\documentclass{article}

\usepackage[arxiv]{iclr2027}
\usepackage[T1]{fontenc}
\usepackage{times}
\usepackage{url}
\usepackage{graphicx}
\usepackage{algorithm}
\usepackage{algorithmic}
\usepackage{booktabs}
\usepackage{xcolor}
\usepackage{colortbl}
\usepackage{amsmath,amssymb}
\usepackage{placeins}
\usepackage{needspace}
\usepackage{marvosym}
\usepackage{hyperref}
\ificlrarxiv
    \hypersetup{hidelinks}
\fi

\usepackage{xspace}

\newcommand{\method}{WorldSync\xspace}
\newcommand{\benchmark}{WorldEcho\xspace}

\definecolor{methodband}{HTML}{EDF5F2}
\definecolor{methodlabel}{HTML}{256B5A}
\definecolor{tablerule}{HTML}{404040}
\newcommand{\tablemethod}[1]{\textcolor{methodlabel}{\textbf{#1}}}

\arrayrulecolor{tablerule}

\title{Do Robotic World Models Really Follow\\Actions? Diagnosing and Aligning\\Action-Conditioned Generation for Policy Learning}
\ificlrarxiv
    \usepackage{tcolorbox}
\definecolor{paperaccent}{HTML}{081B4A}
\definecolor{paperecho}{HTML}{F36A21}
\definecolor{papersync}{HTML}{D91F45}
\definecolor{papermuted}{HTML}{505A60}
\definecolor{paperabstract}{HTML}{F6F5FC}
\definecolor{paperrule}{HTML}{69717D}
\renewcommand{\iclrpreprinttitle}{%
    \makebox[\textwidth][s]{%
        \includegraphics[height=21pt]{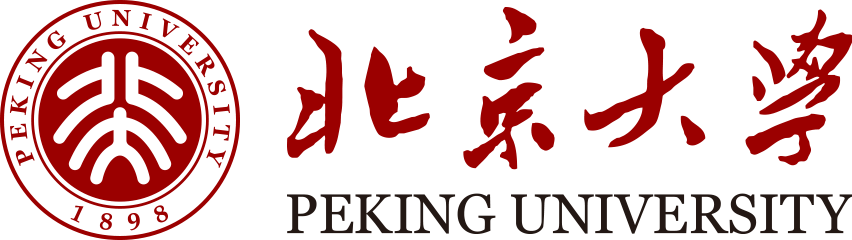}%
        \hfill
        \raisebox{5pt}{\includegraphics[width=91pt]{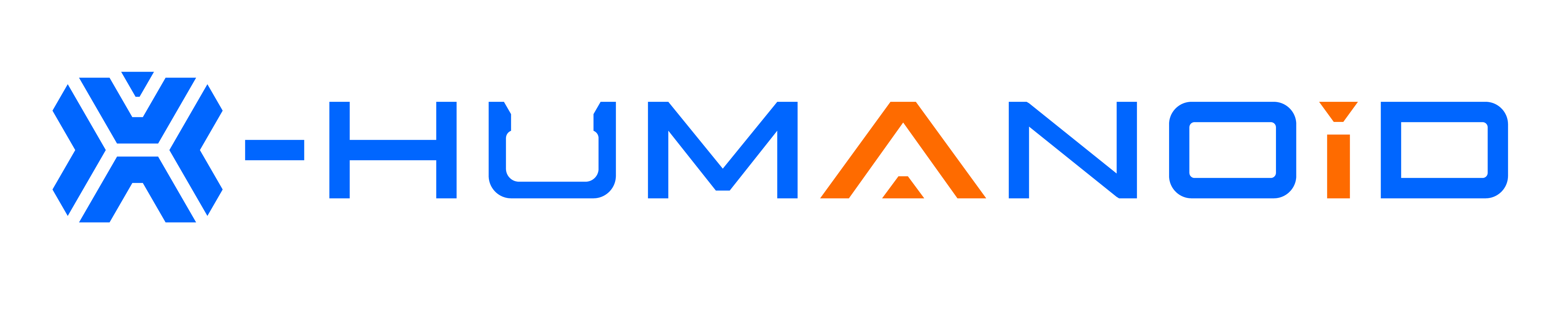}}%
    }\par
    \vspace{4pt}%
    {\color{paperrule}\hrule height 0.5pt}%
    \vspace{12pt}%
    {\normalfont\fontsize{13.5}{16}\selectfont\bfseries
    \raisebox{-4pt}{\includegraphics[height=18pt]{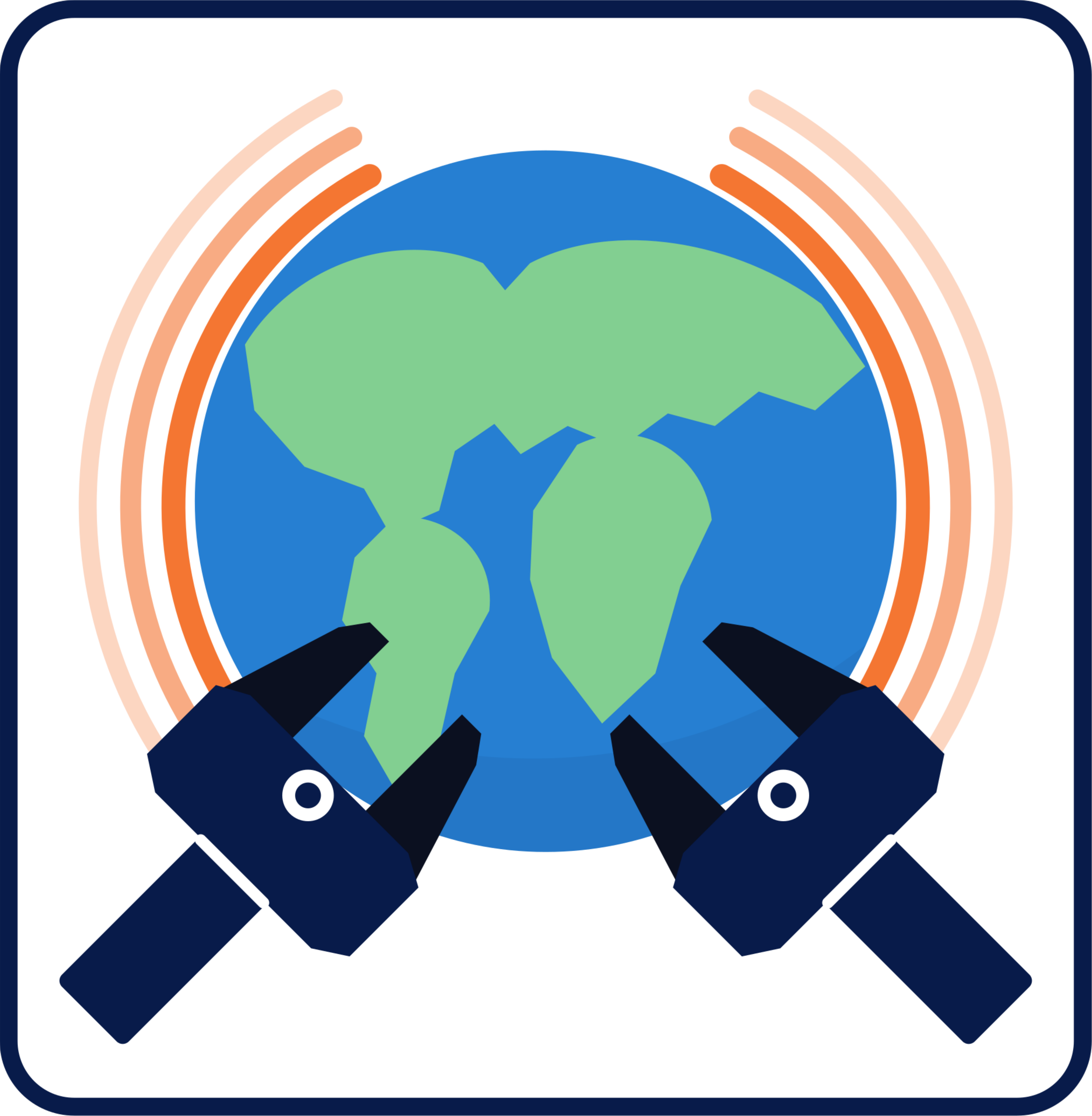}}\hspace{4pt}%
    \textcolor{paperaccent}{World}\textcolor{paperecho}{Echo}\hspace{18pt}%
    \raisebox{-4pt}{\includegraphics[height=18pt]{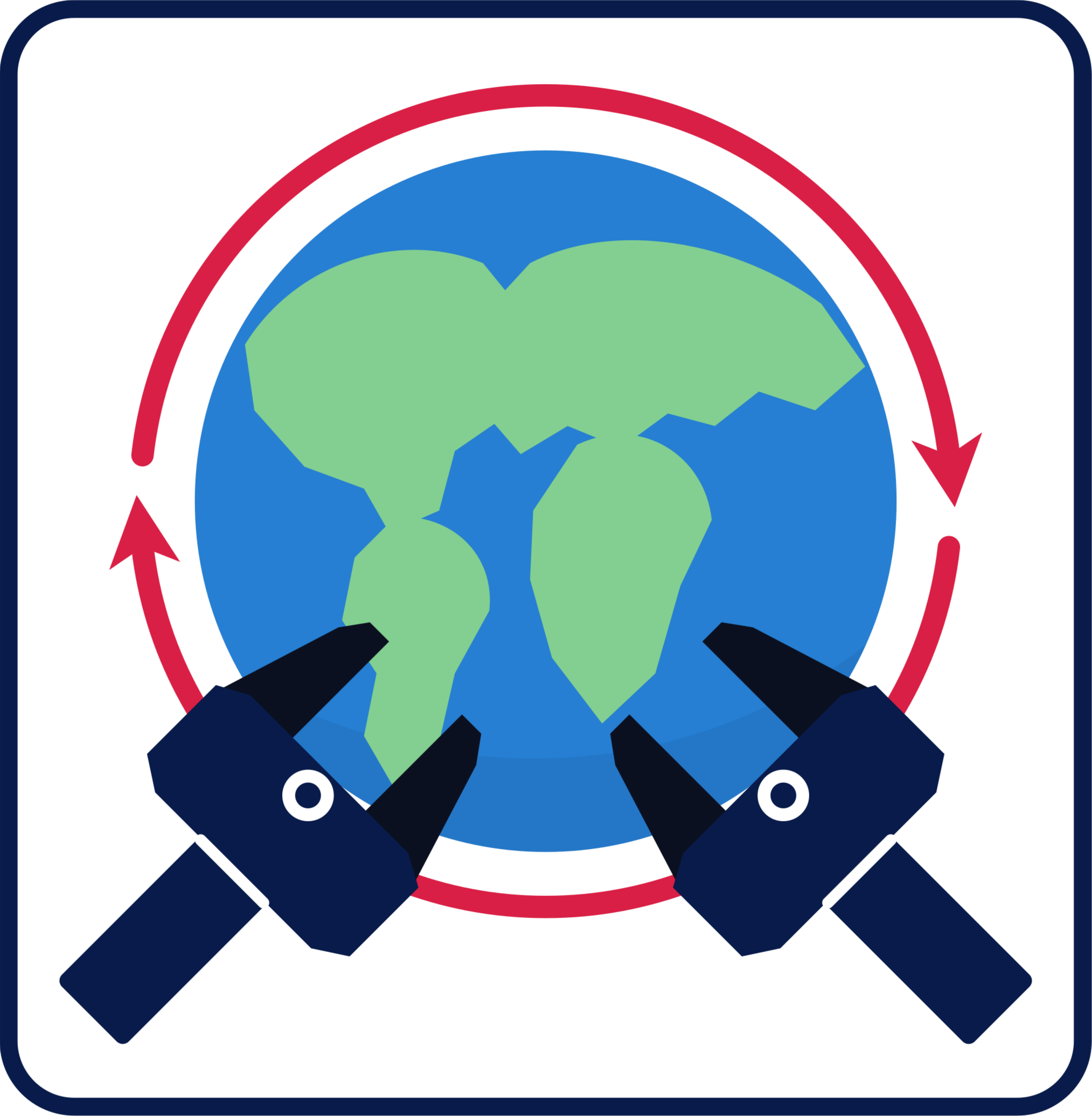}}\hspace{4pt}%
    \textcolor{paperaccent}{World}\textcolor{papersync}{Sync}\par}%
    \vspace{9pt}%
    {\normalfont\fontsize{18}{21}\selectfont\bfseries
    \mbox{Do Robotic World Models Really Follow Actions?}\par}%
    \vspace{4pt}%
    {\normalfont\fontsize{11.5}{14}\selectfont\color{papermuted}%
    \mbox{Diagnosing and Aligning Action-Conditioned Generation for Policy Learning}\par}%
}
\renewenvironment{abstract}{%
    \begin{tcolorbox}[
        colback=paperabstract, colframe=paperabstract,
        boxrule=0pt, arc=5pt,
        left=12pt, right=12pt, top=10pt, bottom=10pt,
        before skip=4pt, after skip=12pt
    ]%
    {\normalfont\normalsize\bfseries Abstract\par}%
    \vspace{5pt}%
    \normalfont\normalsize\noindent\ignorespaces
}{%
    \par\vspace{6pt}%
    {\normalfont\fontsize{9.5}{12}\selectfont\mdseries\upshape
    \noindent \textbf{Webpage:} \textcolor{paperaccent}{%
        \href{https://pku-hmi-lab.github.io/WorldEcho/}{https://pku-hmi-lab.github.io/WorldEcho/}}\\
    \textbf{Code:} \textcolor{paperaccent}{%
        \href{https://github.com/PKU-HMI-Lab/WorldEcho}{https://github.com/PKU-HMI-Lab/WorldEcho}}\par}%
    \end{tcolorbox}%
}
\renewcommand{\iclrpreprintlinks}{}
\hypersetup{pdftitle={Do Robotic World Models Really Follow Actions? Diagnosing and Aligning Action-Conditioned Generation for Policy Learning}}

\fi

\ificlrarxiv
    \ificlrarxiv
\newcommand{\paperindex}[1]{\raisebox{0.65ex}{\normalfont\fontsize{6}{7}\selectfont #1}}
\newcommand{\paperauthor}[2]{\mbox{#1\paperindex{\textcolor{paperaccent}{#2}}}}
\author{\parbox[t]{\textwidth}{\centering\normalfont
    {\fontsize{9}{13}\selectfont
    \paperauthor{Sixiang Chen}{1}\hspace{0.5em}%
    \paperauthor{Jiaming Liu}{1,*}\hspace{0.5em}%
    \paperauthor{Jixian Wu}{2,3,*}\hspace{0.5em}%
    \paperauthor{Yichen Guo}{5,*}\hspace{0.5em}%
    \paperauthor{Tinghao Wang}{2,4,*}\hspace{0.5em}%
    \paperauthor{QR Li}{7}\hspace{0.5em}%
    \paperauthor{Siyuan Qian}{1}\\
    \paperauthor{Peterson Co}{1}\hspace{0.8em}%
    \paperauthor{Hao Chen}{6}\hspace{0.8em}%
    \paperauthor{Jiajun Cao}{2,1}\hspace{0.8em}%
    \paperauthor{Zhoujian Li}{1}\hspace{0.8em}%
    \paperauthor{Lingfei Zhai}{1}\hspace{0.8em}%
    \paperauthor{Jian Tang}{2,\Letter}\hspace{0.8em}%
    \paperauthor{Shanghang Zhang}{1,\Letter}\endgraf}%
    \vspace{6pt}%
    {\fontsize{8.2}{11.2}\selectfont\color{papermuted}
    \mbox{\paperindex{1}State Key Laboratory of Multimedia Information Processing, School of Computer Science, Peking University}\\
    \mbox{\paperindex{2}Beijing Innovation Center of Humanoid Robotics}\quad
    \mbox{\paperindex{4}University of Electronic Science and Technology of China}\\
    \mbox{\paperindex{3}New York University}\quad
    \mbox{\paperindex{5}Nanyang Technological University}\quad
    \mbox{\paperindex{6}The Chinese University of Hong Kong}\quad
    \mbox{\paperindex{7}Tsinghua University}\endgraf}%
    \vspace{4pt}%
    {\fontsize{8}{10}\selectfont\color{papermuted}
    \paperindex{*}Core contributors\hspace{1em}%
    \paperindex{\Letter}Corresponding authors\endgraf}%
}}
\else
\author{\parbox[t]{\dimexpr\textwidth-2\tabcolsep\relax}{\small\raggedright
Sixiang Chen$^1$, Jiaming Liu$^{1,*}$, Jixian Wu$^{2,3,*}$,
Yichen Guo$^{5,*}$, Tinghao Wang$^{2,4,*}$, QR Li$^7$,
Siyuan Qian$^1$, Peterson Co$^1$, Hao Chen$^6$,
Jiajun Cao$^{2,1}$,
Zhoujian Li$^1$, Lingfei Zhai$^1$,
Jian Tang$^2$\textsuperscript{\Letter}, Shanghang Zhang$^1$\textsuperscript{\Letter}\\[4pt]
$^1$State Key Laboratory of Multimedia Information Processing, School of Computer Science, Peking University\\
$^2$Beijing Innovation Center of Humanoid Robotics\\
$^3$New York University \quad $^4$University of Electronic Science and Technology of China\\
$^5$Nanyang Technological University \quad $^6$The Chinese University of Hong Kong\\
$^7$Tsinghua University\\
$^*$Core contributors. \quad \textsuperscript{\Letter}Corresponding authors.
}}
\fi
\hypersetup{pdfauthor={Sixiang Chen, Jiaming Liu, Jixian Wu, Yichen Guo, Tinghao Wang, QR Li, Siyuan Qian, Peterson Co, Hao Chen, Jiajun Cao, Zhoujian Li, Lingfei Zhai, Jian Tang, Shanghang Zhang}}

\else\ificlrfinal
    
\else
    \author{Anonymous authors}
    \hypersetup{pdfauthor={Anonymous authors}}
\fi\fi

\begin{document}
\maketitle

\begin{abstract}

Action-conditioned world models increasingly serve as learned simulators for robotic policy improvement, yet their reliability rests on an unverified assumption: generated futures faithfully reflect the consequences of arbitrary valid actions.
Systematic evaluation against action-specific ground truth remains limited beyond expert demonstrations.
To address this gap, we introduce \benchmark, which evaluates action following over a broader action distribution through visual integrity and $\mathrm{SE}(3)$ end-effector trajectory alignment.
Our diagnosis shows that expert-trained models perform better on demonstrated actions but struggle with off-expert trajectories, producing motion inconsistent with the commanded actions or visually invalid rollouts.
To mitigate these failures, we propose \method, which broadens action coverage, grounds intermediate video representations in action-induced robot dynamics through an Action-Forcing Expert, and aligns predicted and ground-truth changes under action interventions.
Experiments show that \method reduces integrity-gated error on RoboTwin and, when used for iterative policy improvement, yields higher success rates in simulation and on real robots.
\ificlrarxiv\else
\ificlrfinal
Project page: \url{https://pku-hmi-lab.github.io/WorldEcho/}.
\fi
\fi

\end{abstract}
\suppressfloats[t]

\begin{figure}[t]
    \centering
    \includegraphics[width=\textwidth]{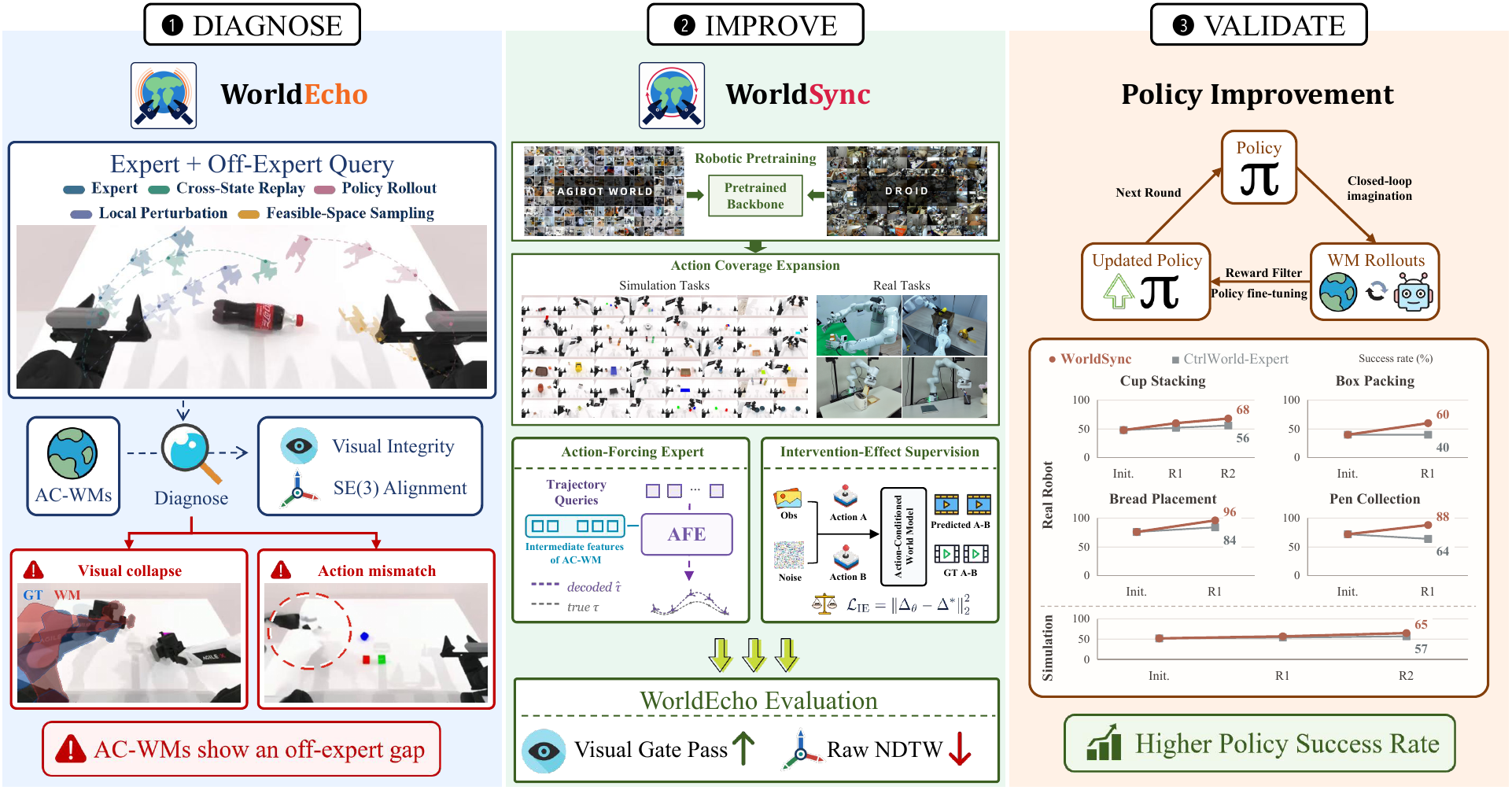}
    \caption{Diagnose--improve--validate pipeline. WorldEcho evaluates action following, WorldSync aligns action-conditioned generation, and simulation and real-robot experiments assess downstream policy improvement.}
    \label{fig:teaser}
\end{figure}

\section{Introduction}

Large-scale pretraining has improved the success rates and generalization of Vision-Language-Action (VLA) models~\citep{RT2,OpenXEmbodiment,pi0.5} and World Action Models (WAM)~\citep{GR1,LingBotVA,DreamZero}.
Task-specific online post-training can further improve downstream performance~\citep{piStar0.6,SOP,VLAW}, but requires costly real-world interaction~\citep{RehearseVLA}.
To reduce this burden, action-conditioned world models (AC-WMs)~\citep{IRASim,CtrlWorld} serve as learned simulators for policy evaluation~\citep{WorldGym,WorldEval} and provide synthetic experience for post-training~\citep{VLAW,WMDAgger,RehearseVLA,VLARFT,WoVR}.
Their use as simulators~\citep{HiWM} raises a critical question: \textbf{do generated futures faithfully reflect the consequences of the queried actions?}

Recent studies report improved policy performance from world-model-generated experience~\citep{VLAW,WoVR}.
Such gains demonstrate downstream utility, but do not directly measure whether generated futures faithfully reflect the queried actions.
Synthetic experience may benefit policy learning through visual augmentation even when its action responses are inaccurate.
Direct evaluation must therefore compare model predictions with the consequences of executing the same actions from the same initial state.
Learned policies may deviate from expert behavior during evaluation and improvement, so these comparisons must also cover off-expert actions~\citep{DAgger,WMDAgger,VLAW}.
Yet existing benchmarks mainly assess perceptual and semantic quality, reference similarity, or downstream executability~\citep{EWMBench,WorldArena,RoboWMBench}.
Although recent studies probe out-of-distribution or failure-inducing actions~\citep{WorldGym,MiraBench}, systematic comparisons against action-specific ground truth across broad numerical action queries remain limited.

To address this gap, \benchmark evaluates action following on demonstrated queries and four complementary off-expert categories (Figure~\ref{fig:teaser}).
Each query is replayed in RoboTwin to obtain its ground-truth future, including unsuccessful outcomes.
We assess generated rollouts through visual integrity and $\mathrm{SE}(3)$ alignment between generated and reference end-effector trajectories.
Relative to demonstrated queries, off-expert queries increase integrity-gated error across six expert-trained models (Figure~\ref{fig:diagnosis-coverage}), with both visually plausible action mismatches and visual degradation (Figure~\ref{fig:motivation}).

These findings motivate \method, which combines distributional coverage, representational grounding, and intervention-effect alignment.
Simulated expert and off-expert trajectories broaden action support, while a small amount of target-domain real-world data preserves visual fidelity.
Because coverage alone does not ground intermediate video representations in action consequences, an \emph{Action-Forcing Expert} (AFE) predicts future end-effector state changes.
AFE supervises individual rollouts but does not explicitly constrain changes across actions.
\emph{Intervention-Effect} (IE) supervision therefore aligns predicted and ground-truth changes across paired trajectories with shared observations and different actions.

On RoboTwin~\citep{RoboTwin2.0}, \method with and without additional robotic pretraining achieves the two lowest integrity-gated error point estimates among evaluated configurations (Table~\ref{tab:main}).
When used for policy improvement, it achieves higher success rates than expert-trained CtrlWorld in simulation and across four real-robot tasks on three platforms (Figure~\ref{fig:policy-improvement}).

\textbf{Our contributions are as follows.}
(1) \benchmark measures visual integrity and action-specific $\mathrm{SE}(3)$ trajectory alignment across demonstrated and off-expert queries.
(2) Our evaluation characterizes off-expert failures across six world models, distinguishing action-inconsistent motion from visual degradation.
(3) \method improves action following through expanded coverage, AFE, and IE supervision, with downstream evaluation of policy improvement.

\section{Related Work}

\subsection{Action-Conditioned Robotic World Models}
AC-WMs advance action control through architecture \citep{IRASim,CtrlWorld,MemWorld}, broader pretraining \citep{DreamDojo,OSCAR,A2World}, and spatially grounded representations \citep{OSCAR,BridgeV2W,ViPSim,EAWM}.
Perturbations and counterfactual behaviors extend beyond standard replay, but provide mostly indirect or coarse-grained evidence of action following rather than ground-truth motion for each command \citep{IRASim,DreamDojo,A2World,Mask2RealWM,MiraBench}.

\subsection{World Models for Policy Evaluation and Improvement}
AC-WMs support policy evaluation \citep{WorldGym,WorldEval,RoboWorld,PiLWorld} and improvement using synthetic experience \citep{WMDAgger,RehearseVLA,VLARFT,WoVR}.
To handle policy queries beyond expert-dominated data, systems use exploratory or corrective interactions \citep{PlayWorld,HiWM,RehearseVLA}, generation filtering, or joint policy and world-model updates \citep{WoVR,VLAW,WorldVLALoop}.
These downstream gains motivate direct evaluation of the queried actions' consequences.

\subsection{Robotic World Model Evaluation}
Benchmarks assess motion and physical consistency \citep{EWMBench,WorldArena,WorldArena2}, or recover actions to test executability \citep{RoboWMBench,WoWWorldEval}.
Most closely related, MiraBench shows that visually plausible predictions may ignore failure-inducing perturbations or exhibit optimism bias \citep{MiraBench}.
Beyond image-space motion, recovered-action execution, and task-level failure judgments, \benchmark aligns generated end-effector trajectories with simulator-replayed references in $\mathrm{SE}(3)$ across broader numerical action queries.

\section{Method}

\subsection{Problem Formulation}

Given a current multi-view observation $o_0$, language instruction $c$, and
$H$-step numerical action sequence $a_{1:H}$, an action-conditioned world model
generates a future multi-view video
\begin{equation}
    \hat{I}_{1:H}\sim p_{\theta}(\cdot\mid o_0,c,a_{1:H}).
\end{equation}
Executing the same actions from the corresponding initial environment state
gives the ground-truth future $I^{\mathrm{GT}}_{1:H}$.
For a trajectory extractor $\Phi$, the generated and reference end-effector
trajectories are $\hat{\tau}=\Phi(\hat{I}_{1:H})$ and
$\tau^{\mathrm{GT}}=\Phi(I^{\mathrm{GT}}_{1:H})$.
Our goal is to evaluate and improve visual integrity and agreement with the
action-specific reference trajectories beyond expert demonstrations.

\Needspace{0.28\textheight}
\subsection{Motivation}
\suppressfloats[t]

\noindent\begin{minipage}{\textwidth}
\begin{minipage}[t]{0.46\textwidth}
    \vspace{0pt}
    \centering
    \includegraphics[width=\linewidth]{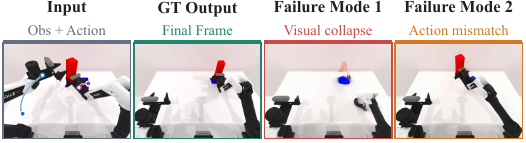}
    \makeatletter\def\@captype{figure}\makeatother
    \caption{Off-expert failures. Left to right: input and actions,
    ground-truth final frame, visual collapse, and plausible but
    action-inconsistent motion.}
    \label{fig:motivation}
\end{minipage}\hfill
\begin{minipage}[t]{0.51\textwidth}
\vspace{0pt}
Following prior work~\citep{IRASim,CtrlWorld}, we fine-tune
Cosmos-Predict2.5~\citep{CosmosPredict2.5} on RoboTwin expert
trajectories~\citep{RoboTwin,RoboTwin2.0}.
From the same initial state, we compare predictions for expert and feasible
off-expert actions with matched simulator futures.
Expert actions are replayed accurately, but off-expert predictions lose
visual integrity or show plausible, action-inconsistent motion
(Figure~\ref{fig:motivation}). These failures motivate evaluating both
visual integrity and action-induced motion beyond expert demonstrations.
\end{minipage}
\end{minipage}

\subsection{\benchmark}

\begin{figure}[!htbp]
    \centering
    \includegraphics[width=\textwidth]{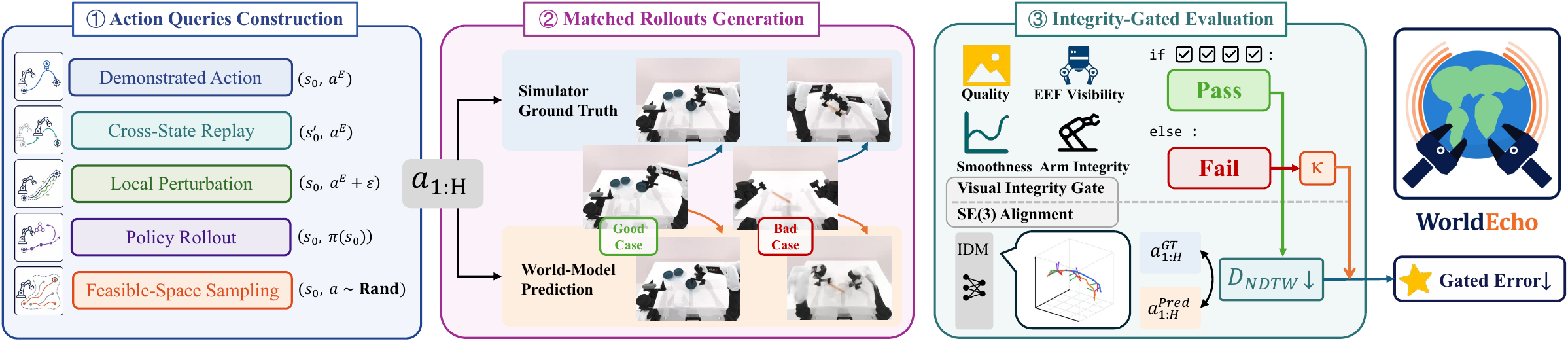}
    \caption{\benchmark evaluation across five action-query categories, combining visual integrity with $\mathrm{SE}(3)$ end-effector trajectory alignment against matched simulator references.}
    \label{fig:benchmark}
\end{figure}

\benchmark evaluates visual integrity and trajectory alignment against
action-specific simulator references across demonstrated and off-expert
queries (Figure~\ref{fig:benchmark}).

\subsubsection{Action Query Construction}

\emph{Demonstrated Action} replays the expert sequence for the current
observation as an in-distribution baseline. Four off-expert categories probe
beyond this support. \emph{Cross-State Replay} transfers expert actions to
another initial state, testing context-dependent effects including failure.
\emph{Local Perturbation} applies bounded changes to expert actions to test
sensitivity to nearby alternatives. \emph{Policy Rollout} uses learned-policy
actions to capture deviations during evaluation and improvement. Beyond
these policy-dependent queries, \emph{Feasible-Space Sampling} draws valid
actions from the broader action space to test controllability. All off-expert
queries undergo feasibility filtering before replay in RoboTwin.

\subsubsection{Visual Integrity Assessment}

We assess four aspects of visual integrity so trajectory comparisons concern
valid depictions of the robot and scene. Following WorldArena~\citep{WorldArena},
image quality $q$ uses MUSIQ~\citep{MUSIQ}, while motion smoothness $m$ uses
frame-interpolation consistency~\citep{VFIMamba}.
With $\mathbb{I}$ denoting an indicator, thresholds $\tau_q,\tau_m$ give
$G_{\mathrm{quality}}=\mathbb{I}[q\geq\tau_q]$ and
$G_{\mathrm{motion}}=\mathbb{I}[m\geq\tau_m]$.
For end-effector visibility, $G_{\mathrm{EEF}}$ passes when either SAM-based
video tracking or framewise text-prompted detection meets its requirement~\citep{SAM3}.
Arm integrity $G_{\mathrm{arm}}$ is assessed by a vision-language evaluator
that detects blurred, broken, or disappearing arms~\citep{Qwen3VL}.
Using the same rules and thresholds for all models, we define
\begin{equation}
    G_{\mathrm{vis}}
    =G_{\mathrm{quality}}\land G_{\mathrm{motion}}
    \land G_{\mathrm{EEF}}\land G_{\mathrm{arm}}.
\end{equation}

\subsubsection{End-Effector Trajectory Alignment}

Visual validity does not ensure that motion follows the queried actions, so
we compare generated and ground-truth end-effector trajectories.
We use an inverse dynamics model (IDM) $\Phi$~\citep{AnyPos} to extract
end-effector positions $p_t^e\in\mathbb{R}^3$ and rotations $R_t^e\in\mathrm{SO}(3)$
from each video, where $e\in\{L,R\}$ denotes the left or right end effector.
Across all 1,750 \benchmark reference videos, frame-aligned IDM predictions
have mean Euclidean position and geodesic rotation errors of 3.28~mm and
$1.42^\circ$ against simulator states, averaged over frames and both arms.
For generated frame $i$ and reference frame $j$, the local pose discrepancy is
\begin{equation}
    d_e(i,j)=\Big[
        w_p^2\lVert\hat p_i^e-p_j^{e,\mathrm{GT}}\rVert_2^2
        +w_R^2 d_{\mathrm{SO}(3)}^2
        (\hat R_i^e,R_j^{e,\mathrm{GT}})
    \Big]^{1/2},
\end{equation}
where hats and GT denote generated and reference poses, and $w_p,w_R$ weight
translation and rotation. The rotation distance is
\begin{equation}
    d_{\mathrm{SO}(3)}(R_1,R_2)
        =\arccos\!\left(\operatorname{clip}(\xi,-1,1)\right),
    \qquad
    \xi=\frac{\operatorname{tr}(R_1^\top R_2)-1}{2},
\end{equation}
with $\operatorname{tr}$ denoting the trace and $\operatorname{clip}$ clamping
to $[-1,1]$.
To accommodate temporal misalignment, we use pose-aware normalized dynamic
time warping (NDTW)~\citep{DTW,FastDTW,EWMBench,WorldArena}:
\begin{equation}
    D_{\mathrm{NDTW}}=\frac{1}{|\mathcal{A}|}
    \sum_{e\in\mathcal{A}}\frac{1}{|\pi_e|}
    \sum_{(i,j)\in\pi_e}d_e(i,j),
\end{equation}
where $\mathcal{A}$ contains valid end effectors and $\pi_e$ is the optimal
warping path for $e$.
We normalize by path length, but not by the reference trajectory's spatial
extent, to preserve
absolute pose error and avoid amplifying pose-estimation noise for short motions.

\subsubsection{Evaluation Protocol}

To retain diagnostic information, we compute raw NDTW $D_n^{\mathrm{NDTW}}$
for every query $n$, including visually invalid rollouts.
Given its visual-gate result $G_n$, the official error instead assigns a
fixed penalty $\kappa$ to visual failures:
\begin{equation}
    S_n=
    \begin{cases}
        D_n^{\mathrm{NDTW}}, & G_n=1,\\
        \kappa, & G_n=0.
    \end{cases}
\end{equation}
We first average $S_n$ within each task, then equally across tasks, so unequal
query counts do not dominate the ranking.
We also report visual pass rate and raw NDTW.
Algorithm~\ref{alg:benchmark} summarizes evaluation;
Appendices~\ref{app:query-category-results}, \ref{app:metric-distributions-thresholds},
and~\ref{app:human-evaluation-agreement} provide query-category results,
metric distributions, and human comparisons, respectively.

\begin{algorithm}[!hb]
\caption{Integrity-Gated Action-Following Evaluation}
\label{alg:benchmark}
\begin{algorithmic}[1]
\REQUIRE AC-WM $\mathcal{W}_{\theta}$, action queries $\mathcal{Q}$,
failure penalty $\kappa$
\FOR{each query $n$: $(o_0,c,a_{1:H},I_{1:H}^{\mathrm{GT}})\in\mathcal{Q}$}
    \STATE Generate $\hat I_{1:H}\sim
    p_{\theta}(\cdot\mid o_0,c,a_{1:H})$
    \STATE Evaluate $G_n=G_{\mathrm{vis}}(\hat I_{1:H})$
    \STATE Extract $\hat\tau=\Phi(\hat I_{1:H})$ and
    $\tau^{\mathrm{GT}}=\Phi(I_{1:H}^{\mathrm{GT}})$
    \STATE Compute pose-aware NDTW error $D_n^{\mathrm{NDTW}}$
    \STATE Set $S_n\leftarrow D_n^{\mathrm{NDTW}}$ if $G_n=1$; otherwise
    $S_n\leftarrow\kappa$
\ENDFOR
\STATE Aggregate $\{S_n\}$ within each task and then across tasks
\RETURN task-macro integrity-gated error, visual pass rate, and raw NDTW errors
\end{algorithmic}
\end{algorithm}

\subsection{\method}

\begin{figure}[!htbp]
    \centering
    \includegraphics[width=\textwidth]{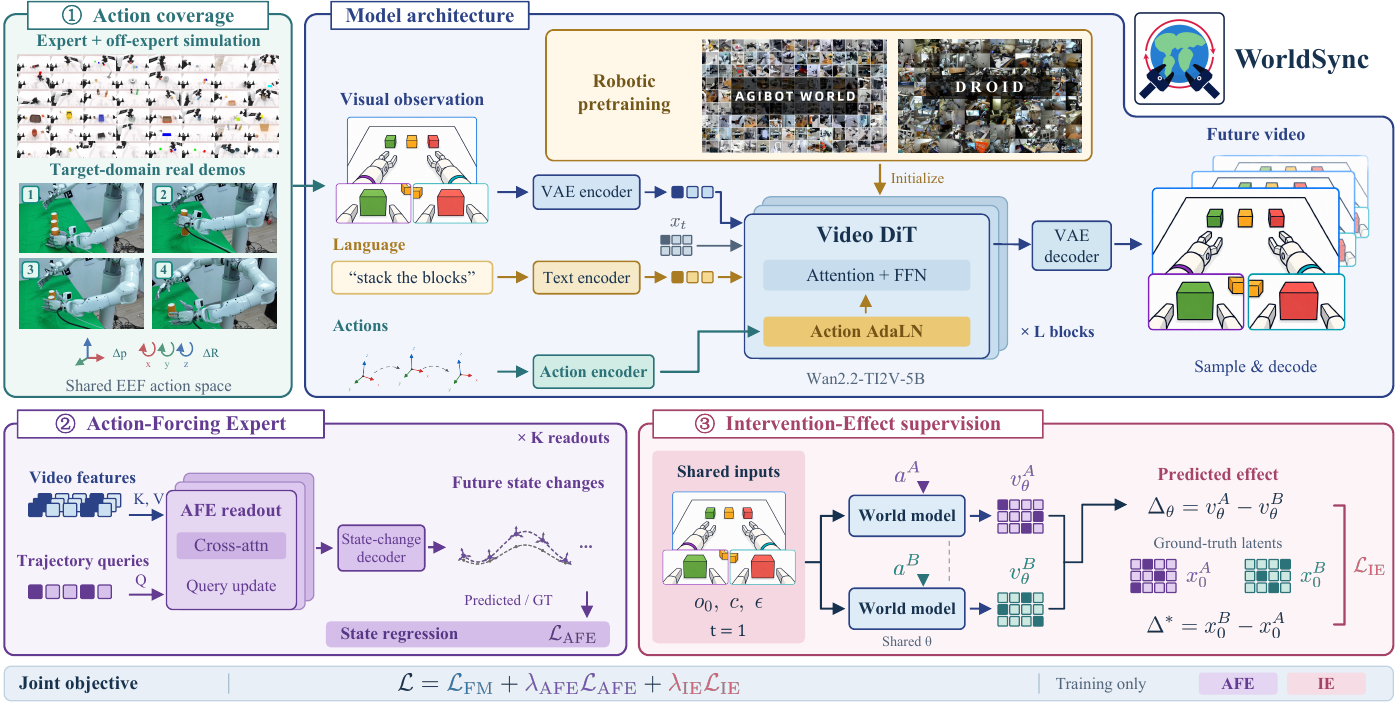}
    \caption{\method overview: action coverage, the pretrained video backbone, and training-only supervision from the Action-Forcing Expert (AFE) and Intervention-Effect (IE) objectives.}
    \label{fig:method-overview}
\end{figure}

To address these failures, \method combines expanded action coverage with
AFE feature grounding and IE supervision of action-dependent changes
(Figure~\ref{fig:method-overview}).

We condition the video blocks of Wan2.2-TI2V-5B~\citep{Wan2.2} on robot
actions through Action AdaLN (Appendix~\ref{app:action-conditioning}) and
train the backbone by flow matching. For clean target-video latent $x_0$, Gaussian noise
$\epsilon\sim\mathcal{N}(0,I)$, and flow time $t\in[0,1]$, we construct
$x_t=(1-t)x_0+t\epsilon$ and optimize
\begin{equation}
    \mathcal{L}_{\mathrm{FM}}
    =\mathbb{E}_{x_0,\epsilon,t}
    \left[
    \left\lVert
    v_\theta(x_t,t\mid o_0,c,a_{1:H})-(\epsilon-x_0)
    \right\rVert_2^2
    \right],
\end{equation}
where $v_\theta$ is the world model's action-conditioned flow velocity.

\paragraph{Robotic Pretraining.}
For the additional robotic pretraining variant, we first optimize the backbone
on approximately 1,011 hours of robot interaction data from
AgiBot-World-Alpha~\citep{AgiBotWorld} and DROID~\citep{DROID} using
$\mathcal{L}_{\mathrm{FM}}$. AFE and IE are introduced during subsequent
\method training.
Dataset statistics and pretraining configurations are provided in
Appendix~\ref{app:robotic-pretraining}.

\subsubsection{Action Coverage Expansion}

We broaden expert-only training with multi-task simulated expert trajectories,
local perturbations, cross-state replays, policy rollouts, and broad feasible actions.
We mix in a small set of target-task real-robot demonstrations to preserve
target-domain visual fidelity. Across simulated and real-robot data, actions
use relative Cartesian end-effector pose displacements in the robot base frame
and gripper opening. This shared representation allows the model to learn
relationships between actions and their consequences across the two domains.
Appendix~\ref{app:data-composition-ablations} examines the effect of training-data composition.

\subsubsection{Action-Forcing Expert}

Broader coverage alone does not ensure that intermediate video representations
encode action-induced robot dynamics. To provide an auxiliary feature-level
grounding signal, AFE maintains one learnable trajectory query per future step.
These queries progressively cross-attend to the intermediate features of successive video blocks and
predict future end-effector state changes relative to the initial state.
The targets include both arms' position, rotation, and gripper state.
Given standardized cumulative-change targets $y_{1:H}$ and predictions
$\hat y_{1:H}$, we optimize
\begin{equation}
    \mathcal{L}_{\mathrm{AFE}}
    =\operatorname{MSE}_{\mathrm{valid}}(\hat y_{1:H},y_{1:H}),
\end{equation}
where the masked mean squared error uses valid state labels.
Appendix~\ref{app:afe-ablations} specifies target preprocessing and design choices.
AFE neither reads actions directly nor writes back to the video stream;
its loss updates the backbone through video features, and the expert is
removed at inference time.

\subsubsection{Intervention-Effect Supervision}

While AFE grounds individual rollouts, it does not directly supervise how
the generated future should change when the conditioned action changes. IE
provides this complementary relational signal through paired trajectories
that share the current observation and instruction but execute different actions.
To isolate the effect of changing actions, we evaluate both branches at the noise
endpoint $t=1$, where shared noise gives identical input latents $x_1^A=x_1^B=\epsilon$.
With the observation and instruction also shared, only the action inputs differ.
The predicted intervention effect is
$\Delta_\theta=v_\theta^A-v_\theta^B$. Since the shared noise cancels
between the target velocities, the corresponding ground-truth effect is
$\Delta^*=x_0^B-x_0^A$. We align these effects over future video latents:
\begin{equation}
    \mathcal{L}_{\mathrm{IE}}
    =\left\lVert\Delta_\theta-\Delta^*\right\rVert_2^2.
\end{equation}

\subsubsection{Training Objective}

Combining the flow-matching generation loss with the two auxiliary
objectives gives
\begin{equation}
    \mathcal{L}
    =\mathcal{L}_{\mathrm{FM}}
    +\lambda_{\mathrm{AFE}}\mathcal{L}_{\mathrm{AFE}}
    +\lambda_{\mathrm{IE}}\mathcal{L}_{\mathrm{IE}}.
\end{equation}
We apply $\mathcal{L}_{\mathrm{AFE}}$ when future state labels are
available and $\mathcal{L}_{\mathrm{IE}}$ to intervention pairs.

\section{Experiments}
\label{sec:experiments}

\subsection{Experimental Setup}
\label{sec:exp:setup}

\paragraph{Benchmarks and evaluation sets.}
We evaluate action following on 50 RoboTwin tasks~\citep{RoboTwin,RoboTwin2.0}
using 1,750 queries from \benchmark's five categories, and component contributions on ten
tasks under the same protocol. Policy improvement is evaluated in RoboTwin
and on four real-robot tasks across three platforms
(Appendix~\ref{app:real-robot-setup}).

\paragraph{Baselines.}
We compare \method with CtrlWorld~\citep{CtrlWorld},
Cosmos-Predict2.5~\citep{CosmosPredict2.5}, Cosmos3~\citep{Cosmos3},
DreamDojo~\citep{DreamDojo}, Motus~\citep{Motus}, and LingBotVA~\citep{LingBotVA}.
Each baseline is trained with Expert Demonstrations or Expanded Action Coverage
and evaluated on the same task split and queries. All baselines and both
\method variants are evaluated after 40k fine-tuning updates.

\paragraph{Metrics.}
Integrity-gated error is the primary metric. Raw pose-aware NDTW and visual
pass rate distinguish action mismatch from visual failure. Raw and gated
errors are task-macro averages. Visual pass rate is the fraction of accepted
rollouts. Table~\ref{tab:main} specifies video-metric aggregation. Video metrics are evaluated on all 1,750 queries.

\subsection{Action-Following Evaluation}
\label{sec:exp:action-following}
\label{sec:exp:diagnosis}

\begin{figure}[!htbp]
    \centering
    \includegraphics[width=\textwidth]{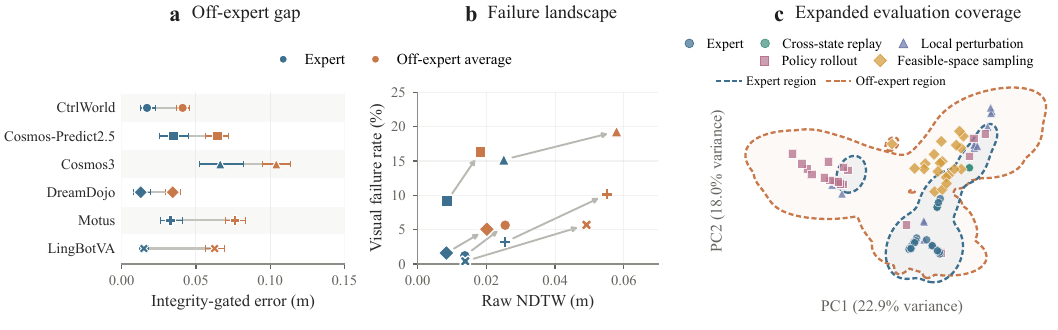}
    \caption{Off-expert action following. (a) Gated error with equally weighted off-expert categories within tasks. Error bars show 95\% confidence intervals from task-level resampling. (b) Expert-to-off-expert changes in raw NDTW and visual failure rate. (c) Action PCA with 95\% highest-density regions (off-expert: union across categories).}
    \label{fig:diagnosis-coverage}
\end{figure}

\textbf{Expert-only evaluation understates off-expert failures.}
Across all six expert-trained models, off-expert queries increased gated error
by 0.022--0.048~m (Figure~\ref{fig:diagnosis-coverage}a).
This gap involved both trajectory inconsistency and visual degradation:
raw NDTW increased by 0.010--0.035~m and visual failure rate by 3.45--7.15
percentage points across all models (Figure~\ref{fig:diagnosis-coverage}b).
Their relative contributions varied by model, so either metric alone would
miss part of the failure.
The off-expert categories occupy a broader region than expert actions in the
projected action space (Figure~\ref{fig:diagnosis-coverage}c). This expanded
coverage exposes errors beyond expert replay, motivating both broader training
support and stronger dependence on the queried actions.

\begin{table}[!htbp]
    \centering
    \footnotesize
    \renewcommand{\arraystretch}{1.12}
    \setlength{\tabcolsep}{1.5pt}
    \caption{Main \benchmark comparison. Expert and Expanded denote Expert
    Demonstrations and Expanded Action Coverage. Bold marks the best scores.
    Underlining marks second-best \benchmark scores. Emb. PT marks embodied-data
    pretraining before task-specific fine-tuning, excluding generic-only video
    pretraining (Appendix~\ref{app:baseline-pretraining}). $\dagger$ Video metrics use
    reconstructed references, not verified as pixel-identical to the original
    RGB videos. Baseline video and \benchmark scores use separate prediction batches
    at the same checkpoint. PSNR, SSIM, and LPIPS average over frames,
    clips, then tasks. FVD pools all clips.}
    \label{tab:main}
    \begin{tabular}{@{}lclccccccc@{}}
        \toprule
        & & & \multicolumn{3}{c}{\textbf{\benchmark metrics}} & \multicolumn{4}{c}{\textbf{Video generation metrics} $\dagger$} \\
        \cmidrule(lr){4-6}\cmidrule(l){7-10}
        \textbf{Model} & \shortstack{\textbf{Emb.}\\\textbf{PT}} & \shortstack{\textbf{Training}\\\textbf{regime}} & \shortstack{\textbf{Gated}\\\textbf{error} $\downarrow$} & \shortstack{\textbf{Raw}\\\textbf{NDTW} $\downarrow$} & \shortstack{\textbf{Visual}\\\textbf{pass (\%)} $\uparrow$} & \textbf{PSNR} $\uparrow$ & \textbf{SSIM} $\uparrow$ & \textbf{LPIPS} $\downarrow$ & \textbf{FVD} $\downarrow$ \\
        \midrule
        CtrlWorld & $\checkmark$ & Expert & 0.0379 & 0.0266 & 95.83 & 19.41 & 0.8466 & 0.1137 & 113.58 \\
        CtrlWorld & $\checkmark$ & Expanded & 0.0293 & 0.0210 & 96.97 & 20.83 & 0.8665 & 0.0915 & 69.58 \\
        Cosmos-Predict2.5 & $\checkmark$ & Expert & 0.0551 & 0.0190 & 87.09 & 21.10 & 0.8750 & 0.0848 & 95.69 \\
        Cosmos-Predict2.5 & $\checkmark$ & Expanded & 0.0445 & \textbf{0.0127} & 88.80 & \textbf{22.95} & 0.8952 & \textbf{0.0658} & 51.89 \\
        Cosmos3 & $\checkmark$ & Expert & 0.0989 & 0.0572 & 82.51 & 14.45 & 0.6846 & 0.1833 & 122.58 \\
        Cosmos3 & $\checkmark$ & Expanded & 0.0799 & 0.0419 & 85.09 & 15.10 & 0.7003 & 0.1525 & 86.36 \\
        DreamDojo & $\checkmark$ & Expert & 0.0313 & 0.0210 & 96.29 & 20.49 & 0.8909 & 0.1049 & 81.83 \\
        DreamDojo & $\checkmark$ & Expanded & 0.0237 & \underline{0.0151} & 96.97 & 21.17 & \textbf{0.8959} & 0.0949 & \textbf{30.77} \\
        Motus & $\checkmark$ & Expert & 0.0708 & 0.0548 & 91.94 & 17.21 & 0.8164 & 0.1620 & 106.76 \\
        Motus & $\checkmark$ & Expanded & 0.0371 & 0.0292 & 97.14 & 19.22 & 0.8448 & 0.1182 & 68.78 \\
        LingBotVA & $\checkmark$ & Expert & 0.0572 & 0.0473 & 95.60 & 16.54 & 0.8084 & 0.1717 & 128.21 \\
        LingBotVA & $\checkmark$ & Expanded & 0.0446 & 0.0340 & 96.06 & 17.60 & 0.8272 & 0.1450 & 81.16 \\
        \addlinespace[2pt]
        \rowcolor{methodband}
        \tablemethod{\method} & -- & Expanded & \underline{0.0234} & 0.0176 & \underline{97.94} & 21.42 & 0.8743 & 0.0875 & 46.97 \\
        \rowcolor{methodband}
        \tablemethod{\method w/ pretrain} & $\checkmark$ & Expanded & \textbf{0.0228} & 0.0180 & \textbf{98.23} & 21.36 & 0.8738 & 0.0887 & 45.61 \\
        \bottomrule
    \end{tabular}
\end{table}

\textbf{Expanded-coverage models improve all three \benchmark metrics.}
Compared with Expert Demonstrations, Expanded Action Coverage lowers gated
error and raw NDTW and raises visual pass rates for all six baselines
(Table~\ref{tab:main}), improving both trajectory alignment and visual integrity.

\textbf{\method achieves the lowest integrity-gated error.}
Without additional robotic pretraining, \method achieves a lower gated-error
point estimate than every baseline (0.0234 versus 0.0237 for the strongest
baseline, expanded-coverage DreamDojo). Additional robotic pretraining further
lowers this estimate to 0.0228. The pretrained and non-pretrained variants' visual pass rates were also
highest (98.23\% and 97.94\%, versus 97.14\% for expanded-coverage Motus).
However, expanded-coverage Cosmos-Predict2.5 had lower raw NDTW
(0.0127 versus 0.0180 and 0.0176), so the gated ranking does not imply
the lowest trajectory error. Both \method variants ranked among the top
three on PSNR and FVD. Relative to the non-pretrained variant,
pretraining improved visual pass rate but slightly increased raw NDTW.
Appendix~\ref{app:additional-qualitative-results} presents qualitative comparisons and per-case metrics.

\subsection{Policy Improvement}
\label{sec:exp:policy-improvement}

We adapt the VLAW framework~\citep{VLAW} to compare \method with expert-trained
CtrlWorld under a shared policy-improvement procedure. Both conditions use
the same initial policy and matched interaction, world-model rollout, and
policy-training budgets within each task. Appendix~\ref{app:policy-improvement-pipeline}
details the simulation procedure and condition-specific world-model adaptation.

\begin{figure}[!htbp]
    \centering
    \includegraphics[width=\textwidth]{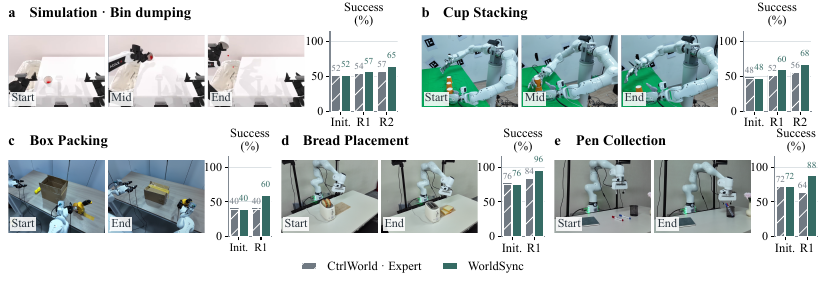}
    \caption{Policy improvement in simulation and four real-robot tasks. Each real-robot success rate uses 25 trials. Init. denotes the initial policy, and R1/R2 denote improvement rounds.}
    \label{fig:policy-improvement}
\end{figure}

\textbf{Simulation.}
On Bin Dumping, both conditions started at 52\% success. \method reached 65\%
versus 57\% for CtrlWorld after two rounds (Figure~\ref{fig:policy-improvement}a).
From 3,000 imagined rollouts per round, the same task-progress and
action-consistency filters retained 553 versus 452 rollouts in Round 1 and
365 versus 323 in Round 2 (Appendix~\ref{app:policy-improvement-results}).
Thus, \method provided 22.3\% and 13.0\% more usable imagined rollouts
under the same generation budget, alongside larger policy gains.

\textbf{Real-robot tasks.}
The four tasks span three manipulation platforms
(Figure~\ref{fig:policy-improvement}b--e and Appendix~\ref{app:real-robot-tasks}).
Cup Stacking uses two TianJi arms with Sharpa dexterous hands to nest
three cups through sequential left- and right-hand placements.
Box Packing uses the dual-arm Simplexity i7 Pro with Xense grippers to
place an object in a box, close its flaps, and tape the top seam.
On the single-arm Franka Research 3, Bread Placement transfers two slices
from a toaster to a cutting board, while Pen Collection places two pens
into a holder. These tasks cover coordinated dual-arm manipulation,
dexterous-hand control, and sequential single-arm pick-and-place.

\textbf{Real-robot data.}
The demonstration datasets contain 200 trajectories for Cup Stacking,
200 for Box Packing, 300 for Bread Placement, and 500 for Pen Collection
(Appendix~\ref{app:robot-platforms}). All three platforms provide workspace
and wrist camera observations, with training trajectories sampled at 10\,Hz.
The TianJi platform controls 22 independently actuated degrees of freedom
per hand, while the i7 Pro and Franka platforms use gripper-opening commands.
Success requires completing the full task sequence, including closing and
sealing the box or placing both bread slices or pens at their targets.

\textbf{Real-robot results.}
Each success rate uses 25 trials. On Cup Stacking, both conditions started
at 48\%. \method reached 68\% versus 56\% after two rounds.
The other tasks use one round. Box Packing improved from 40\% to 60\%
with \method while CtrlWorld remained at 40\%. Bread Placement improved
from 76\% to 96\% versus 84\%. Pen Collection improved from 72\% to
88\% versus 64\%. \method therefore improved the initial policy on all
four tasks, with final success rates 12--24 percentage points above
CtrlWorld. The gains span both dexterous hands and grippers, including
Box Packing and Pen Collection, where CtrlWorld yielded no improvement
or a decline. Together with the simulation results, these experiments
connect stronger \benchmark performance with larger downstream policy gains.

\subsection{Component Contributions and Interactions}
\label{sec:exp:ablation}

\begin{table}[htbp]
    \centering
    \footnotesize
    \renewcommand{\arraystretch}{1.12}
    \setlength{\tabcolsep}{2.5pt}
    \caption{Data and component ablation on ten RoboTwin tasks at 40k training steps.
    Bold denotes the best reported value, including ties. Video metrics use separate
    generations at the same checkpoint. PSNR, SSIM, and LPIPS average over
    frames, clips, then tasks. FVD pools all video pairs.}
    \label{tab:ablation}
    \begin{tabular}{@{}lccccccc@{}}
        \toprule
        Variant & Gated $\downarrow$ & Raw $\downarrow$ & Pass (\%) $\uparrow$ & PSNR $\uparrow$ & SSIM $\uparrow$ & LPIPS $\downarrow$ & FVD $\downarrow$ \\
        \midrule
        Expert Base & 0.0352 & 0.0314 & \textbf{98.57} & 21.23 & 0.8585 & 0.1154 & 175.09 \\
        Base & 0.0264 & 0.0204 & 97.86 & 20.66 & 0.8591 & 0.1087 & 141.10 \\
        + IE & 0.0214 & 0.0176 & \textbf{98.57} & 21.05 & 0.8665 & 0.0963 & 125.92 \\
        + AFE & 0.0232 & 0.0172 & 97.86 & 21.30 & 0.8683 & 0.0962 & 118.81 \\
        \rowcolor{methodband}
        \tablemethod{IE + AFE} & \textbf{0.0193} & \textbf{0.0138} & 97.86 & \textbf{22.33} & \textbf{0.8821} & \textbf{0.0830} & \textbf{105.83} \\
        \bottomrule
    \end{tabular}
\end{table}

In Table~\ref{tab:ablation}, both Base variants disable IE and AFE. Expert Base
uses expert-only data. All other rows use Expanded data. Expanded Base lowers
gated and raw errors by 25.0\%
and 35.0\%, respectively, despite lower visual pass rate and PSNR.
IE and AFE each further reduce both errors. Their combination achieves the
lowest errors and best video metrics, reducing gated and raw errors by 9.57\%
and 21.68\% relative to IE, with a 0.71-percentage-point drop in visual pass.
Thus, both objectives contribute at this checkpoint, while visual pass does
not improve uniformly (Appendix~\ref{app:afe-ablations}).

\FloatBarrier

\section{Conclusion and Limitations}

In this work, we introduced \benchmark to evaluate action-conditioned world
models beyond the narrow distribution of expert demonstrations, jointly
measuring visual integrity and action-induced trajectory alignment. Our
evaluation shows that the six expert-trained models degrade under
feasible off-expert actions, exposing failures overlooked by expert-only
protocols. Guided by this diagnosis, we proposed \method, which combines
distributional coverage, representational grounding, and intervention-effect
alignment for faithful action-conditioned generation. \method reduced
integrity-gated error on RoboTwin and yielded larger policy gains than
expert-trained CtrlWorld in simulation and on real robots. Although \benchmark
substantially broadens evaluation coverage, comprehensively probing
long-horizon interactions across diverse embodiments and open-world
environments remains a shared challenge for the field and an important
direction for future work.

\label{page:main-text-end}

\ificlrarxiv
    \section*{Acknowledgements}
This work was supported by the National Natural Science Foundation of China
(62476011), the Beijing Natural Science Foundation (L252060), and the Beijing
Major Science and Technology Project under Contract no. Z191100010618003.

\else\ificlrfinal
    \section*{Acknowledgements}
This work was supported by the National Natural Science Foundation of China
(62476011), the Beijing Natural Science Foundation (L252060), and the Beijing
Major Science and Technology Project under Contract no. Z191100010618003.

\fi\fi

\clearpage
\subsection*{AI Use Statement}
Generative AI tools were used to assist with manuscript formatting and language
editing. All AI-assisted content was reviewed by the authors, who take
responsibility for the final content of this work.

\bibliography{references}
\bibliographystyle{iclr2027}

\clearpage
\appendix
\section*{Appendix}
\section{Real-Robot Configuration}
\label{app:real-robot-setup}
This appendix describes the real-robot platforms and manipulation tasks.
Section~\ref{app:robot-platforms} details the hardware, action spaces,
data collection, and camera views for each platform.
Section~\ref{app:real-robot-tasks} introduces the task instructions and
manipulation sequences, illustrated through third-person and robot-camera views.

\subsection{Robot Platforms}
\label{app:robot-platforms}

\subsubsection{Dual TianJi-Sharpa (Robot 1)}
\label{app:robot-tianji}

\begin{figure}[htbp]
    \centering
    \includegraphics[width=0.95\linewidth]{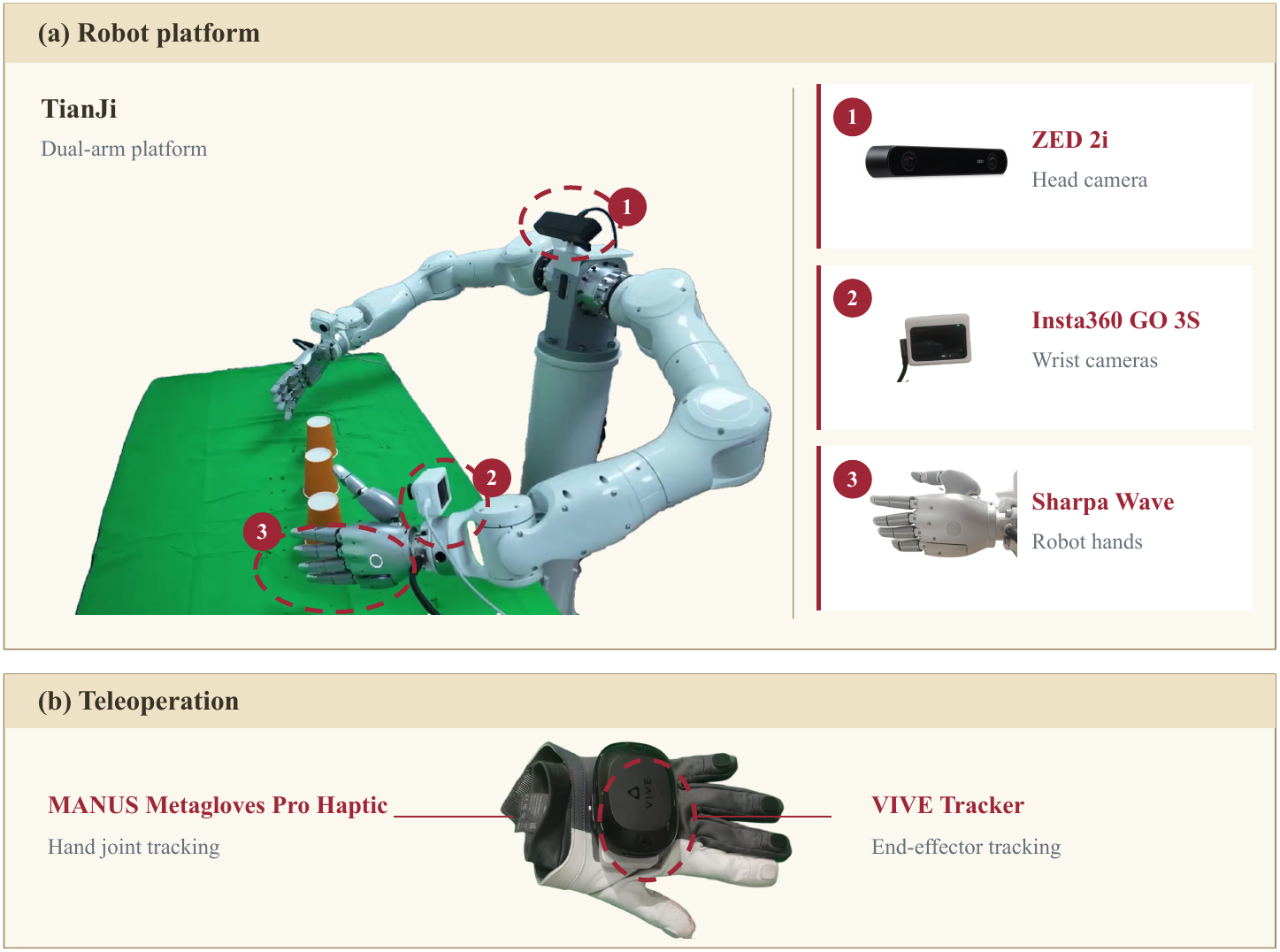}
    \caption{TianJi hardware and teleoperation setup. Numbered markers match
    the installed devices to the equipment list; the lower panel identifies
    the hand-motion and end-effector tracking devices.}
    \label{fig:real-robot-setup}
\end{figure}

\paragraph{Hardware.}
Our first platform uses two
\href{https://en.tianjizn.com/products/marvin-series/}{TianJi robotic arms},
each with seven degrees of freedom (DoF), equipped with
\href{https://www.sharpa.com/pages/wave}{Sharpa Wave} dexterous hands.
Each hand has 22 independently actuated DoF.
A \href{https://www.stereolabs.com/store/products/zed-2i}{ZED 2i}
head camera provides a workspace view, while
\href{https://www.insta360.com/cn/product/insta360-go3s}{Insta360 GO 3S}
cameras provide wrist views.
\href{https://www.vive.com/cn/accessory/vive-ultimate-tracker/}{VIVE Ultimate Trackers}
control the arm end-effector motion during teleoperation, and
\href{https://www.manus-meta.com/products/metagloves-pro-haptic}{MANUS Metagloves Pro Haptic}
provide the operator's hand motion for controlling the dexterous hands.
Figure~\ref{fig:real-robot-setup} identifies the platform hardware and
teleoperation devices.

\Needspace{8\baselineskip}
\paragraph{Action Space.}
We retain \method's relative Cartesian end-effector pose commands
(Delta EE) for each arm and extend its hand control to dexterous manipulation.
For each hand, the single gripper-opening component is replaced by
22 absolute joint-position targets. The combined command is
\begin{equation}
    a_t = \bigl(\Delta T_t^L, q_t^L, \Delta T_t^R, q_t^R\bigr),
    \qquad \Delta T_t^e \in \mathrm{SE}(3),\quad
    q_t^e \in \mathbb{R}^{22},\quad e\in\{L,R\}.
    \label{eq:real-robot-action}
\end{equation}
Here, $\Delta T_t^e$ specifies the commanded end-effector pose change and
$q_t^e$ specifies the corresponding hand-joint targets.

\paragraph{Data Collection.}
We collect 200 teleoperated demonstrations of Cup Stacking for the initial
baseline dataset, before iterative policy improvement.
Observations and robot commands are recorded at 30\,Hz and downsampled to
10\,Hz for training. The demonstrations pair camera observations with
end-effector motion commands and dexterous-hand joint targets.

\paragraph{Camera Views.}
The head view shows the tabletop and both hands, while the wrist views
show the nearby hand and objects during manipulation.
The RGB frames have a resolution of
$640\times360$ pixels and retain the full stored field of view.

\subsubsection{Simplexity i7 Pro (Robot 2)}
\label{app:robot-i7pro}

\begin{figure}[htbp]
    \centering
    \includegraphics[width=0.95\linewidth]{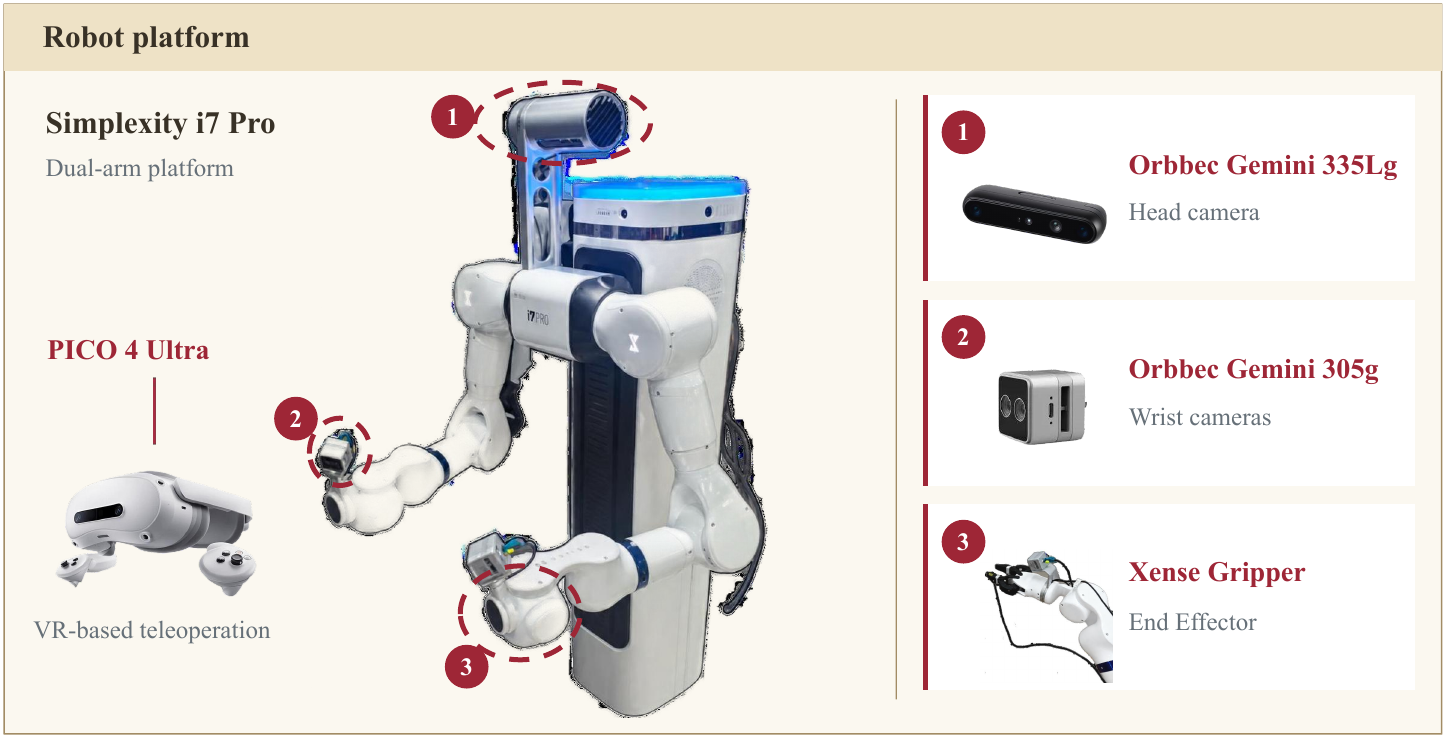}
    \caption{i7 Pro hardware and teleoperation setup. Numbered markers link
    the installed cameras and grippers to the equipment list; the left
    side shows the PICO teleoperation device.}
    \label{fig:i7pro-hardware}
\end{figure}

\paragraph{Hardware.}
Our second platform is a Simplexity i7 Pro dual-arm robot equipped with
\href{https://www.xenserobotics.com/product/35}{Xense grippers}. The base and torso remain fixed during the task.
An \href{https://store.orbbec.com/products/gemini-335lg}{Orbbec Gemini 335Lg} head camera provides a workspace view, and
\href{https://www.orbbec.com/gemini-305g/}{Orbbec Gemini 305g} cameras provide left and right wrist views.
Figure~\ref{fig:i7pro-hardware} identifies the installed devices and the
\href{https://www.picoxr.com/uk/products/pico4-ultra}{PICO 4 Ultra} teleoperation interface.

\Needspace{8\baselineskip}
\paragraph{Action Space.}
Following \method's shared end-effector action formulation, we control
both arms through relative Cartesian end-effector pose commands (Delta EE)
and a continuous gripper-opening command for each arm. The combined command is
\begin{equation}
    a_t = \bigl(\Delta T_t^L, g_t^L, \Delta T_t^R, g_t^R\bigr),
    \qquad \Delta T_t^e \in \mathrm{SE}(3),\quad
    g_t^e \in \mathbb{R},\quad e\in\{L,R\}.
    \label{eq:i7pro-action}
\end{equation}
Here, $\Delta T_t^e$ specifies the commanded end-effector pose change and
$g_t^e$ specifies the corresponding gripper opening.

\paragraph{Data Collection.}
We collect 200 demonstrations of Box Packing through PICO teleoperation.
Camera observations and robot commands are recorded at 30\,Hz and
downsampled to 10\,Hz for training.

\paragraph{Camera Views.}
The head camera observes the tabletop and both arms, while the wrist
cameras show the grippers and nearby objects during manipulation.
The three camera views are synchronized.

\subsubsection{Franka Research 3 (Robot 3)}
\label{app:robot-fr3}

\begin{figure}[htbp]
    \centering
    \includegraphics[width=0.95\linewidth]{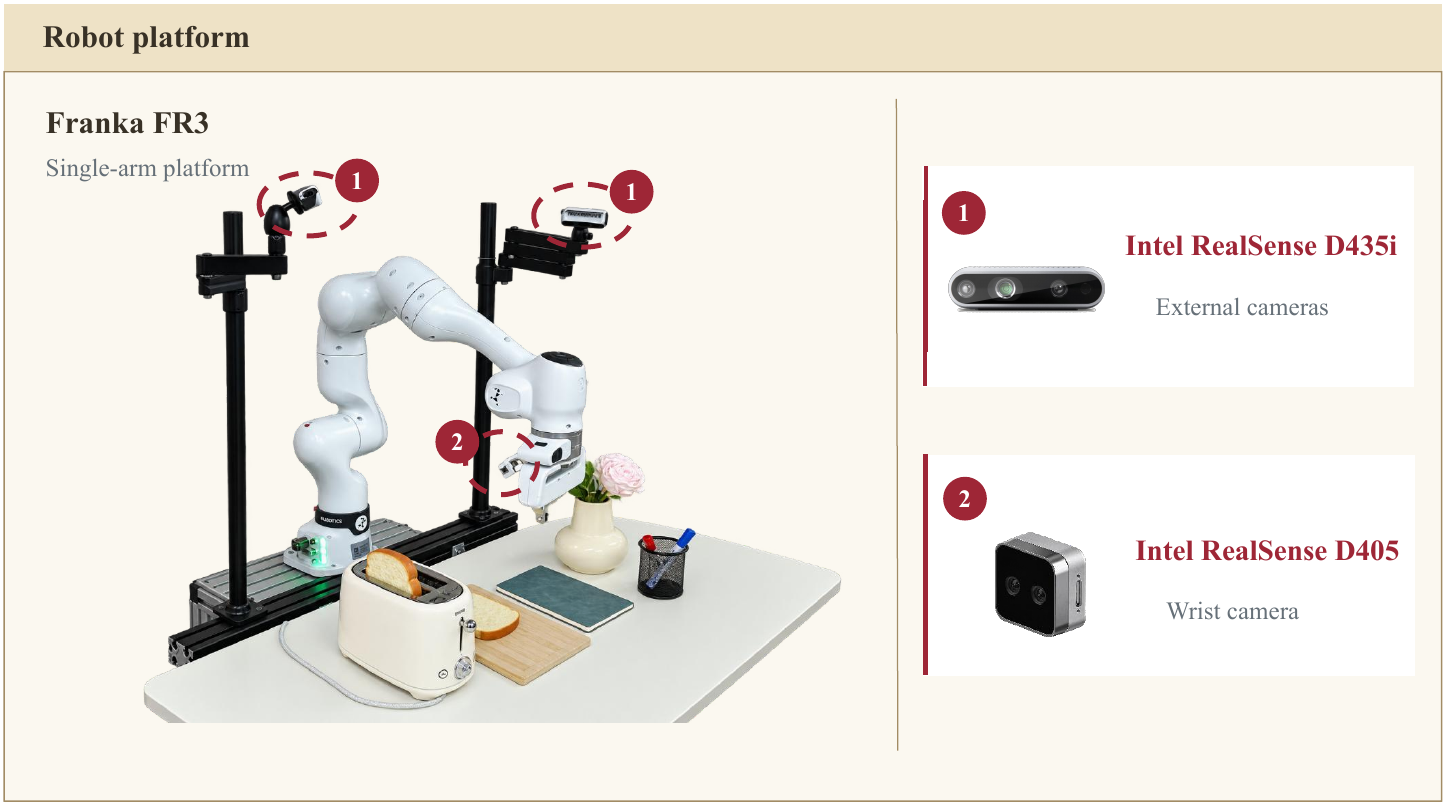}
    \caption{Franka FR3 hardware and camera setup. Numbered markers link the
    camera locations to the equipment list.}
    \label{fig:fr3-hardware}
\end{figure}

\paragraph{Hardware.}
Our third platform uses a \href{https://franka.de/franka-research-3}{Franka Research 3 (FR3)} robotic arm with seven
degrees of freedom (DoF), equipped with a parallel-jaw gripper. Two \href{https://store.realsenseai.com/buy-realsense-depth-camera-d435i.html}{Intel
RealSense D435i} cameras provide complementary
external views of the workspace. An \href{https://store.realsenseai.com/buy-realsense-depth-camera-d405.html}{Intel RealSense D405}
mounted near the end effector provides a close-range wrist view.
Figure~\ref{fig:fr3-hardware} identifies the platform and installed cameras.

\Needspace{8\baselineskip}
\paragraph{Action Space.}
We control the FR3 through relative Cartesian end-effector pose commands
(Delta EE) and a continuous gripper-opening command, following \method's
shared end-effector action formulation. The single-arm command is
\begin{equation}
    a_t = \bigl(\Delta T_t, g_t\bigr),
    \qquad \Delta T_t \in \mathrm{SE}(3),\quad g_t \in \mathbb{R}.
    \label{eq:fr3-action}
\end{equation}
Here, $\Delta T_t$ specifies the commanded end-effector pose change and
$g_t$ specifies gripper opening. Both FR3 tasks use this action representation.

\paragraph{Data Collection.}
We collect 300 demonstrations for Bread Placement and 500 demonstrations for
Pen Collection. The recorded trajectories are downsampled to 10\,Hz for training.
Each demonstration pairs the three camera observations with the corresponding
relative end-effector pose commands and continuous gripper-opening commands.

\paragraph{Camera Views.}
The two external cameras observe the workspace from complementary
viewpoints, while the wrist-mounted camera shows the gripper and nearby
objects during manipulation.

\clearpage
\begingroup
\setlength{\intextsep}{6pt}
\setlength{\abovecaptionskip}{6pt}
\subsection{Tasks}
\label{app:real-robot-tasks}

\subsubsection{Cup Stacking}
\label{app:task-stack-cups}

\paragraph{Instruction.}
``Stack the three upside-down cups into a single stack at the center of the table.''

\paragraph{Task Description.}
Three paper cups are initially placed upside down on the table in front
of the TianJi platform. The robot first uses its left hand to place the
left cup over the center cup. It then uses its right hand to place the
right cup over the resulting stack. The task ends with all three cups
nested in a single stack at the center position.

\paragraph{Task Visualization.}
Figures~\ref{fig:stack-cups-third-person} and~\ref{fig:stack-cups-robot}
illustrate Cup Stacking from third-person and robot-camera views, respectively.

\begin{figure}[!htbp]
    \centering
    \includegraphics[width=\linewidth]{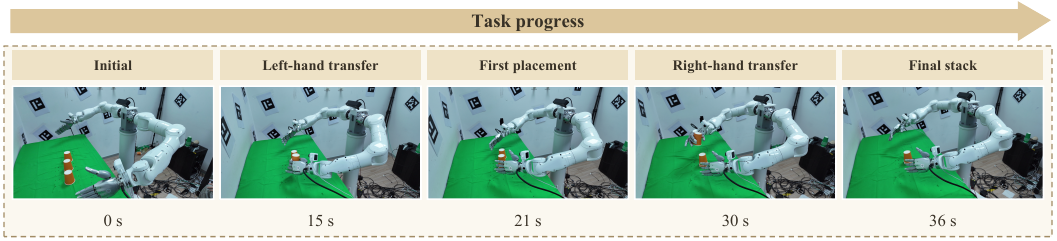}
    \caption{Cup Stacking from a third-person viewpoint. Read left to right;
    timestamps mark selected frames from one rollout.}
    \label{fig:stack-cups-third-person}
\end{figure}

\begin{figure}[!htbp]
    \centering
    \includegraphics[width=0.95\linewidth,height=0.47\textheight,keepaspectratio]{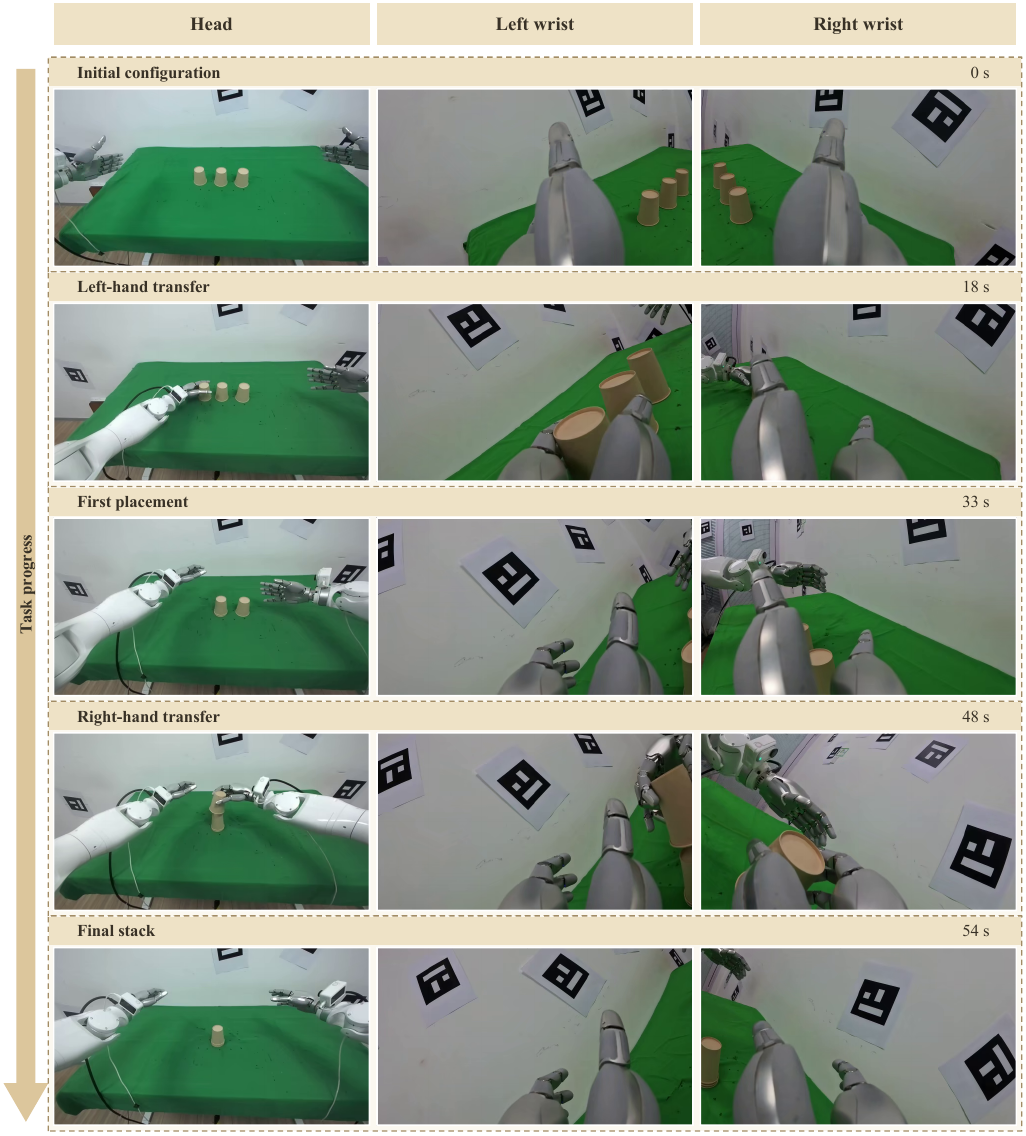}
    \caption{Cup Stacking through the robot's cameras. Columns show the head,
    left wrist, and right wrist; rows advance in time, with synchronized
    views within each row. This rollout is separate from
    Figure~\ref{fig:stack-cups-third-person}; timestamps are relative to each recording.}
    \label{fig:stack-cups-robot}
\end{figure}

\clearpage
\subsubsection{Box Packing}
\label{app:task-box-packing}

\paragraph{Instruction.}
``Place the mango in the box, close the flaps, and seal the box with tape.''

\paragraph{Task Description.}
A mango model, an open cardboard box, and a tape dispenser are initially
placed on the table in front of the i7 Pro platform. The robot places the
mango in the box, folds the top flaps closed, and uses the tape dispenser
to apply tape across the top seam. The task ends with the mango inside
the closed box and tape applied across the seam.

\paragraph{Task Visualization.}
Figures~\ref{fig:box-packing-third-person} and~\ref{fig:box-packing-robot}
illustrate Box Packing from third-person and robot-camera views, respectively.

\begin{figure}[!htbp]
    \centering
    \includegraphics[width=\linewidth]{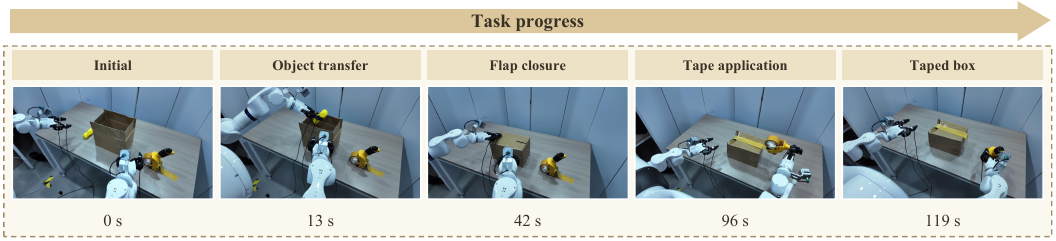}
    \caption{Box Packing from a third-person viewpoint. Read left to right;
    timestamps mark selected frames from one rollout.}
    \label{fig:box-packing-third-person}
\end{figure}

\begin{figure}[!htbp]
    \centering
    \includegraphics[width=0.95\linewidth,height=0.47\textheight,keepaspectratio]{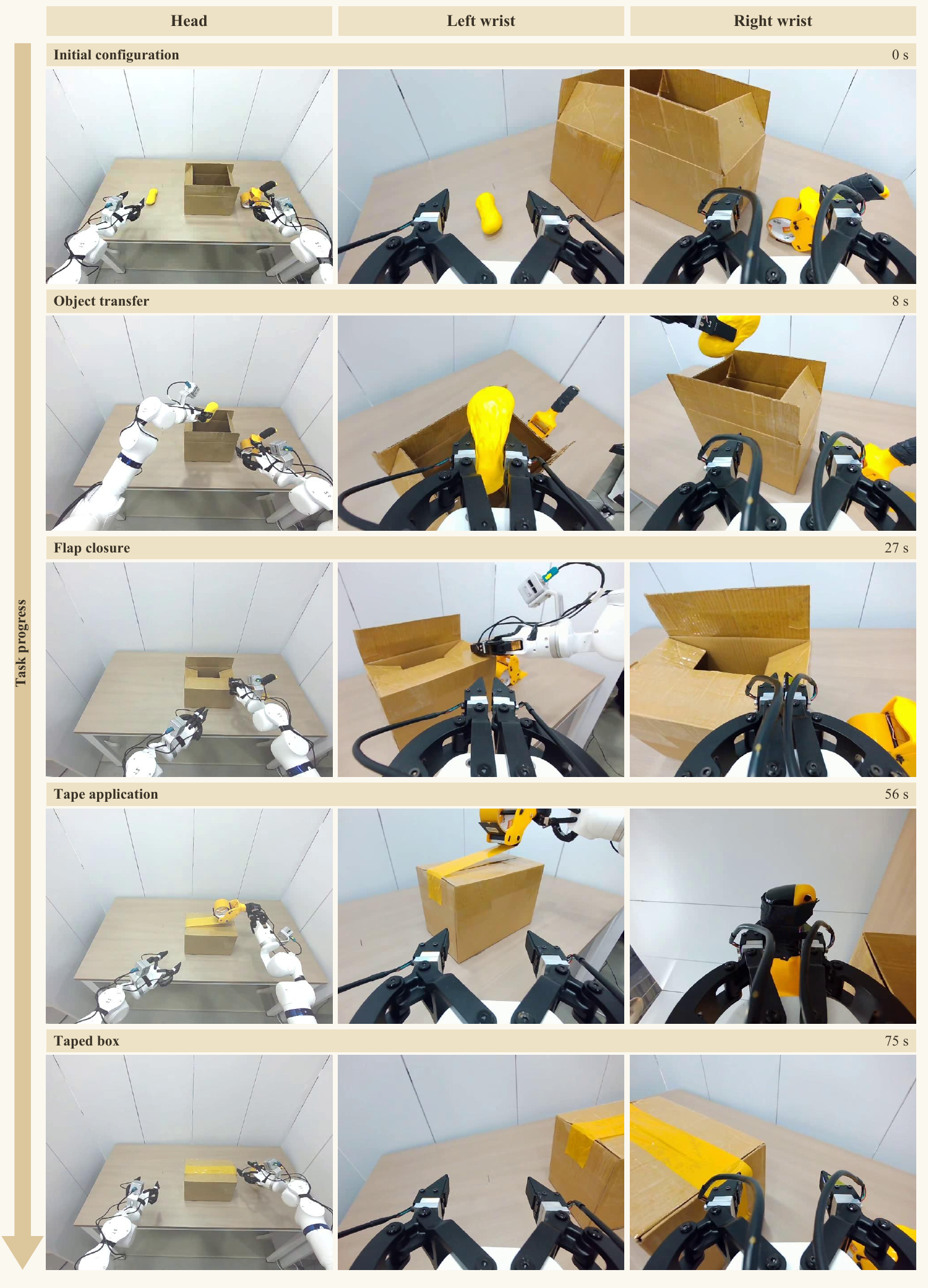}
    \caption{Box Packing through the robot's cameras. Columns show the head,
    left wrist, and right wrist; rows advance in time, with synchronized
    views within each row. This rollout is separate from
    Figure~\ref{fig:box-packing-third-person}; timestamps are relative to each recording.}
    \label{fig:box-packing-robot}
\end{figure}

\clearpage
\subsubsection{Bread Placement}
\label{app:task-bread-placement}

\paragraph{Instruction.}
The recorded task instruction is ``Take the bread out of the toaster and
place it on the cutting board.''

\paragraph{Task Description.}
Two slices of bread are initially placed in the toaster in front of the FR3
platform. The robot sequentially grasps each slice, removes it from the
toaster, transfers it to the cutting board, and releases it. The task ends
when both slices have been successfully placed on the cutting board.

\paragraph{Task Visualization.}
Figures~\ref{fig:bread-third-person} and~\ref{fig:bread-robot-progress}
illustrate Bread Placement from third-person and robot-camera views, respectively.

\begin{figure}[!htbp]
    \centering
    \includegraphics[width=\linewidth]{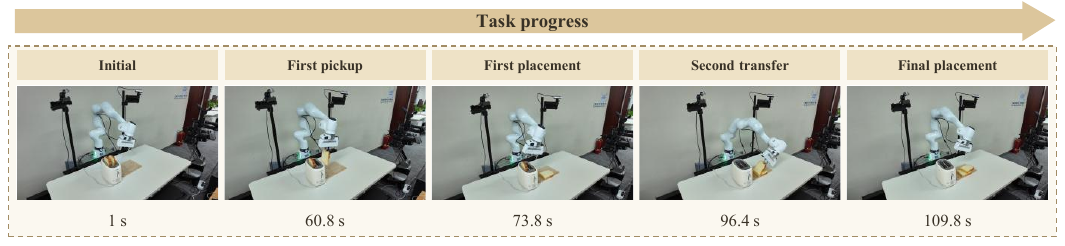}
    \caption{Bread Placement from a third-person viewpoint. Read left to right;
    timestamps mark selected frames from one rollout.}
    \label{fig:bread-third-person}
\end{figure}

\begin{figure}[!htbp]
    \centering
    \includegraphics[width=0.95\linewidth,height=0.47\textheight,keepaspectratio]{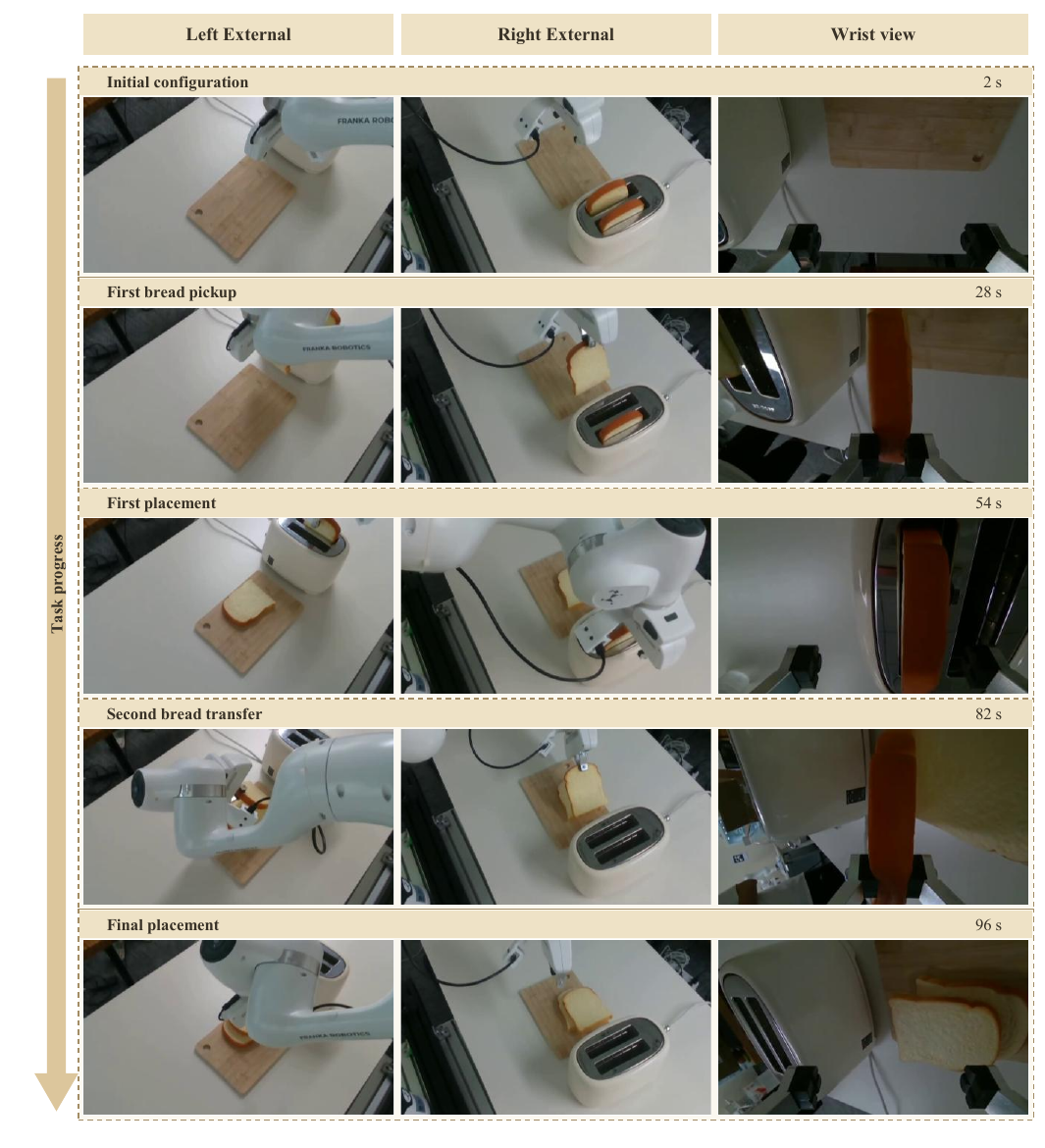}
    \caption{Bread Placement through the robot's cameras. Columns show the left
    external, right external, and wrist views; rows advance in time, with
    corresponding views within each row. This rollout is separate from
    Figure~\ref{fig:bread-third-person}; timestamps are relative to each recording.}
    \label{fig:bread-robot-progress}
\end{figure}

\clearpage
\subsubsection{Pen Collection}
\label{app:task-pickpen}

\paragraph{Instruction.}
The recorded task instruction is ``Pick up the pens one at a time and
place them into the pen holder.''

\paragraph{Task Description.}
Two pens are initially placed on the tabletop in front of the FR3 platform.
The robot grasps the pens one at a time, transfers each pen toward the pen
holder, and releases it into the holder. The task ends when both pens have
been successfully placed inside the pen holder.

\paragraph{Task Visualization.}
Figures~\ref{fig:pickpen-third-person} and~\ref{fig:pickpen-robot-progress}
illustrate Pen Collection from third-person and robot-camera views, respectively.

\begin{figure}[!htbp]
    \centering
    \includegraphics[width=\linewidth]{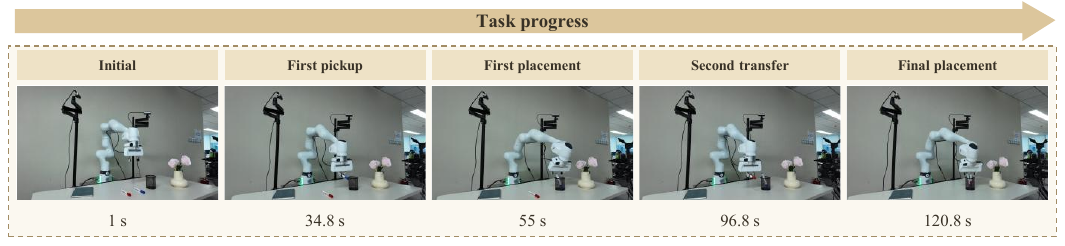}
    \caption{Pen Collection from a third-person viewpoint. Read left to right;
    timestamps mark selected frames from one rollout.}
    \label{fig:pickpen-third-person}
\end{figure}

\begin{figure}[!htbp]
    \centering
    \includegraphics[width=0.95\linewidth,height=0.47\textheight,keepaspectratio]{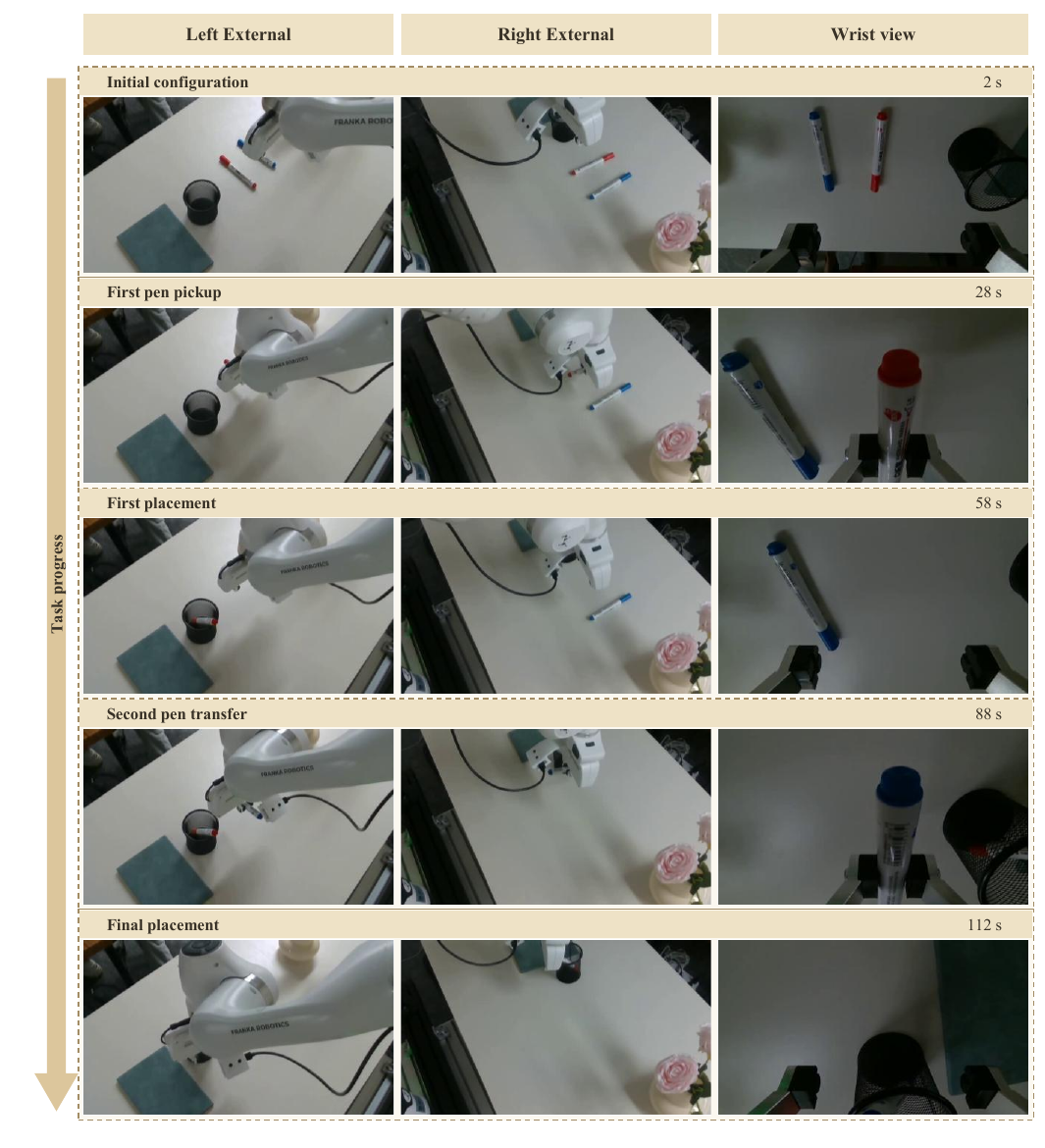}
    \caption{Pen Collection through the robot's cameras. Columns show the left
    external, right external, and wrist views; rows advance in time, with
    corresponding views within each row. This rollout is separate from
    Figure~\ref{fig:pickpen-third-person}; timestamps are relative to each recording.}
    \label{fig:pickpen-robot-progress}
\end{figure}

\FloatBarrier
\clearpage
\endgroup

\section{Robotic Pretraining}
\label{app:robotic-pretraining}

\paragraph{Data.}
The additional robotic pretraining variant of WorldSync uses AgiBot-World-Alpha~\citep{AgiBotWorld}
and DROID~\citep{DROID}. After cleaning, we retain 32,836 AgiBot episodes:
31,953 gripper episodes and 883 dexterous-hand episodes.
From DROID's cleaned subset of 92,998 episodes, we exclude 81 that are
shorter than the required 33-frame clip, leaving 92,917 episodes.
Table~\ref{tab:pretraining-data} summarizes the retained data.

\begin{table}[ht]
    \centering
    \small
    \setlength{\tabcolsep}{5pt}
    \caption{Robotic pretraining data after filtering. Steps count recorded
    timesteps, including episode endpoints, rather than optimizer updates or
    sampled clips; synchronized camera views are counted once. Hours give the
    corresponding video duration. Mean length is the number of recorded steps
    per episode, with the total-row mean computed over all retained episodes.}
    \label{tab:pretraining-data}
    \begin{tabular}{@{}lrrrr@{}}
        \toprule
        \textbf{Dataset} & \textbf{Episodes} & \textbf{Steps} & \shortstack{\textbf{Mean length}\\\textbf{(steps)}} & \textbf{Hours} \\
        \midrule
        AgiBot-World-Alpha & 32,836 & 55,255,078 & 1,682.76 & 511.62 \\
        DROID & 92,917 & 26,959,694 & 290.15 & 499.25 \\
        \midrule
        Total & 125,753 & 82,214,772 & 653.78 & 1,010.87 \\
        \bottomrule
    \end{tabular}
\end{table}

\paragraph{Action Space.}
Both datasets use a unified 20-dimensional action representation with ten
dimensions per arm: three for translation increments, six for relative
rotation, and one for gripper opening at the next timestep. Pose increments are computed
between adjacent observed states in the robot base frame. Given successive
end-effector rotations $R_t$ and $R_{t+1}$, we form the relative rotation
$\Delta R_t=R_{t+1}R_t^\top$. We use a 6D rotation representation
by concatenating the first two columns of $\Delta R_t$.
We apply min--max normalization only to translation. The six rotation
components are already bounded in $[-1,1]$; we retain their original values
to preserve the unit-length, orthogonal column structure without
dataset-dependent rescaling. Gripper opening ranges from zero (closed) to one
(open).

We adapt this representation to each dataset. DROID's single-arm actions use
the ten components allocated to the right arm. The unused left-arm components
encode zero translation, identity rotation, and a zero gripper value. For
AgiBot dexterous-hand episodes, we retain end-effector motion and set the
gripper components to zero without encoding individual finger motion.

\paragraph{Video and Sampling.}
Each training clip contains one current frame, 32 future frames, and 32
actions. All three views are resized to $320\times256$ pixels. AgiBot uses
head and wrist views, while DROID's two external cameras and wrist camera
are assigned to the three view inputs in that order. Language conditions use episode descriptions; the 18,863
retained DROID episodes without descriptions use an empty string.

Clips are sampled with a stride of 32, with a final overlapping clip covering
each episode's tail. All clips stay within episode boundaries and contain real
future frames. This yields 1,741,651 AgiBot clips and 884,670 DROID clips.
We shuffle and traverse these clips without replacement, giving each
dataset a sampling share proportional to its clip count.

\paragraph{Training Configuration.}
We initialize from the official Wan2.2-TI2V-5B checkpoint~\citep{Wan2.2} and train the video
diffusion transformer (DiT) and Action AdaLN modules described in
Appendix~\ref{app:action-conditioning}, using the future-video flow-matching objective,
keeping the VAE and text encoder frozen. AFE and IE are disabled during
pretraining. We train for one epoch on 16 NVIDIA A100 GPUs over 10 days, accumulating gradients
over four microbatches per optimizer update for a global batch size of 128.
Training uses bfloat16 precision, ZeRO stage 2, and gradient checkpointing.

We use AdamW with a learning rate of $10^{-5}$, 500 warmup steps, and cosine
decay to $5\times10^{-7}$. The optimizer uses
$(\beta_1,\beta_2)=(0.9,0.95)$, $\epsilon=10^{-8}$, weight decay $10^{-8}$,
and gradient clipping at a norm of 1.0. The random seed is 45. Model weights
are saved every 5,000 updates and at the end of pretraining.

\clearpage
\paragraph{Baseline Pretraining.}
\label{app:baseline-pretraining}
In Table~\ref{tab:main}, Emb. PT indicates training on curated robot or human
egocentric interaction data before our task-specific fine-tuning, with or
without explicit action labels. A checkmark includes such data in an earlier
backbone-training stage; a dash denotes generic video pretraining only, not
training from scratch. Our CtrlWorld baseline starts from its DROID-trained
checkpoint, whereas \method without additional robotic pretraining starts
from Wan2.2-TI2V-5B.
The other baseline initializations are Cosmos-Predict2.5-2B's action-conditioned
checkpoint, Cosmos3-Nano, DreamDojo-2B's GR-1 post-trained checkpoint, Motus's
stage-2 checkpoint, and LingBotVA's base checkpoint.

The reported corpora provide context for these initializations.
CtrlWorld~\citep{CtrlWorld} is trained on DROID with approximately 95k
trajectories across 564 scenes.
Cosmos-Predict2.5~\citep{CosmosPredict2.5} includes approximately 481k
robot-domain videos, obtained by summing the camera-view counts in its
robotics data table; these are not unique episodes or hours.
Cosmos3~\citep{Cosmos3} reports 61.3k hours for action mid-training across
egocentric motion, robotics, driving, and camera motion, including 41.3k
hours of egocentric motion and 5.36k hours of robotics.
DreamDojo~\citep{DreamDojo} reports 44,711 hours of egocentric human video
before robot post-training, while LingBotVA~\citep{LingBotVA} reports
approximately 16k hours of robot manipulation data.
Our additional robotic pretraining uses 1,010.87 hours from AgiBot-World-Alpha
and DROID (Table~\ref{tab:pretraining-data}), counting synchronized views once.
These figures describe the reported data stages, not cumulative training
exposure or compute. Differences in data domains, counting rules, supervision,
and training schedules preclude interpreting their ratios as controlled
data-efficiency comparisons.
\clearpage

\section{Policy Improvement Approach}
\label{app:policy-improvement-pipeline}

\begingroup
\raggedbottom

\makeatletter
\newenvironment{policytable}{%
    \begin{table}[htbp]
    \centering\small
    \setlength{\tabcolsep}{4pt}%
    \renewcommand{\arraystretch}{1.12}%
    \long\def\@makecaption##1##2{%
        {\normalsize\raggedright ##1: ##2\par}%
        \vskip6pt}%
}{\end{table}}
\makeatother

Following the diagnose--improve--validate workflow in the Introduction,
we evaluate world-model quality through downstream policy improvement.
While \benchmark directly measures visual integrity and action following,
policy success after post-training assesses the value of generated experience
for learning to complete a task. We build on VLAW~\citep{VLAW} to compare
world models through a common policy-improvement procedure.
\S\ref{app:rollout-diagnosis} examines failures in our reproduction and
explains the resulting changes to the pipeline.
\S\ref{app:policy-improvement-setup} describes the shared experimental protocol
and implementation, and \S\ref{app:policy-improvement-results} reports
rollout retention and policy success over two improvement rounds.

\subsection{Diagnosis and Adaptations}
\label{app:rollout-diagnosis}

\paragraph{Reproduction and diagnosis.}
VLAW~\citep{VLAW} alternates world-model adaptation and policy improvement
using online and imagined experience. In our reproduction with CtrlWorld
and Qwen-OFT on RoboTwin's \texttt{rotate\_qrcode} task, policy success
decreased from 59\% to 46\% and then 14\%
(Table~\ref{tab:vlaw-reproduction}). This decline motivates examining
whether imagined rollouts provide reliable observation--action supervision.
We use this reproduction to diagnose failures before comparing world models
under the adapted pipeline on the bin-dumping task below.

\begin{policytable}
    \caption{\textbf{Reproduction of the VLAW framework.}
    Policy success rates (\%) on \texttt{rotate\_qrcode}, measured over 100
    episodes at each stage. Both CtrlWorld and Qwen-OFT are updated each round.}
    \label{tab:vlaw-reproduction}
    \begin{tabular*}{\linewidth}{@{\extracolsep{\fill}}lrrr@{}}
        \toprule
        \textbf{Setting} & \textbf{Initial}
          & \textbf{Round 1} & \textbf{Round 2} \\
        \midrule
        CtrlWorld + Qwen-OFT & 59 & 46 & 14 \\
        \bottomrule
    \end{tabular*}
\end{policytable}

Two weaknesses undermine this supervision. First, generated rollouts may
fail to advance the task or depict apparent progress through visual artifacts,
such as distorted arms or disappearing grippers (Figure~\ref{fig:motivation}).
Second, visually plausible motion may not follow the conditioning actions.
Figure~\ref{fig:imagined-rollout-replay} shows a bin-dumping example:
the policy's actions fail to pick up the cup and empty its contents in the
simulator, yet the world model predicts successful lifting and pouring.
This mismatch is consistent with a bias toward successful expert behavior
when responding to off-expert actions.
This discrepancy motivates screening both predicted task progress and action
consistency before using imagined rollouts for policy training.

\begin{figure}[htbp]
    \centering
    \includegraphics[width=\textwidth]{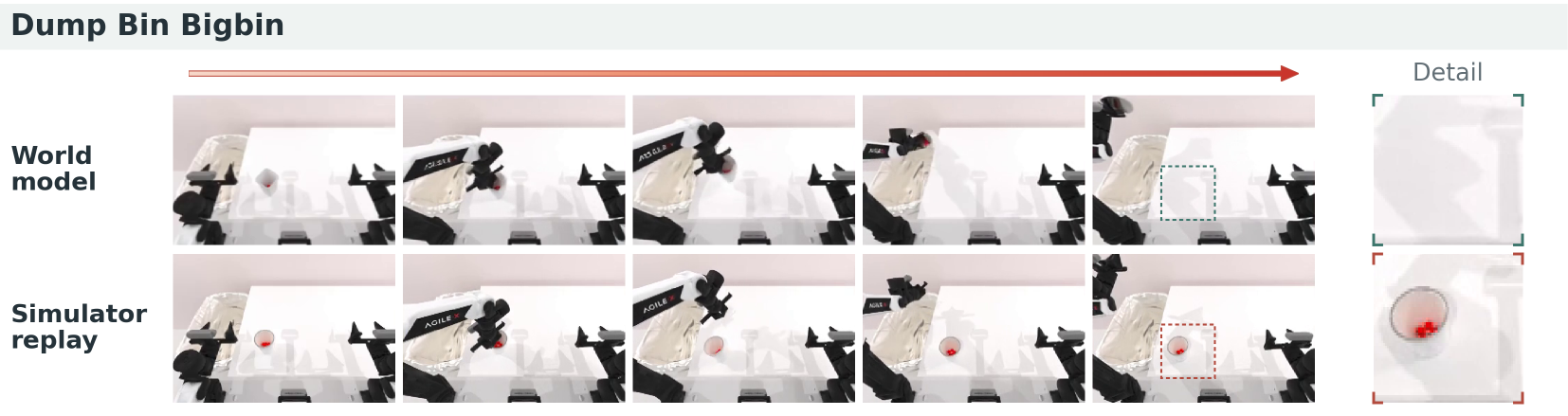}
    \caption{\textbf{Action mismatch under identical conditioning actions.}
    World-model prediction (top) and simulator replay (bottom) on
    \texttt{dump\_bin\_bigbin}, using the same actions from matched initial
    conditions. The prediction depicts lifting and pouring, whereas the
    simulator replay leaves the cup and its contents in place.
    The arrow indicates temporal order; rightmost panels magnify the boxed
    final-frame regions.}
    \label{fig:imagined-rollout-replay}
\end{figure}

\paragraph{Adaptations to VLAW.}
We retain VLAW's iterative organization of online collection, world-model
adaptation, closed-loop imagination, and supervised policy updates.
Within this organization, we refine how imagined experience is selected and
adapt the policy update to Qwen-OFT. We use Robo-Dopamine~\citep{RoboDopamine}
as the task-progress reward model $\mathcal R$ to remove rollouts with
insufficient predicted progress.
Because apparent progress can coexist with action mismatch, action-consistency
filtering then combines video--action rotation agreement with a paired-expert
reference feature, using AnyPos~\citep{AnyPos} as the pose extractor $\Phi$.
The first stage selects predicted progress, while the second checks the
associated motion. Neither score replaces the simulator's success label.

\begin{samepage}
Online world-model adaptation addresses the changing policy distribution.
It uses current-round successes and failures together with offline data,
while successful online rollouts accumulate for policy training.
For policy training, an L1 behavior-cloning objective combines those online
successes with retained imagined rollouts and expert demonstrations.
We apply the resulting pipeline to both world-model conditions under
the shared budgets specified in \S\ref{app:policy-improvement-setup}.
\par\end{samepage}

Imagined rollouts are regenerated each round, and both updated models
initialize the next round.
Algorithm~\ref{alg:policy-improvement} specifies the update sequence.
Here $\pi_k$ and $\mathcal W_{\theta_k}$ denote the policy and world model
before round $k+1$, and $\mathcal E$ is the simulator.

\begin{algorithm}[ht]
\caption{Policy Improvement with Filtered Imagined Rollouts}
\label{alg:policy-improvement}
\begin{algorithmic}[1]
\REQUIRE Initial models $(\pi_0,\mathcal{W}_{\theta_0})$, offline data
$\mathcal{D}_{\mathrm{off}}$, expert demonstrations $\mathcal{D}_{\mathrm{exp}}$,
environment $\mathcal{E}$, reward model $\mathcal{R}$, pose extractor $\Phi$,
fixed filter calibration and evaluation seeds $\mathcal{S}_{\mathrm{eval}}$
\STATE Initialize cumulative online successes $\mathcal{D}_{\mathrm{on}}^{+}\leftarrow\emptyset$
\STATE Evaluate $\pi_0$ on fixed evaluation seeds $\mathcal{S}_{\mathrm{eval}}$
\FOR{$k=0,1$}
    \item[] \textit{(a) Online collection and world-model adaptation}
    \STATE Collect online rollouts $\mathcal{D}_{\mathrm{on}}^k$ using $\pi_k$ in $\mathcal{E}$
    \STATE Accumulate simulator-verified successes in $\mathcal{D}_{\mathrm{on}}^{+}$
    \STATE Update $\mathcal{W}_{\theta_k}\to\mathcal{W}_{\theta_{k+1}}$
    using current $\mathcal{D}_{\mathrm{on}}^k$ and $\mathcal{D}_{\mathrm{off}}$ (ordinary sampling $1:1$),
    with the condition-specific objective and auxiliary data in \S\ref{app:policy-improvement-setup}
    \item[] \textit{(b) Imagined rollout generation}
    \STATE Generate imagined rollouts $\mathcal{D}_{\mathrm{syn}}^k$ from expert starts
    using $\pi_k$ and $\mathcal{W}_{\theta_{k+1}}$ in closed loop
    \item[] \textit{(c) Sequential filtering}
    \STATE Select $\mathcal{D}_{\mathcal{R}}^k\leftarrow
    \{z\in\mathcal{D}_{\mathrm{syn}}^k:s_{\mathcal{R}}(z)>0.8\}$
    \STATE Extract visual poses with $\Phi$ and reconstruct trajectories from conditioning actions
    \STATE Compute action-consistency scores $q_k(z)$ using Eq.~\eqref{eq:policy-consistency-filter}
    \STATE Select $\mathcal{D}_{\mathrm{keep}}^k\leftarrow
    \{z\in\mathcal{D}_{\mathcal{R}}^k:q_k(z)\leq\eta\}$
    \item[] \textit{(d) Policy update and evaluation}
    \STATE Construct policy-training examples from $\mathcal{D}_{\mathrm{keep}}^k$
    \STATE Fine-tune $\pi_k\to\pi_{k+1}$ on
    $\mathcal{D}_{\mathrm{on}}^{+}$, $\mathcal{D}_{\mathrm{keep}}^k$, and
    $\mathcal{D}_{\mathrm{exp}}$ (sampling $1:1:1$)
    \STATE Evaluate $\pi_{k+1}$ on $\mathcal{S}_{\mathrm{eval}}$
\ENDFOR
\RETURN $(\pi_2,\mathcal{W}_{\theta_2})$
\end{algorithmic}
\end{algorithm}
\FloatBarrier

\subsection{Experimental Setup}
\label{app:policy-improvement-setup}

\paragraph{Task, initialization, and controls.}
We detail the RoboTwin \texttt{dump\_bin\_bigbin} task, hereafter the
bin-dumping task, evaluated in Figure~\ref{fig:policy-improvement}.
All conditions start from the same Qwen-OFT policy checkpoint and their
respective world-model checkpoints.
We compare \method with CtrlWorld trained on expert demonstrations.
The policy initialization, expert demonstrations and starts, filter settings,
per-round budgets, and evaluation seeds are shared across conditions
(Table~\ref{tab:policy-improvement-protocol}). Online observations and retained
imagined rollouts may differ as the policies evolve, while the number of
policy optimizer updates remains fixed.
World-model architectures, training objectives, and offline data remain
condition-specific, as described below.

Each arm's action contains three translation increments, a 6D rotation
representation of the relative rotation, and one gripper value.
Together, the two arms give a 20-dimensional action
(Appendix~\ref{app:robotic-pretraining}).

\begin{policytable}
    \caption{\textbf{Shared simulation protocol.}
    Counts are per condition and per round unless stated otherwise.
    Sampling ratios do not denote unique rollout counts.}
    \label{tab:policy-improvement-protocol}
    \begin{tabular}{@{}>{\raggedright\arraybackslash}p{0.30\linewidth}
        >{\raggedright\arraybackslash}p{\dimexpr0.70\linewidth-2\tabcolsep\relax}@{}}
        \toprule
        \textbf{Setting} & \textbf{Configuration} \\
        \midrule
        Improvement rounds & 2 \\
        Online collection & 30 rollouts, including successes and failures \\
        Imagined rollouts & $3{,}000=50$ expert starts $\times60$ \\
        Action representation & 20-dimensional actions, 32 actions per chunk \\
        Generation horizon & 7 chunks (224 actions), 20 denoising steps per chunk \\
        World-model update & 1,000 updates, global batch 16, learning rate $10^{-5}$ \\
        World-model sampling & Current online : offline $=1:1$ for ordinary sampling \\
        Policy sampling & Cumulative online successes : current retained imagined : expert $=1:1:1$ \\
        Policy batch size & Global batch 32 \\
        Policy optimizer / loss & AdamW, learning rate $10^{-5}$, gradient clip 1, L1 behavior cloning \\
        Evaluation & Same ordered 100 seeds, initially and after each round \\
        \bottomrule
    \end{tabular}
\end{policytable}

\paragraph{Online collection and world-model adaptation.}
\label{app:policy-improvement-adaptation}
To follow the updated policy's state--action distribution, each round adapts
the world model using all 30 current online rollouts, including failures.
Offline replay preserves the original training distribution:
\method uses its original expert/off-expert mixture,
whereas CtrlWorld uses expert data.
\method retains video flow matching, AFE, and IE.
Its intervention-pair sampling fraction is 0.25, separate
from the ordinary online:offline mixture. CtrlWorld retains
its native video-denoising objective.

\paragraph{Sequential rollout filtering.}
\label{app:policy-improvement-selection}
To select useful experience without relying on apparent task completion alone,
we apply the complementary filters motivated in \S\ref{app:rollout-diagnosis}.
First, we use the Robo-Dopamine-GRM-2.0-8B-Preview checkpoint as $\mathcal R$
to score predicted task progress from the instruction, generated multi-view
observations, and paired expert terminal image.
It evaluates frame pairs at a nominal stride of eight, up to eight
pairs. For longer videos, it samples nine uniformly spaced frames including
the endpoints. With incremental scoring disabled, the rollout score
$s_{\mathcal R}(z)$ is the maximum over evaluated frames and forward/backward
scoring modes. Only rollouts with $s_{\mathcal R}(z)>0.8$ proceed to
action-consistency filtering.

For each such rollout $z$, $\Phi$ estimates both end-effector poses from the
generated head-camera video. We reconstruct action-conditioned trajectories
from the paired initial robot state using
$p_t=p_{t-1}+\Delta p_t$ and $R_t=\Delta R_tR_{t-1}$.
Both trajectories are expressed in physical units and aligned at video
timestamps. Their relative rotations, $\bar R_t=R_tR_0^\top$, define a
geodesic error $e_{\mathrm{rot}}(z)$ in degrees, averaged over both arms
and aligned frames. We normalize this error within the reward-passing set
$\mathcal D_{\mathcal R}^k$:
\begin{equation}
 r_k(z)=\frac{e_{\mathrm{rot}}(z)}
 {\operatorname{median}_{z'\in\mathcal D_{\mathcal R}^k}e_{\mathrm{rot}}(z')}.
 \label{eq:policy-rotation-feature}
\end{equation}

An expert-referenced feature provides additional information about the
action-conditioned motion. We select the arm with the longer expert Cartesian
path, choosing the left arm in a tie. For candidate and expert separately,
closure is the first frame of a four-frame window containing at least three
gripper values below 0.5, or the final frame if none exists.
The feature $d(z)$ is the Euclidean distance in meters between the candidate's
action-integrated position and the expert position at their respective closure
frames. It is a closure-position proxy, not a measured physical-contact event.
The joint retention rule is
\begin{equation}
 q_k(z)=\frac{r_k(z)-\mu_r}{s_r}-\lambda\frac{d(z)-\mu_d}{s_d},
 \qquad
 \mathcal D_{\mathrm{keep}}^k
 =\{z\in\mathcal D_{\mathcal R}^k:q_k(z)\leq\eta\}.
 \label{eq:policy-consistency-filter}
\end{equation}
Here $\mu$ and $s$ are calibration medians and interquartile ranges.
These statistics, the closure-feature weight $\lambda$, and the retention
threshold $\eta$ remain fixed across conditions and rounds.
Only the rotation batch median in
Eq.~\eqref{eq:policy-rotation-feature} is recomputed for each condition and round.
Because the implemented joint score assigns a negative coefficient to $d(z)$,
the rule does not impose separate upper bounds on both features.
Filtering stops if pose extraction is incomplete or feature statistics are invalid.

\FloatBarrier

\subsection{Results}
\label{app:policy-improvement-results}

\paragraph{Sequential rollout filtering.}
Table~\ref{tab:policy-improvement-funnel} tracks each round's 3,000 imagined
rollouts through both filters, reporting rollout counts and retention rates.

\begin{policytable}
    \caption{\textbf{Sequential filtering counts and retention rates.}
    Stage 1 applies task-progress filtering with $s_{\mathcal R}(z)>0.8$,
    followed by action-consistency filtering with $q_k(z)\leq\eta$ in Stage 2.
    Stage-wise rates use the preceding stage's count, whereas overall rates
    use all $3{,}000$ generated rollouts.}
    \label{tab:policy-improvement-funnel}
    \begingroup
    \setlength{\tabcolsep}{3pt}
    \begin{tabular*}{\linewidth}{@{\extracolsep{\fill}}
        >{\raggedright\arraybackslash}p{0.20\linewidth}
        >{\centering\arraybackslash}p{0.065\linewidth}
        *{6}{>{\centering\arraybackslash}p{0.095\linewidth}}@{}}
        \toprule
        & & \multicolumn{3}{c}{\textbf{Rollout counts}: $N_0\rightarrow N_1\rightarrow N_2$}
          & \multicolumn{3}{c}{\textbf{Retention (\%)}} \\
        \cmidrule(lr){3-5}\cmidrule(lr){6-8}
        \textbf{Method} & \textbf{Round}
          & \shortstack{\textbf{Initial}\\$N_0$}
          & \shortstack{\textbf{After S1}\\$N_1$}
          & \shortstack{\textbf{After S2}\\$N_2$}
          & \shortstack{\textbf{Stage 1}\\$N_1/N_0$}
          & \shortstack{\textbf{Stage 2}\\$N_2/N_1$}
          & \shortstack{\textbf{Overall}\\$N_2/N_0$} \\
        \midrule
        \method & 1 & 3,000 & 1,114 & 553 & 37.1 & 49.6 & 18.4 \\
        \method & 2 & 3,000 & 1,088 & 365 & 36.3 & 33.5 & 12.2 \\
        \addlinespace[3pt]
        CtrlWorld & 1 & 3,000 & 962 & 452 & 32.1 & 47.0 & 15.1 \\
        CtrlWorld & 2 & 3,000 & 909 & 323 & 30.3 & 35.5 & 10.8 \\
        \bottomrule
    \end{tabular*}
    \endgroup
\end{policytable}

Under the same generation budget and filtering criteria, \method retained
more imagined rollouts than CtrlWorld in both improvement rounds.
The final counts were 553 versus 452 in Round 1 and 365 versus 323 in
Round 2, corresponding to 22.3\% and 13.0\% more retained rollouts,
respectively. Thus, \method supplied more imagined rollouts that met both
the task-progress and action-consistency criteria for policy training from
the same number of generation attempts.
We next examine whether this higher data yield was accompanied by larger
policy gains under the fixed policy-training budget.

\paragraph{Downstream policy success.}
Table~\ref{tab:policy-improvement-success} gives the exact simulation values
plotted in Figure~\ref{fig:policy-improvement}.
Under the shared pipeline and budgets, \method achieved higher policy success
after both rounds than CtrlWorld. It reached 65\% after
two rounds, gaining 13 percentage points, whereas CtrlWorld gained
5 points and finished at 57\%.
The higher yield of accepted imagined rollouts and larger policy gains
together support higher world-model quality for downstream policy learning
under the evaluated protocol. Combined with the direct measurements
in \benchmark, these results provide complementary evidence for improved action
following and policy-learning utility.

\begin{policytable}
    \caption{\textbf{Policy success over two improvement rounds.}
    Success rates use the same ordered set of 100 evaluation seeds.
    Gains are percentage-point changes from each condition's measured initial rate.}
    \label{tab:policy-improvement-success}
    \begin{tabular*}{\linewidth}{@{\extracolsep{\fill}}lrrrr@{}}
        \toprule
        \textbf{Method} & \textbf{Initial (\%)} & \textbf{Round 1 (\%)}
          & \textbf{Round 2 (\%)} & \textbf{Gain (pp)} \\
        \midrule
        \method & 52 & 57 & \textbf{65} & \textbf{+13} \\
        CtrlWorld & 52 & 54 & 57 & +5 \\
        \bottomrule
    \end{tabular*}
\end{policytable}

\FloatBarrier
\par\endgroup

\section{Analysis of Action-Conditioning Mechanisms}
\label{app:action-conditioning}
\subsection{Action-Conditioning Mechanisms}

\begin{figure}[t]
\centering
\includegraphics[width=\textwidth,height=0.82\textheight,keepaspectratio]{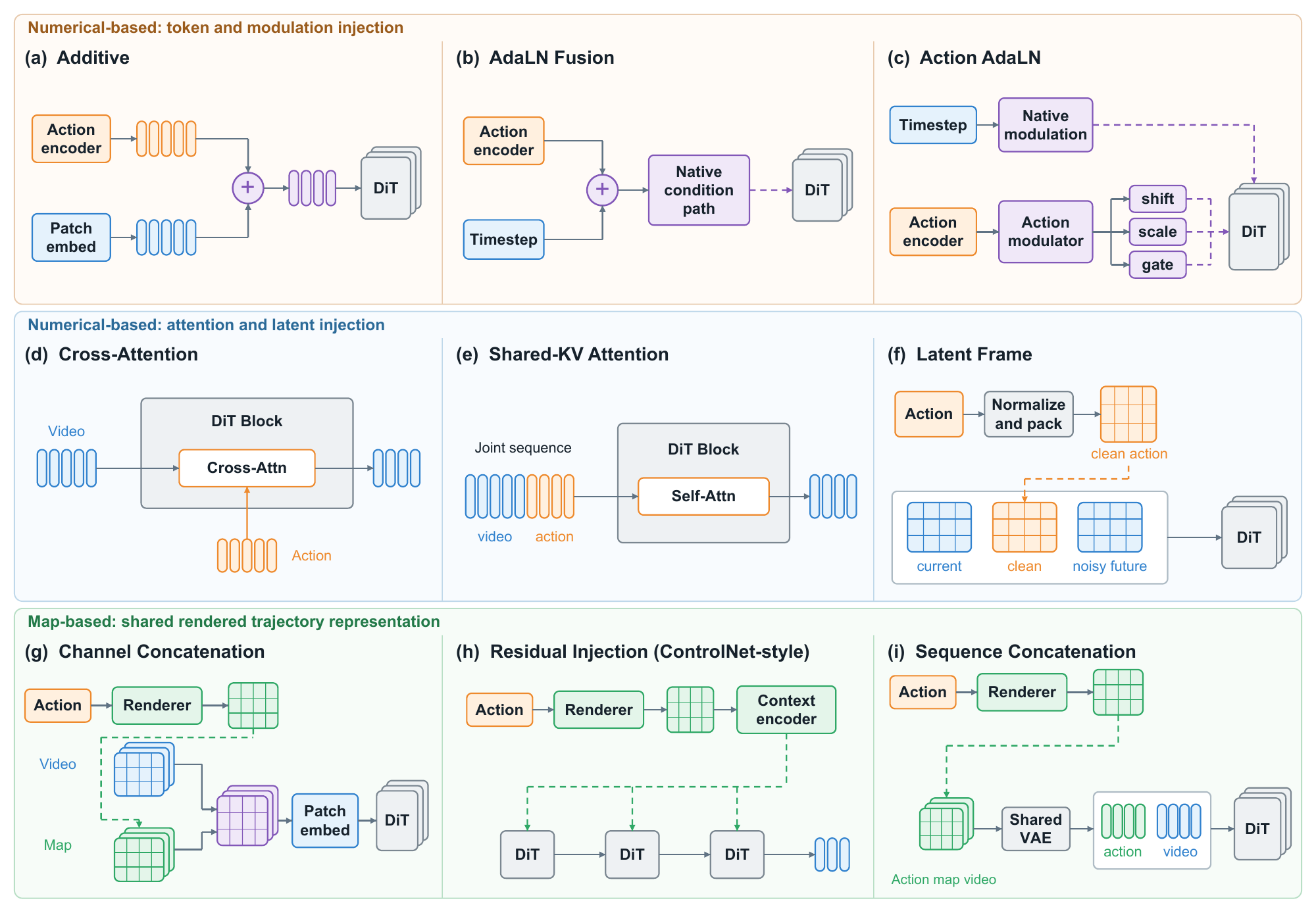}
\caption{Overview of the evaluated action-conditioning mechanisms. Numerical-based methods use feature addition, modulation, attention, or latent frames, whereas map-based methods use rendered action maps. Blue, orange, green, purple, and gray denote video features, numerical actions, action maps, fusion operations, and DiT blocks, respectively.}
\label{fig:injection-mechanisms}
\end{figure}

Action conditioning is central to robot world models because the model must
translate commanded robot motion into its visual consequences. For
video-DiT-based world models, this requirement raises two coupled design
questions of how actions should be represented and where they should interact
with video features. We therefore systematically compare nine mechanisms
selected from representative works and reimplemented within the same backbone,
organizing them along four design axes of action representation, temporal or
spatial alignment, injection site, and fusion.
Table~\ref{tab:injection-definitions} defines these mechanisms, while
Figure~\ref{fig:injection-mechanisms} illustrates their corresponding data
flows through the DiT backbone.

\begin{table}[t]
\centering
\caption{Action-conditioning mechanisms and representative works. All variants use the same video diffusion backbone.}
\label{tab:injection-definitions}
\small
\setlength{\tabcolsep}{2pt}
\renewcommand{\arraystretch}{1.22}
\begin{tabular}{@{}>{\raggedright\arraybackslash}p{0.17\textwidth}
                    >{\raggedright\arraybackslash}p{0.18\textwidth}
                    >{\raggedright\arraybackslash}p{0.15\textwidth}
                    >{\raggedright\arraybackslash}p{0.18\textwidth}
                    >{\raggedright\arraybackslash}p{0.24\textwidth}@{}}
\toprule
\textbf{Mechanism} & \textbf{Representative work} & \textbf{Alignment} & \textbf{Injection site} & \textbf{Fusion} \\
\midrule
\multicolumn{5}{@{}l}{\textbf{Numerical-based}}\\
\addlinespace[2pt]
Additive & NanoWM~\citeyearpar{NanoWM} & Frame-aligned & Patch features & Element-wise addition \\
AdaLN Fusion & Cosmos-Predict2.5~\citeyearpar{CosmosPredict2.5} & Frame-aligned & Native AdaLN path & Native-condition fusion \\
\rowcolor{methodband}
\tablemethod{Action AdaLN} & IRASim~\citeyearpar{IRASim} & Frame-aligned & Per-block attention/FFN & Low-rank shift/scale/gate \\
Cross-Attention & CtrlWorld~\citeyearpar{CtrlWorld} & Frame-aligned, local & Native attention & Gated action attention \\
Shared-KV Attention & Motus~\citeyearpar{Motus} & Global sequence & Native self-attention & Shared video/action K/V; video Q \\
Latent Frame & Cosmos Policy~\citeyearpar{CosmosPolicy} & Global chunk & Latent condition slot & Native attention across slots \\
\addlinespace[3pt]
\midrule
\multicolumn{5}{@{}l}{\textbf{Map-based} (shared RGB trajectory maps)}\\
\addlinespace[2pt]
Channel Concatenation & GE-Sim 2.0~\citeyearpar{GESim2.0} & Latent-aligned & Patch input & Channel concatenation \\
Residual Injection\newline{\footnotesize (ControlNet-style)} & ABot-PhysWorld~\citeyearpar{ABotPhysWorld} & Latent-aligned & Selected DiT blocks & Zero-initialized residuals \\
Sequence Concatenation & Action Images~\citeyearpar{ActionImages} & Full map video & Action/video sequence & Temporal concatenation; video-only loss \\
\bottomrule
\end{tabular}
\end{table}

Six numerical-based mechanisms retain actions as low-dimensional signals.
Additive follows NanoWM~\citep{NanoWM} and adds frame-aligned
action features directly to the patch features. AdaLN Fusion keeps the
same frame alignment but routes actions through the native AdaLN path,
where the action embedding is fused into the timestep pathway that drives the
backbone's adaptive layer normalization, as in
Cosmos-Predict2.5~\citep{CosmosPredict2.5} and DreamDojo~\citep{DreamDojo}. Action AdaLN then moves this
modulation from this shared conditioning path into each block; following
IRASim~\citep{IRASim}, it produces low-rank shift, scale, and gate offsets for
the attention and feed-forward sublayers. The next two mechanisms instead
introduce actions through attention. Cross-Attention follows
CtrlWorld~\citep{CtrlWorld} and lets video tokens query a separate action
branch through a gated attention output. Shared-KV Attention follows
Motus~\citep{Motus} and merges video and action keys and values in the native
self-attention while retaining video-only queries. DreamZero~\citep{DreamZero}
shares this joint video--action attention design, but further assigns action
tokens their own queries and prediction targets. Finally, Latent Frame follows
Cosmos Policy~\citep{CosmosPolicy} and reshapes a global action chunk to fill a
dedicated latent-frame-sized slot in the native DiT sequence. Together, these representative mechanisms span feature
addition, adaptive modulation, attention-based fusion, and latent-frame
conditioning.

\FloatBarrier

\begin{figure}[t]
\centering
\includegraphics[width=\textwidth]{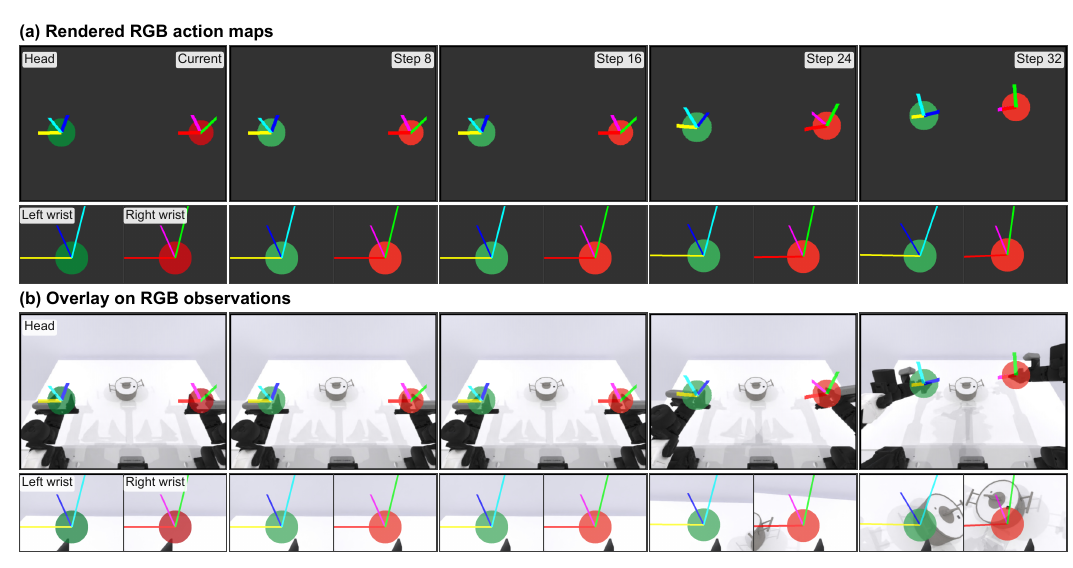}
\caption{RGB action-map representation. (a) Rendered action maps at the current frame and four future steps. (b) Corresponding overlays on the RGB observations. Each column shows the head, left-wrist, and right-wrist views; colored markers encode the projected end-effector poses.}
\label{fig:rgb-action-map-representation}
\end{figure}

However, compact numerical signals do not explicitly indicate where the
commanded motion should appear in each camera view. Some studies instead
project robot motion into image-aligned maps, providing a direct spatial
correspondence between actions and predicted visual changes. We therefore
evaluate three map-based mechanisms that share the RGB trajectory
representation used by GE-Sim 2.0~\citep{GESim2.0}. As shown in
Figure~\ref{fig:rgb-action-map-representation}, we project the end-effector
poses into each camera view and render trajectories over the prediction
horizon. Channel Concatenation follows GE-Sim 2.0 and injects these maps at the
patch input. Residual Injection adapts the ControlNet-style pathway
of ABot-PhysWorld~\citep{ABotPhysWorld}, adding map features to selected
backbone blocks through zero-initialized projections. Sequence Concatenation
follows Action Images~\citep{ActionImages} and places action-map and video
latents in a joint temporal sequence. By making spatial alignment explicit,
these mechanisms aim to give the model a more direct cue for where
action-induced changes should appear. Holding the action-map representation
fixed further isolates the effect of their injection paths.

\subsection{Experimental Setup}
All nine mechanisms are implemented on the same video diffusion backbone,
Wan2.2-TI2V-5B~\citep{Wan2.2}, a 5B-parameter video DiT initialized from the
official checkpoint. All variants are trained with the same flow-matching
objective on the same simulated trajectory mixture, which spans
demonstrated and off-expert actions. Each model generates 32 future frames
across three camera views from the current observation, conditioned on a
32-step action chunk in the shared end-effector
action space used by \method. Only the representation and injection of the
action condition differ across variants. We save checkpoints every 10k training
steps up to 40k; the fixed 40k checkpoint supports the primary comparison in
Table~\ref{tab:injection-results}, and Figure~\ref{fig:injection-curves}
examines the full checkpoint series.

All variants are evaluated following the \benchmark protocol (Algorithm~\ref{alg:benchmark}),
generating each rollout with 20 denoising steps. Each variant is evaluated on
four RoboTwin tasks and all five action-query categories. We
report raw pose-aware NDTW, integrity-gated error, and visual pass rate. Raw
NDTW measures the agreement between the generated and ground-truth end-effector
trajectories, the integrity-gated error assigns a fixed penalty $\kappa{=}0.3$
to visually invalid rollouts, and the visual pass rate isolates the frequency
of valid rollouts; we interpret the three jointly rather than as a single
ranking. Table~\ref{tab:injection-results} additionally
reports the trainable parameters added by each mechanism and the additional
inference compute relative to unconditioned video generation, estimated at the
operator level over 20 denoising steps. These performance and cost dimensions
are interpreted separately.

\begin{table}[t]
\centering
\caption{Comparison at the last training checkpoint. Params and FLOPs denote additional trainable parameters and inference compute relative to unconditioned video generation. Best performance is bold.}
\label{tab:injection-results}
\small
\setlength{\tabcolsep}{3.5pt}
\renewcommand{\arraystretch}{1.2}
\begin{tabular}{@{}>{\raggedright\arraybackslash}p{0.25\textwidth}rrrrr@{}}
\toprule
\textbf{Mechanism} & \textbf{Raw NDTW $\downarrow$} &
\textbf{Gated $\downarrow$} & \textbf{Pass (\%) $\uparrow$} &
\textbf{Params (M)} & \textbf{FLOPs (T)} \\
\midrule
Additive & 0.0320 & 0.0403 & 96.43 & 568.07 & 0.04 \\
AdaLN Fusion & 0.0330 & 0.0389 & 97.86 & 568.07 & 4.93 \\
\rowcolor{methodband}
\tablemethod{Action AdaLN} & \textbf{0.0137} & \textbf{0.0176} & \textbf{98.57} & 572.01 & 0.09 \\
Cross-Attention & 0.0302 & 0.0354 & 97.86 & 566.49 & 0.82 \\
Shared-KV Attention & 0.0332 & 0.0370 & \textbf{98.57} & 566.49 & 1.27 \\
Latent Frame & 0.0239 & 0.0297 & 97.86 & 0 & 69.39 \\
\midrule
Channel Concatenation & 0.0342 & 0.0379 & \textbf{98.57} & 0.04 & $<0.01$ \\
Residual Injection (ControlNet-style) & 0.0338 & 0.0413 & 97.14 & 88.17 & 7.61 \\
Sequence Concatenation & 0.0318 & 0.0392 & 97.14 & 0 & 520.90 \\
\bottomrule
\end{tabular}
\end{table}

\subsection{Comparison Results}
Table~\ref{tab:injection-results} reports the common 40k checkpoint under
the WorldEcho evaluation protocol. Action AdaLN attains the lowest raw NDTW (0.0137) and
integrity-gated error (0.0176), and shares the highest visual pass rate
(98.57\%) with Shared-KV Attention and Channel Concatenation. Latent Frame
has the second-lowest raw NDTW (0.0239) and gated error (0.0297), with a
97.86\% pass rate. The remaining seven mechanisms have raw NDTW values of
0.0302--0.0342. Visual pass rates are similar across mechanisms,
while substantial differences in trajectory agreement remain.

Integrity-gated error combines trajectory agreement with a fixed penalty
$\kappa{=}0.3$ for visually rejected clips. At 40k, pass rates range from
96.43\% to 98.57\%; the remaining failures still increase gated error above
raw NDTW. Channel Concatenation has the lowest gated error among map-based
methods (0.0379), despite having their highest raw NDTW (0.0342).
We therefore read raw NDTW and pass rate separately and gated error as their
synthesis. In this comparison, the three map-based mechanisms have higher raw
NDTW than Action AdaLN despite their explicit image-aligned spatial
correspondence. Their raw NDTW values are close to those of Additive,
AdaLN Fusion, Cross-Attention, and Shared-KV Attention.

\begin{figure}[t]
\centering
\includegraphics[width=\textwidth]{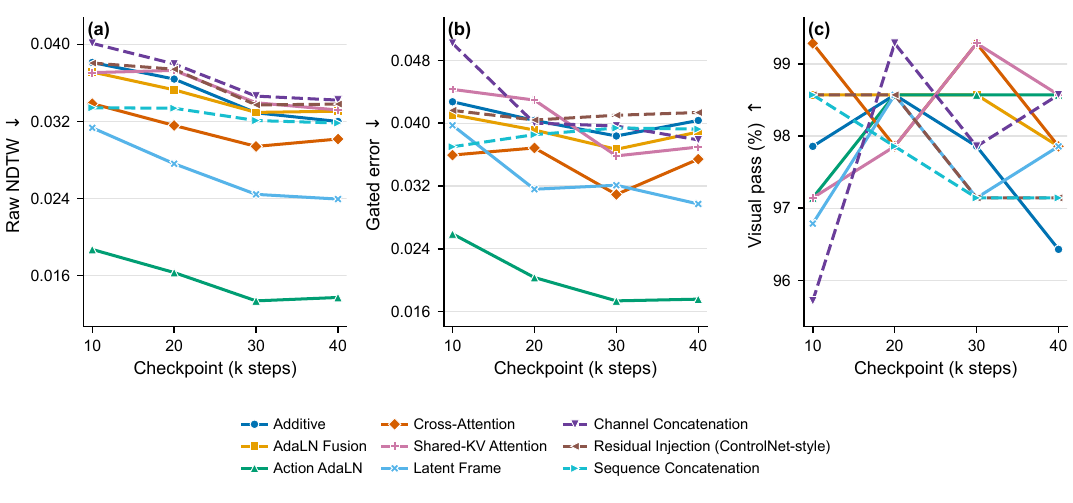}
\caption{Performance across training checkpoints under the WorldEcho evaluation protocol. The panels report raw NDTW, integrity-gated error, and visual pass rate for all nine mechanisms. Raw and gated errors are task-macro averages; pass rate counts samples. Solid and dashed lines denote numerical-based and map-based mechanisms, respectively. Latent Frame uses equal-weight aggregation of its component results.}
\label{fig:injection-curves}
\end{figure}

Figure~\ref{fig:injection-curves} shows that these relationships are stable
across training. Action AdaLN attains the lowest raw NDTW and gated error at
every displayed checkpoint, although its errors bottom out near 30k and rise
slightly at 40k, so its advantage does not grow monotonically with longer
training. Latent Frame's raw NDTW decreases across the displayed checkpoints,
while its gated error is not strictly monotonic. Visual pass rates across all
36 points range from 95.71\% to 99.29\%, and no single mechanism leads this
metric at every checkpoint. We therefore interpret the
curves as descriptive evidence of trend stability and trade-offs, without
claiming statistical significance from them.

\subsection{Computational Cost}
The nine mechanisms introduce overhead through three distinct profiles
(Table~\ref{tab:injection-results}). The additive, modulation, and attention
variants are parameter-dominated, with Additive, AdaLN Fusion, Action AdaLN,
Cross-Attention, and Shared-KV Attention each adding 566.49--572.01M trainable
parameters but little additional inference compute (0.04--4.93 TFLOPs). The
sequence-extending variants are compute-dominated. Latent Frame and Sequence
Concatenation add no trainable parameters, yet their additional latent tokens
are processed by the full backbone at every denoising step, costing 69.39 and
520.90 TFLOPs, respectively. Channel Concatenation is nearly free, with 0.04M
parameters and less than 0.01 TFLOPs, but has higher trajectory error than
Action AdaLN and Latent Frame.
Parameter count and inference compute are therefore decoupled, since a
mechanism can be parameter-free yet expensive per rollout or parameter-heavy
yet cheap to run. Neither number alone summarizes the cost of a design.

\begin{figure}[t]
\centering
\includegraphics[width=\textwidth]{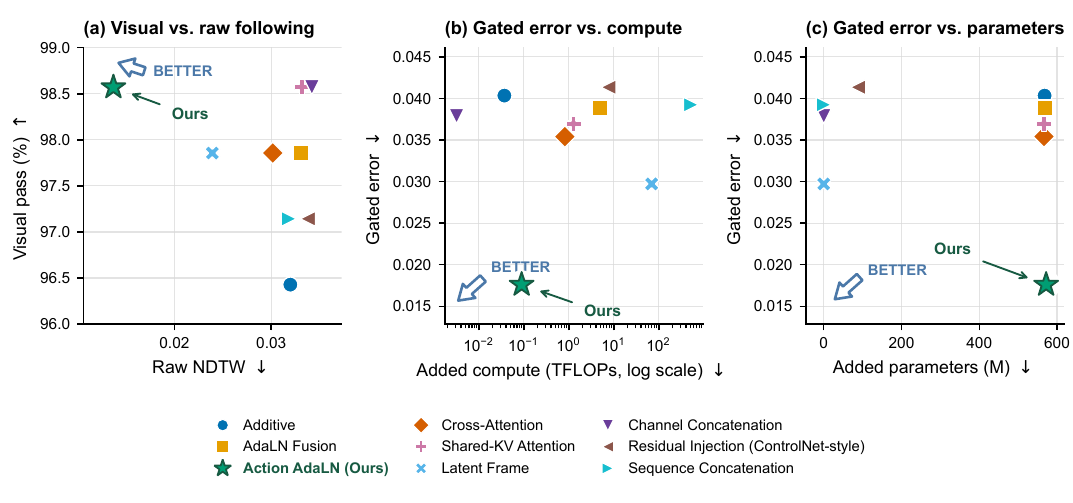}
\caption{Performance--efficiency trade-offs at the common 40k checkpoint under the WorldEcho evaluation protocol. (a) Visual pass rate versus raw NDTW. (b) Integrity-gated error versus additional inference compute on a logarithmic scale. (c) Integrity-gated error versus additional parameters. Arrows indicate preferred directions, and the star marks Action AdaLN (Ours).}
\label{fig:injection-tradeoffs}
\end{figure}

Figure~\ref{fig:injection-tradeoffs} relates these costs to performance. In
panel (a), Action AdaLN has the lowest raw NDTW and ties for the highest
visual pass rate; Latent Frame has the second-lowest raw NDTW.
Panels (b) and (c) show that no mechanism is simultaneously the cheapest and
the most accurate. Sequence Concatenation pays the highest additional compute
for a gated error of 0.0392, the parameter-free variants are not
automatically stronger, and Action AdaLN pairs the lowest gated error with
only 0.09 TFLOPs of additional compute although it is not the most
parameter-efficient design. We note that the reported FLOPs are operator-level
estimates over 20 denoising steps and do not directly measure wall-clock
latency.

\subsection{Discussion and Design Choice}
Two conclusions emerge from this study. First, no action-conditioning
mechanism resolves action following on its own. Across feature addition,
adaptive modulation, attention-based fusion, latent-frame conditioning, and
rendered action maps, every variant still trades trajectory agreement against
visual integrity to some degree, and every variant adds parameters, inference
compute, or both. Improving the injection pathway shifts the operating point
but does not close the gap. Our full model therefore pairs its injection
mechanism with broader action coverage and auxiliary training objectives
rather than relying on injection design alone.

Second, Action AdaLN achieves the lowest errors with a simple modulation
design. It modulates the attention and feed-forward sublayers of each DiT block
through low-rank shift, scale, and gate offsets, without extra tokens, a
separate attention branch, or rendered maps. Yet this standard modulation
design attains the lowest raw NDTW and integrity-gated error in
Table~\ref{tab:injection-results}, retains a 98.57\% visual pass rate, and
adds only 0.09 TFLOPs of inference compute. The more elaborate alternatives,
including extra latent slots, joint action--video sequences, ControlNet-style
residual branches, and explicit spatial maps, fail to surpass this strong
baseline on either error metric.

We therefore adopt Action AdaLN in our full model. This choice should not be
read as uniform superiority: Shared-KV Attention and Channel Concatenation
match its visual pass rate, while Latent Frame adds no trainable parameters.
What Action AdaLN offers is
the most favorable balance of accurate action following, visual integrity,
inference compute, and architectural simplicity among the nine candidates,
which is the combination that matters for a general-purpose robot world
model.

\section{Additional Ablation Studies}
\label{app:additional-ablations}

\subsection{Action-Forcing Expert Ablations}
\label{app:afe-ablations}

\paragraph{Motivation and design space.}
Intervention-Effect (IE) supervision compares generated futures under paired
actions. AFE adds a complementary constraint within each rollout: intermediate
video features should retain the robot-state changes induced by the actions.
This training-only objective raises three design questions. \emph{Where to
read?} Features at different backbone depths may differ in their usefulness
for state prediction. \emph{When to read?} Restricting a future-step query to
its corresponding video time may focus supervision but remove useful context.
\emph{What to predict?} Absolute state, stepwise change, and
initial-relative cumulative change emphasize different aspects of robot motion.
We compare these choices to select the AFE supervision.

\begin{figure}[t]
    \centering
    \includegraphics[width=\textwidth]{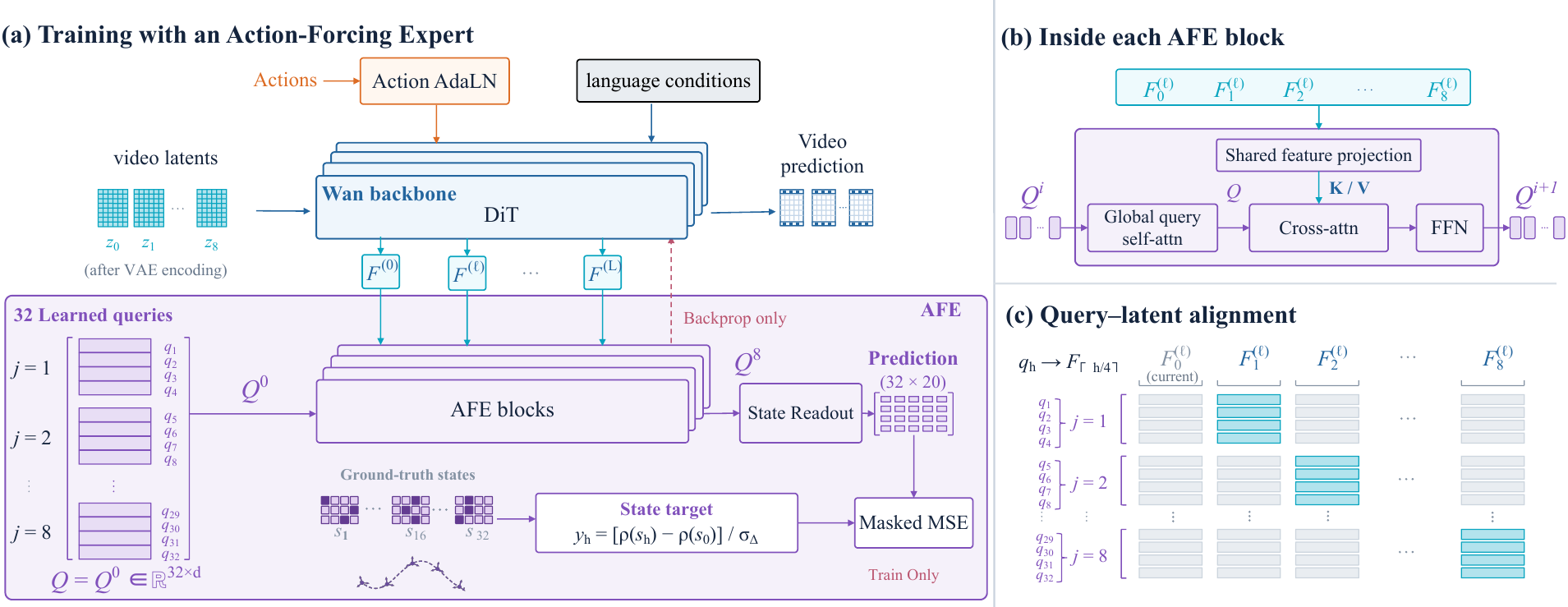}
    \caption{AFE architecture. (a) Training-only state supervision from
    intermediate video features. (b) Query updates within an AFE block.
    (c) Temporal alignment between future-step queries and video features.
    AFE updates the backbone through loss gradients and is removed at inference.}
    \label{fig:afe-architecture}
\end{figure}

\paragraph{Design choices.}
\textbf{AFE architecture.}
Figure~\ref{fig:afe-architecture}(a) shows the two input paths. Actions
condition the Wan video backbone; AFE reads its intermediate video-token
features. Ground-truth robot states enter only the supervision path, where
they define targets for the auxiliary loss. They are not inputs to AFE.
Wan-VAE compresses the current frame and 32 future frames into nine latent
time positions. The backbone processes these noisy latents with visual,
language, and action conditioning.

AFE maintains $H=32$ learnable queries $q_h$, one per future step, arranged
as $Q^0=[q_1;\ldots;q_H]\in\mathbb R^{H\times d}$, where $d$ is the query
feature dimension. The $(i+1)$-th of eight sequential blocks reads the
zero-based backbone layer $\ell_i=8+2i$ and updates $Q^i$ to $Q^{i+1}$,
for $0\leq i<8$. As shown in
Figure~\ref{fig:afe-architecture}(b), each block uses global query
self-attention, video-feature cross-attention, and a feed-forward network,
with pre-normalization and residual connections. A shared LayerNorm--linear
projection maps the backbone features into the AFE feature space.

Let $F_j^{(\ell)}$ contain all spatial tokens from the three camera views
at backbone layer $\ell$ and latent time $j$; $j=0$ denotes the current
frame. Temporal alignment restricts query $q_h$'s video-feature cross-attention to
$F_{\lceil h/4\rceil}^{(\ell)}$. Equivalently, the queries
$\{q_h:4j-3\leq h\leq4j\}$ read future latent time $j$, for
$1\leq j\leq8$ (Figure~\ref{fig:afe-architecture}(c)). These queries predict
distinct future-step targets despite sharing the same video features. A shared
LayerNorm--linear head maps the final queries to the $32\times20$ prediction
$\hat y_{1:H}$.

The selected target is \emph{initial-relative cumulative change}: the future
end-effector state $s_h$ minus the same current state $s_0$, rather than the previous
step's state. Each $s_h\in\mathbb{R}^{20}$ contains both arms' XYZ position,
rotation encoded using the 6D rotation representation, and gripper state. The State target column of
Table~\ref{tab:afe-design-sweep} writes this semantic target as $s_h-s_0$;
the actual training target applies coordinate preprocessing and scaling:
\begin{equation}
    y_h=\frac{\rho(s_h)-\rho(s_0)}{\sigma_{\Delta}},
    \label{eq:afe-cumulative-target}
\end{equation}
where $\rho$ min--max normalizes XYZ and retains the coordinates of the 6D rotation representation and binary
gripper coordinates, and $\sigma_{\Delta}$ is a frozen per-coordinate
standard deviation of the cumulative state changes after applying $\rho$.
Division is element-wise, with $\sigma_\Delta\in\mathbb R^{20}$.
The queries predict $\hat y_{1:H}$ directly and in parallel; no recursive
sum of predicted increments is used. A masked mean squared error loss
supervises the predictions using valid state labels.
This target construction concerns auxiliary supervision, while the world
model is conditioned on actions in the shared delta end-effector action space.
The loss gradients update AFE
and the backbone through the intermediate features. AFE does not write
activations into the video stream and is removed at inference. All results
below therefore evaluate generated videos rather than auxiliary state predictions.

\textbf{Design axes.}
We vary the feature layers, temporal feature access, and state target.
For layers, $\ell_i=\lfloor29i/7+1/2\rfloor$ spans the 30-layer backbone,
whereas $\ell_i=8+2i$ selects its middle layers; both use $0\leq i<8$.
For temporal access, the \emph{Latent times} column in
Table~\ref{tab:afe-design-sweep} contrasts unrestricted access
($0\leq j\leq8$) with temporal alignment ($j=\lceil h/4\rceil$).
Both retain all spatial tokens from the permitted times and global query
self-attention. For the target, we compare absolute state $s_h$,
stepwise change $s_h-s_{h-1}$, and initial-relative cumulative change $s_h-s_0$.
These are semantic descriptions before preprocessing. Absolute-state
supervision uses $\rho(s_h)$, whereas stepwise-change supervision uses
$[\rho(s_h)-\rho(s_{h-1})]/\sigma_{\mathrm{inc}}$.
The selected initial-relative cumulative-change target is $y_h$ in
Equation~\eqref{eq:afe-cumulative-target}. Each change target has its
own frozen per-coordinate standard deviations, estimated after $\rho$ and
floored at $10^{-4}$, without subtracting a mean.

\paragraph{Experimental setup.}
We evaluate AFE design choices on the same queries across four RoboTwin
tasks.
\textbf{Gated error is the primary metric.} It assigns each rollout its
pose-aware normalized dynamic time-warping error (NDTW) if it passes all
visual integrity criteria, and a fixed failure penalty otherwise, then averages within
each task and across tasks. NDTW compares generated and GT trajectories
using position and rotation, with the accumulated alignment cost divided
by the warping-path length.
We also report \emph{raw NDTW} and the
percentage of rollouts passing all visual integrity criteria (\emph{Visual pass}).
Errors are reported in meter-equivalent units and pass rates
as percentages. Bold denotes the best unrounded value in each column.

Table~\ref{tab:afe-design-sweep} reports the design ablation, with IE
disabled and all configurations evaluated at the common 40k checkpoint. Its Baseline is the same action-conditioned Wan model trained
without either auxiliary objective; every other row adds eight AFE blocks.
The individual and joint contributions of AFE and IE are reported in
Table~\ref{tab:ablation}. The design ablation assesses end-effector visibility
using segmentation masks, whereas the component ablation combines masks
with text-based detection at every second frame. Their Gated errors and
Visual pass rates are therefore not directly comparable.

The qualitative cases use Baseline and AFE checkpoints trained for 5k steps
in the design ablation.

\paragraph{Results and analysis.}
\textbf{Design ablation.}
At 40k, middle-layer features improve all three metrics relative to
spanning layers under the same absolute-state target and unrestricted
temporal access (Table~\ref{tab:afe-design-sweep}). Restricting cross-attention
to aligned latent times further improves all three metrics for this target.
Stepwise-change supervision lowers Raw NDTW relative to absolute-state
supervision, but increases Gated error and lowers Visual pass. Cumulative-change
supervision achieves the lowest Gated error and Raw NDTW and the highest
Visual pass among the tested configurations. Relative to Baseline, it reduces
Gated error by 13.02\% and Raw NDTW by 27.33\%, supporting its selection as AFE.

\begin{table}[!htbp]
    \centering
    \small
    \caption{AFE design ablation at 40k training steps with IE disabled. State targets are shown before preprocessing.
    Cumulative-change supervision achieves the lowest Gated error and Raw NDTW.}
    \label{tab:afe-design-sweep}
    \begin{tabular*}{\textwidth}{@{\extracolsep{\fill}}lllrrr@{}}
        \toprule
        Layers $\ell_i$ & Latent times $j$ & State target & Gated error $\downarrow$ & Raw NDTW $\downarrow$ & Visual pass $\uparrow$ \\
        \midrule
        Baseline & -- & -- & 0.0717 & 0.0208 & 82.14 \\
        $\lfloor29i/7+1/2\rfloor$ & $0\leq j\leq8$ & $s_h$ & 0.0816 & 0.0210 & 78.57 \\
        $8+2i$ & $0\leq j\leq8$ & $s_h$ & 0.0784 & 0.0197 & 79.29 \\
        $8+2i$ & $\lceil h/4\rceil$ & $s_h$ & 0.0702 & 0.0191 & 82.14 \\
        $8+2i$ & $\lceil h/4\rceil$ & $s_h-s_{h-1}$ & 0.0714 & 0.0159 & 80.71 \\
        $8+2i$ & $\lceil h/4\rceil$ & $s_h-s_0$ & \textbf{0.0624} & \textbf{0.0151} & \textbf{83.57} \\
        \bottomrule
    \end{tabular*}
\end{table}

This sweep selects a configuration, without isolating temporal alignment
or target semantics: the change targets also alter coordinate scaling,
and cumulative targets with unrestricted temporal access were not tested.

\begin{figure}[!htbp]
    \centering
    \includegraphics[width=0.85\textwidth]{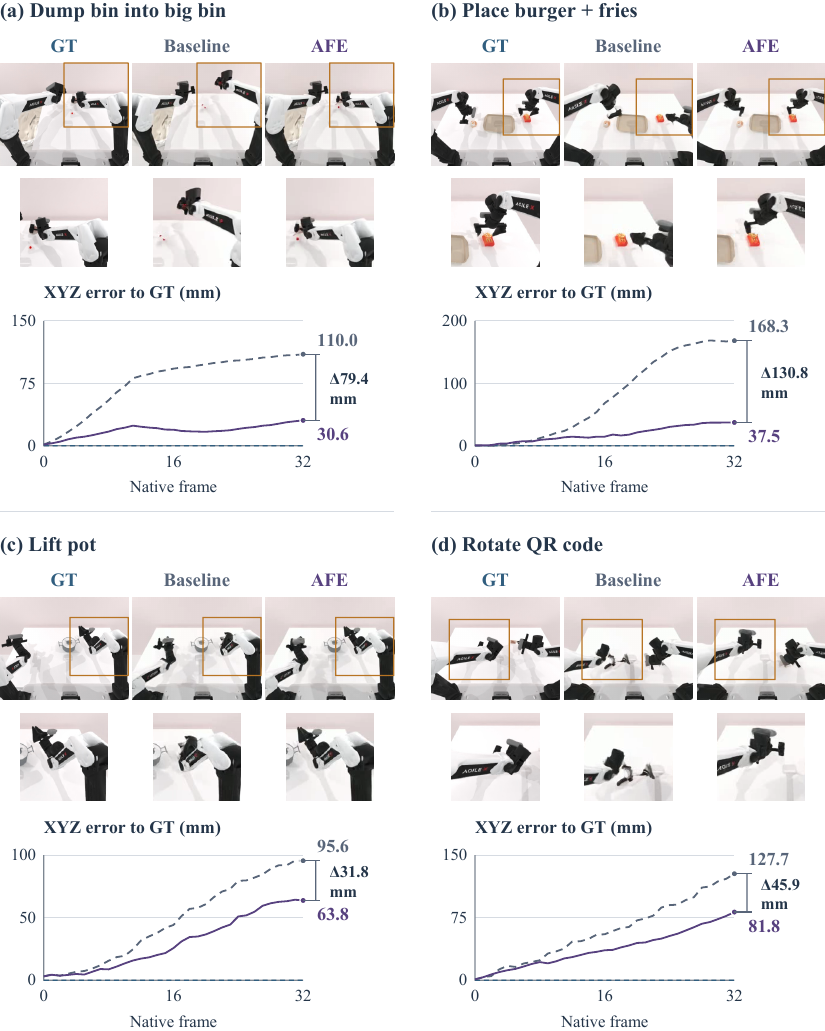}
    \caption{Selected Baseline and AFE video predictions at 5k training
    steps. Images show final frames and hand details; curves report position
    errors averaged over both arms in millimeters. Endpoint labels show final-frame
    errors, and brackets mark the Baseline--AFE differences. Cases illustrate
    improvements rather than average performance.}
    \label{fig:afe-m06-state-trajectories}
\end{figure}

\textbf{Generated-video evidence.}
Figure~\ref{fig:afe-m06-state-trajectories} compares GT, Baseline, and AFE
for matched initial observations and action sequences in the design
ablation. Terminal frames and identical hand-region crops reveal reduced
hand-position drift in the four selected cases. The curves report
framewise position errors estimated using AnyPos and averaged over both arms,
whereas each crop shows one hand. The errors use corresponding frame indices
without DTW alignment, exclude rotation, and are reported in millimeters.
They complement, rather than reproduce, the pose-aware NDTW metric.
For each task, we select the query with the largest absolute Raw NDTW
reduction among those passing both models' visual integrity criteria.
All images use the head camera at frame 32, with identical crops across
methods showing the right hand in panels a--c and the left hand in panel d.
The curves include the initial frame and 32 future frames.

These cases illustrate large improvements selected within each task, not
average behavior or evidence for IE+AFE. Gripper, rotation, and
opposite-arm errors remain visible, particularly in pot lifting and QR-code
rotation. Reduced position error alone does not establish task completion.
The aggregate evidence for the AFE design comes from
Table~\ref{tab:afe-design-sweep}.
\FloatBarrier

\subsection{Data Composition Ablations}
\label{app:data-composition-ablations}

\paragraph{Setup.}
We examine how progressively broadening the training-data composition affects
action following. We use \method with IE and AFE disabled to isolate the
role of data composition. Starting from expert demonstrations, we successively
include feasible-space sampling, local-perturbation, cross-state-replay, and
policy-rollout data. All configurations are evaluated at the 40k checkpoint
on four RoboTwin tasks with three camera views and the shared end-effector
action representation, following the \benchmark evaluation protocol defined
in the main text.

\begin{table}[!htbp]
    \centering
    \small
    \renewcommand{\arraystretch}{1.08}
    \caption{Incremental data-composition ablation at 40k with IE and AFE
    disabled. Each row retains all data families in the preceding row and
    adds the named family. The final row includes all four off-expert
    families and constitutes Expanded Action Coverage. Bold denotes the
    best value in each metric column.}
    \label{tab:data-composition}
    \begin{tabular*}{\textwidth}{@{\extracolsep{\fill}}lrrr@{}}
        \toprule
        Training data & Gated error $\downarrow$ & Raw NDTW $\downarrow$ & Visual pass (\%) $\uparrow$ \\
        \midrule
        Expert Demonstrations & 0.0352 & 0.0314 & \textbf{98.57} \\
        \quad + Feasible-Space Sampling & 0.0336 & 0.0282 & 97.86 \\
        \quad + Local Perturbation & 0.0330 & 0.0270 & 97.86 \\
        \quad + Cross-State Replay & 0.0300 & 0.0242 & 97.86 \\
        \quad \tablemethod{+ Policy Rollout} & \textbf{0.0264} & \textbf{0.0204} & 97.86 \\
        \bottomrule
    \end{tabular*}
\end{table}

\paragraph{Results.}
Table~\ref{tab:data-composition} shows that the complete mixture reduces
integrity-gated error from 0.0352 to 0.0264 and raw NDTW from 0.0314 to 0.0204
relative to expert-only training, corresponding to reductions of 25.0\% and
35.0\%, respectively. Visual pass rate decreases from 98.57\% to 97.86\%.
Both errors decrease at every stage of the incremental comparison. Adding
feasible-space data lowers both errors despite a decrease in visual pass
rate; subsequent additions of local-perturbation and cross-state-replay data
further reduce both errors while maintaining visual pass rate. The final
addition of policy-rollout data yields the largest successive reduction in
both errors while maintaining visual pass rate at 97.86\%.

These results support broadening training beyond expert demonstrations under
the evaluated configuration. The successive gains are conditional on the
data families already present: they characterize this incremental composition
path rather than establish an independent contribution ranking among families.

\FloatBarrier

\clearpage
\section{Additional Quantitative Results}
\label{app:additional-quantitative-results}

\subsection{Results across Query Categories}
\label{app:query-category-results}

\begin{figure}[!ht]
    \centering
    \includegraphics[width=\textwidth]{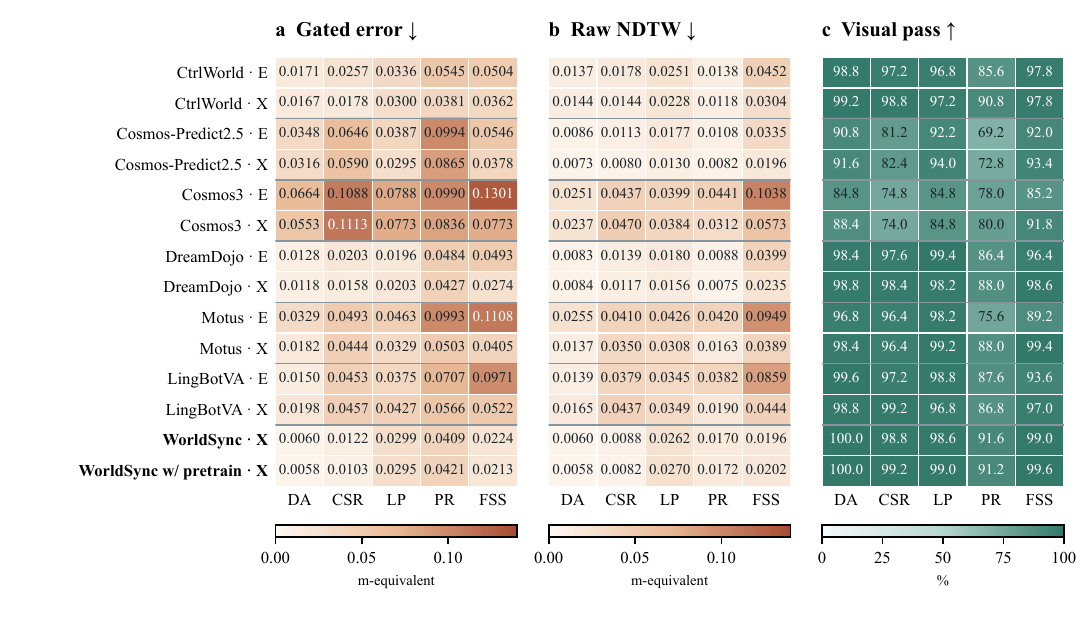}
    \caption{Results across query categories. Panels show gated error, raw NDTW,
    and visual pass rate. E/X denote Expert Demonstrations/Expanded Action Coverage.
    DA, CSR, LP, PR, and FSS denote Demonstrated Action, Cross-State Replay,
    Local Perturbation, Policy Rollout, and Feasible-Space Sampling.
    Cells give point estimates, and both error panels share an absolute color scale.}
    \label{fig:query-category-heatmap}
\end{figure}

\begin{figure}[!ht]
    \centering
    \includegraphics[width=\textwidth]{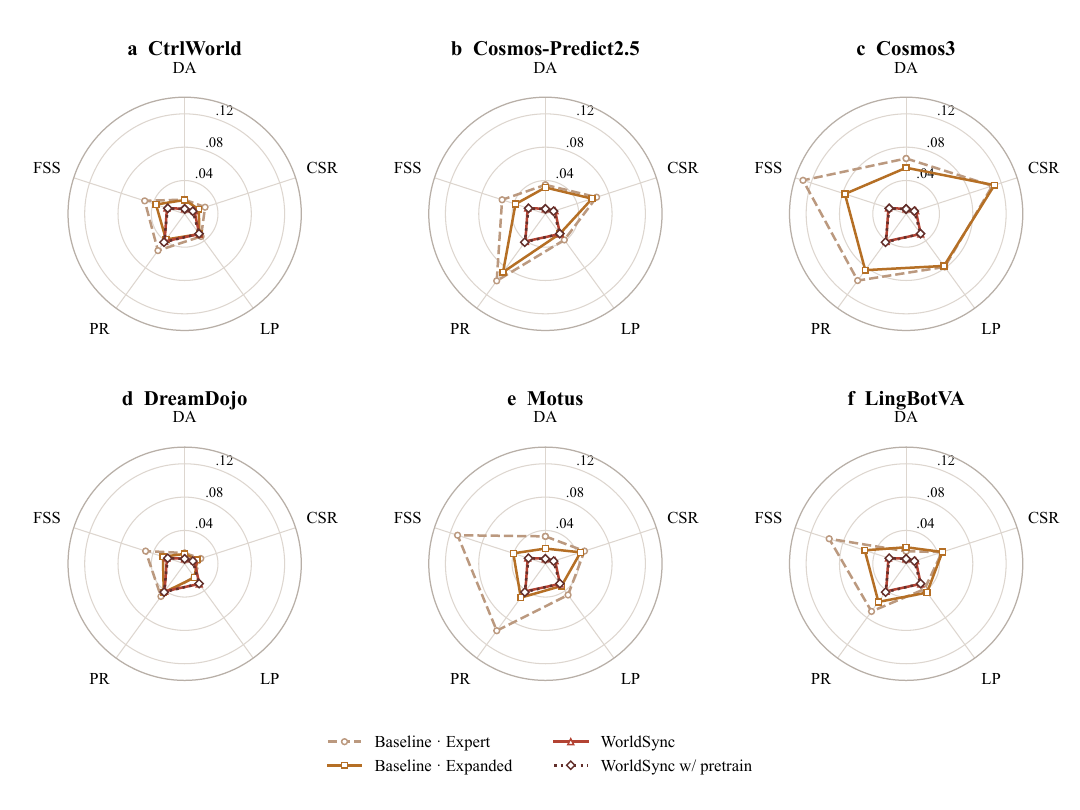}
    \caption{Gated error by query category. Each panel compares one baseline
    under both training regimes with the two \method variants, repeated as
    common references. All axes share a 0--0.14 m-equivalent scale,
    with lower error closer to the center. Category abbreviations follow
    Figure~\ref{fig:query-category-heatmap}.}
    \label{fig:query-category-gated}
\end{figure}

\begin{figure}[!ht]
    \centering
    \includegraphics[width=\textwidth]{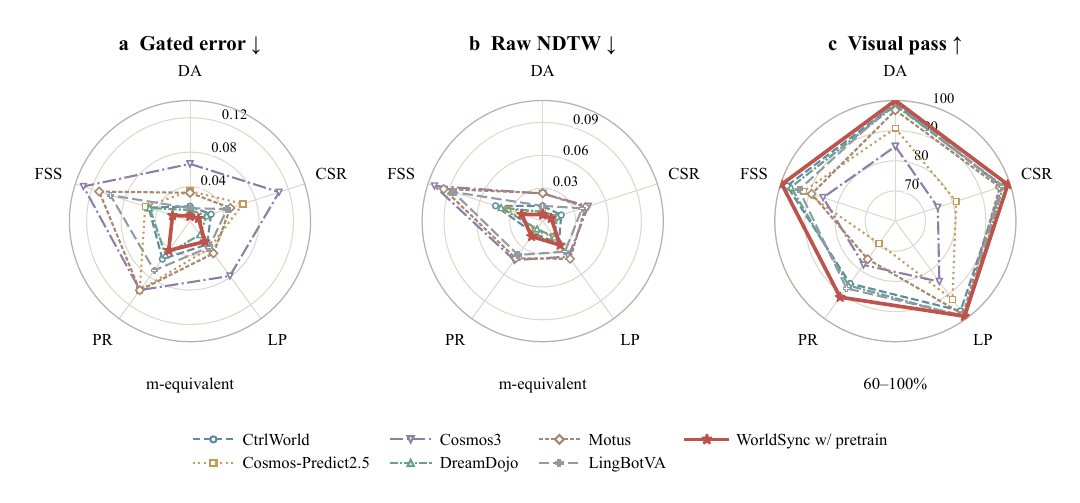}
    \caption{Three-metric profiles of expert-trained baselines and \method
    with additional robotic pretraining and Expanded Action Coverage.
    Error axes start at zero, while visual pass spans 60--100\%.
    Lower error lies closer to the center and higher pass rate farther out.
    Abbreviations follow Figure~\ref{fig:query-category-heatmap}.}
    \label{fig:query-category-radar}
\end{figure}

\paragraph{Evaluation scope.}
We complement the overall comparison in Table~\ref{tab:main} with results
for each of the five action-query categories under the same evaluation protocol.
Figure~\ref{fig:query-category-heatmap} reports all model configurations,
while Figures~\ref{fig:query-category-gated} and~\ref{fig:query-category-radar}
visualize their category-level profiles. Raw NDTW and gated error retain
task-macro aggregation, and visual pass rate counts accepted rollouts.

\paragraph{Expanded action coverage.}
Expanded Action Coverage reduces both raw NDTW and gated error on
Policy Rollout and Feasible-Space Sampling for all six baselines
(Figures~\ref{fig:query-category-heatmap} and~\ref{fig:query-category-gated}).
These gains extend across actions produced during policy execution and
actions sampled from the broader feasible space, providing category-level
support for the overall benefits of expanded training coverage.

\paragraph{WorldSync across categories.}
Both \method variants achieve the two lowest gated-error point estimates on
Demonstrated Action, Cross-State Replay, and Feasible-Space Sampling
(Figure~\ref{fig:query-category-heatmap}). With additional robotic pretraining,
the errors are 0.0058, 0.0103, and 0.0213, respectively. These correspond to
reductions of 50.6\%, 34.7\%, and 22.3\% relative to expanded-coverage
DreamDojo, the lowest-scoring baseline in each category.
The gains therefore span demonstrated actions, replay from a different
initial state, and sampling from the broader feasible action space.

\paragraph{Trajectory agreement and visual integrity.}
The raw NDTW and visual pass results further characterize these advantages
(Figures~\ref{fig:query-category-heatmap} and~\ref{fig:query-category-radar}).
On Demonstrated Action, the two \method variants achieve the lowest two
raw errors and both pass every visual check on all evaluated rollouts.
On Feasible-Space Sampling, the pretrained variant combines the lowest
gated error with the highest visual pass rate of 99.6\%.
These results support the gated-error improvements with specific evidence
of trajectory agreement and visual integrity in the corresponding categories.

\clearpage

\clearpage

\subsection{Metric Distributions and Analysis}
\label{app:metric-distributions-thresholds}

To support policy evaluation, world-model metrics must distinguish differences
in visual integrity and action following. Blurred imagery, distorted robot
structures, and motion inconsistent with the commanded actions can each make
generated rollouts unreliable for policy learning. We therefore examine
whether \benchmark scores and gate decisions correspond to observable
differences in generated videos. We combine quantitative score distributions
and per-check pass/fail rates with qualitative examples across score ranges
to connect the metric values to visual integrity and action following.

\subsubsection{Metric Distributions}
\begin{figure}[!ht]
    \centering
    \includegraphics[width=\textwidth]{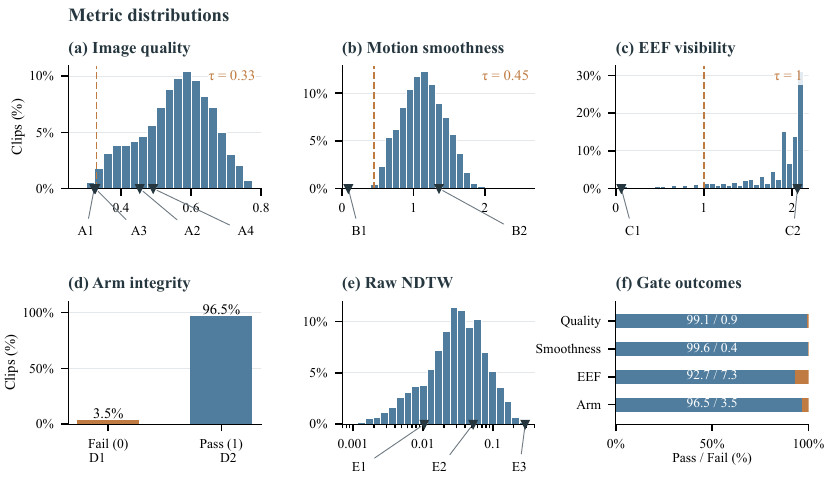}
    \caption{Metric distributions of generated clips.
    Panels (a--e) show the five metrics; case labels correspond to
    Figure~\ref{fig:threshold-cases}, and dashed lines mark thresholds.
    Panel (f) reports each check's Pass/Fail percentages (blue/orange), with
    overlapping failures allowed. Percentages use the full evaluation set,
    including clips omitted from motion scoring.}
    \label{fig:metric-distributions}
\end{figure}

Figure~\ref{fig:metric-distributions} shows metric distributions for over 10k
generated clips from early training checkpoints, spanning 50 tasks and all
five action-query categories. Each category includes over 200 distinct action queries.
We retain all evaluated clips when plotting distributions,
including visually rejected clips. The image-quality and motion-smoothness
thresholds are $\tau_q=0.33$ and $\tau_m=0.45$. To visualize the binary EEF
visibility check, we normalize the detection counts from tracking and
text-prompted detection by their respective required counts and plot the
larger value; a value of at least 1 passes the check. Arm integrity
is binary, while raw NDTW is continuous and has no pass/fail threshold.

EEF visibility has the highest rejection rate in this evaluation set
(Figure~\ref{fig:metric-distributions}(f)): it rejects 7.3\% of clips,
compared with 0.9\% for image quality, 0.4\% for motion smoothness, and
3.5\% for arm integrity. Reporting these outcomes separately reveals the
frequency of violations in each visual dimension alongside the overall
pass/fail decision. Each check is counted separately, so a clip may fail
multiple checks. Clips with no dynamic frame pairs pass the motion check
and are omitted from its score histogram.

\FloatBarrier
\subsubsection{Metric Analysis}
\begin{figure}[p]
    \centering
    \includegraphics[width=\textwidth]{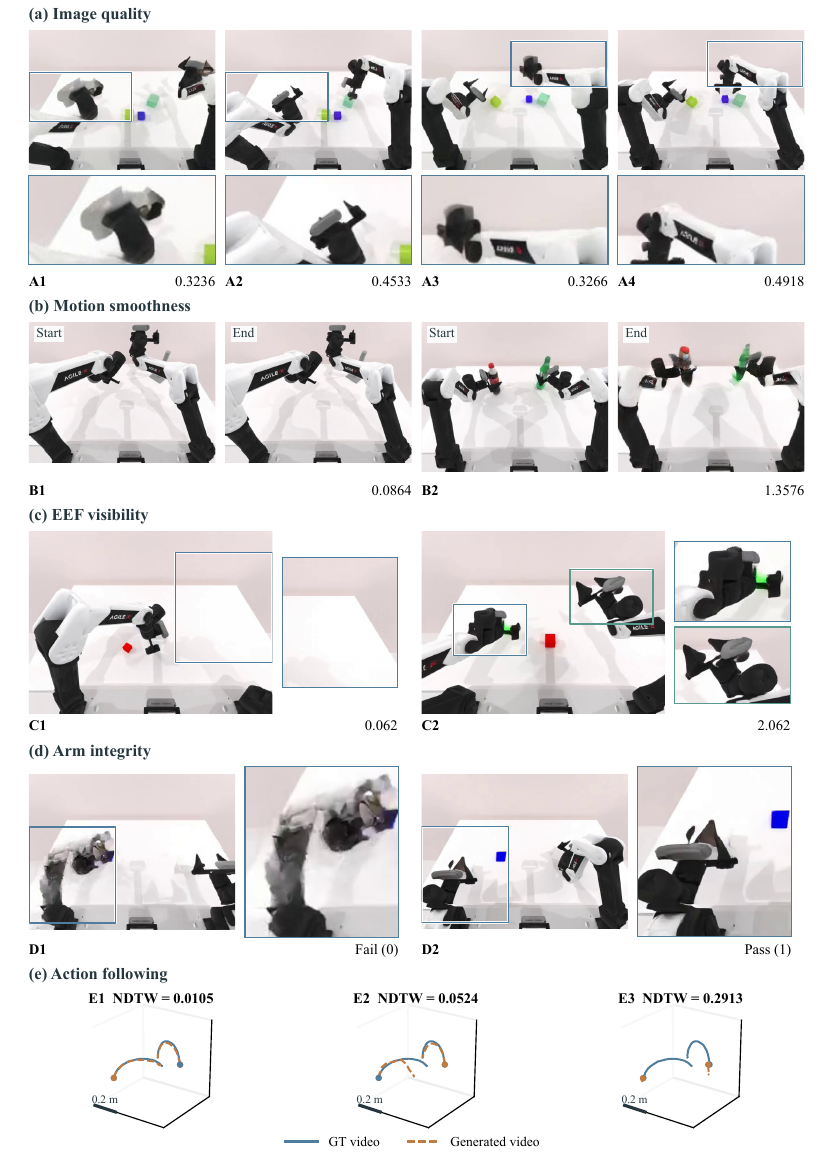}
    \caption{Examples corresponding to the case labels in
    Figure~\ref{fig:metric-distributions}. Scores are computed on full videos.
    Colored boxes locate the enlarged regions;
    C2 enlarges the left and right EEFs above and below, respectively.
    In (e), solid blue and dashed orange paths are estimated from ground-truth
    (GT) and generated videos on shared 3D axes; dots mark starts.}
    \label{fig:threshold-cases}
\end{figure}

Figure~\ref{fig:threshold-cases} connects the metric values to appearance,
robot structure, and motion agreement.
For image quality, each pair (A1/A2 and A3/A4) uses the same action query
and native resolution. The lower-scoring examples show wrist blur
(A1, 0.3236; A3, 0.3266), whereas A2 (0.4533) and A4 (0.4918) are
clearer and exceed 0.33.

Following WorldArena~\citep{WorldArena}, motion smoothness is assessed through
frame-interpolation reconstruction, weighted by motion magnitude.
B1 and B2 receive sequence-level scores of 0.0864 and 1.3576, respectively.
The displayed start and end frames provide visual context for scores
computed over the full sequences.

For EEF visibility, C1 shows the right EEF out of view, while C2 shows
both EEFs; their normalized detection counts are 0.062 and 2.062. These
examples use different queries. For arm integrity, D1 and D2 share the
same query: D1 shows arm distortion and fails the check, whereas D2
preserves arm structure and passes.

For action following, E1--E3 share one query and have raw NDTW errors of
0.0105, 0.0524, and 0.2913. The estimated position paths show increasing
mismatch with trajectories estimated from ground-truth videos. The position
paths visualize the spatial component of the pose discrepancy, while raw
NDTW jointly measures position and rotation agreement.

Together, these examples connect the \benchmark scores to appearance,
robot visibility and structure, and action following, providing an
interpretable diagnosis of generated rollout quality.

\FloatBarrier
\subsection{Agreement with Human Evaluation}
\label{app:human-evaluation-agreement}

We examined how WorldEcho's automatic measurements correspond to human judgments, complementing the metric distributions and examples in Appendix~\ref{app:metric-distributions-thresholds}. We first assess whether raw NDTW captures human judgments of action following, then examine the four visual checks used for preliminary screening.

\paragraph{Evaluation protocol.}
Eleven raters provided 544 ratings for 100 clips, with five or six ratings per clip. Forty action clips received ratings on a five-point scale, with higher scores indicating greater action agreement. The remaining 60 clips received binary pass/fail judgments, with 15 clips each for image quality, motion smoothness, EEF visibility, and arm integrity. We retained all observed ratings, using the median for each action clip and a strict majority for each visual clip. No visual clip had a tied vote.

\paragraph{Sampling and analysis.}
The action set covers four tasks and five action-query categories, with two clips per combination. The tasks are \texttt{dump\_bin\_bigbin}, \texttt{lift\_pot}, \texttt{place\_burger\_fries}, and \texttt{rotate\_qrcode}. For action following, we computed Spearman's correlation between raw NDTW and median human ratings. A 95\% percentile interval was obtained from 10,000 bootstrap resamples of clips within each task, retaining each clip's observed ratings. For visual metrics, we compared the recorded automatic decisions with human majority labels using accuracy, pass recall, failure recall, and Cohen's $\kappa$. Krippendorff's $\alpha$ measures inter-rater agreement, using ordinal distances for action scores and nominal distances for visual labels.

\begin{figure}[!ht]
    \centering
    \includegraphics[width=\textwidth]{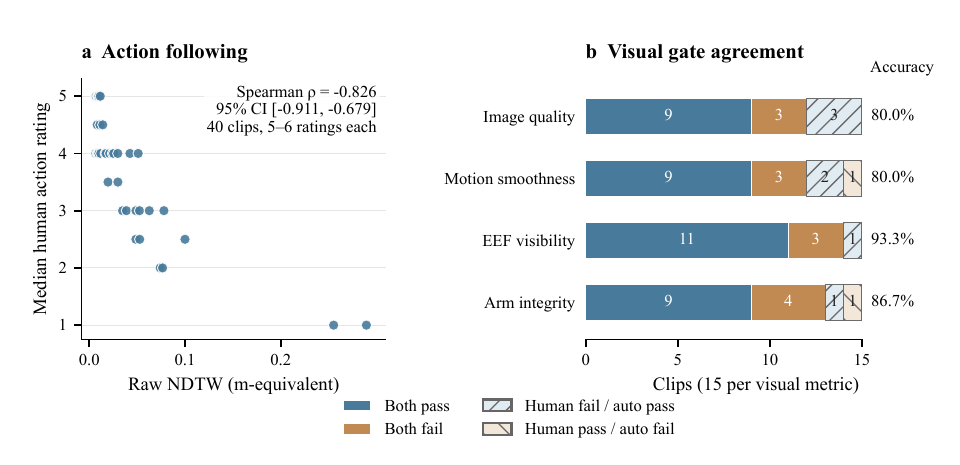}
    \caption{Automatic measurements and human judgments. (a) Raw NDTW versus median human ratings for 40 clips with five or six ratings each. Points retain exact coordinates, and marker area scales with the number of coincident clips. The interval is a task-stratified bootstrap interval for Spearman's $\rho$. (b) Joint human and automatic pass/fail decisions for 15 clips per visual metric. Human labels use a strict majority of five or six ratings. Solid segments indicate agreement, and hatched segments indicate disagreement. Segment labels give clip counts, and percentages give total agreement.}
    \label{fig:human-evaluation}
\end{figure}

\paragraph{Action-following metric.}
Raw NDTW showed a strong negative association with median human action ratings (Figure~\ref{fig:human-evaluation}(a)). Spearman's $\rho$ was $-0.826$, with a 95\% CI of $[-0.911,-0.679]$, indicating that lower trajectory errors corresponded to higher perceived action agreement. This result supports trajectory alignment as the central measure of action following in WorldEcho across the sampled tasks and action-query categories. Ordinal Krippendorff's $\alpha$ was 0.594, so the correlation reflects agreement with aggregated human judgments despite variation among individual raters.

\paragraph{Visual metrics.}
The visual checks are intended to screen for basic visual integrity while preserving clips for action-following evaluation. Automatic decisions matched human majority labels on 80.0\%, 80.0\%, 93.3\%, and 86.7\% of image-quality, motion-smoothness, EEF-visibility, and arm-integrity clips, respectively (Figure~\ref{fig:human-evaluation}(b)). Each check retained 90.0\%--100.0\% of human-passed clips, while rejecting 50.0\%--80.0\% of human-failed clips (Table~\ref{tab:human-visual-agreement}). This asymmetry is consistent with the intended permissive screening, which limits visual rejection so that retained clips can be evaluated for action following. For clips passing all four checks, raw NDTW determines the official error, keeping trajectory agreement central to the evaluation. Some human-rated visual failures nevertheless passed the automatic checks, so this screening does not guarantee visual integrity. These per-check results describe 15 sampled clips per dimension.

\clearpage

\begin{table}[!ht]
    \centering
    \caption{Visual classification and inter-rater agreement. Each metric uses 15 clips with five or six ratings per clip. Pass recall is the fraction of human-passed clips accepted automatically. Failure recall is the fraction of human-failed clips rejected automatically. Cohen's $\kappa$ compares automatic decisions with majority labels, while nominal Krippendorff's $\alpha$ compares individual raters.}
    \label{tab:human-visual-agreement}
    \small
    \setlength{\tabcolsep}{5pt}
    \begin{tabular}{lrrrrr}
        \toprule
        \textbf{Visual metric} & \shortstack{\textbf{Accuracy}\\\textbf{(\%)}} & \shortstack{\textbf{Pass recall}\\\textbf{(\%)}} & \shortstack{\textbf{Failure recall}\\\textbf{(\%)}} & \textbf{Cohen's $\kappa$} & \textbf{Rater $\alpha$} \\
        \midrule
        Image quality & 80.0 & 100.0 & 50.0 & 0.545 & 0.815 \\
        Motion smoothness & 80.0 & 90.0 & 60.0 & 0.526 & 0.605 \\
        EEF visibility & 93.3 & 100.0 & 75.0 & 0.815 & 0.651 \\
        Arm integrity & 86.7 & 90.0 & 80.0 & 0.700 & 0.888 \\
        \bottomrule
    \end{tabular}
\end{table}
\clearpage

\section{Additional Qualitative Results}
\label{app:additional-qualitative-results}

\begin{figure*}[p]
    \centering
    \includegraphics[width=\textwidth]{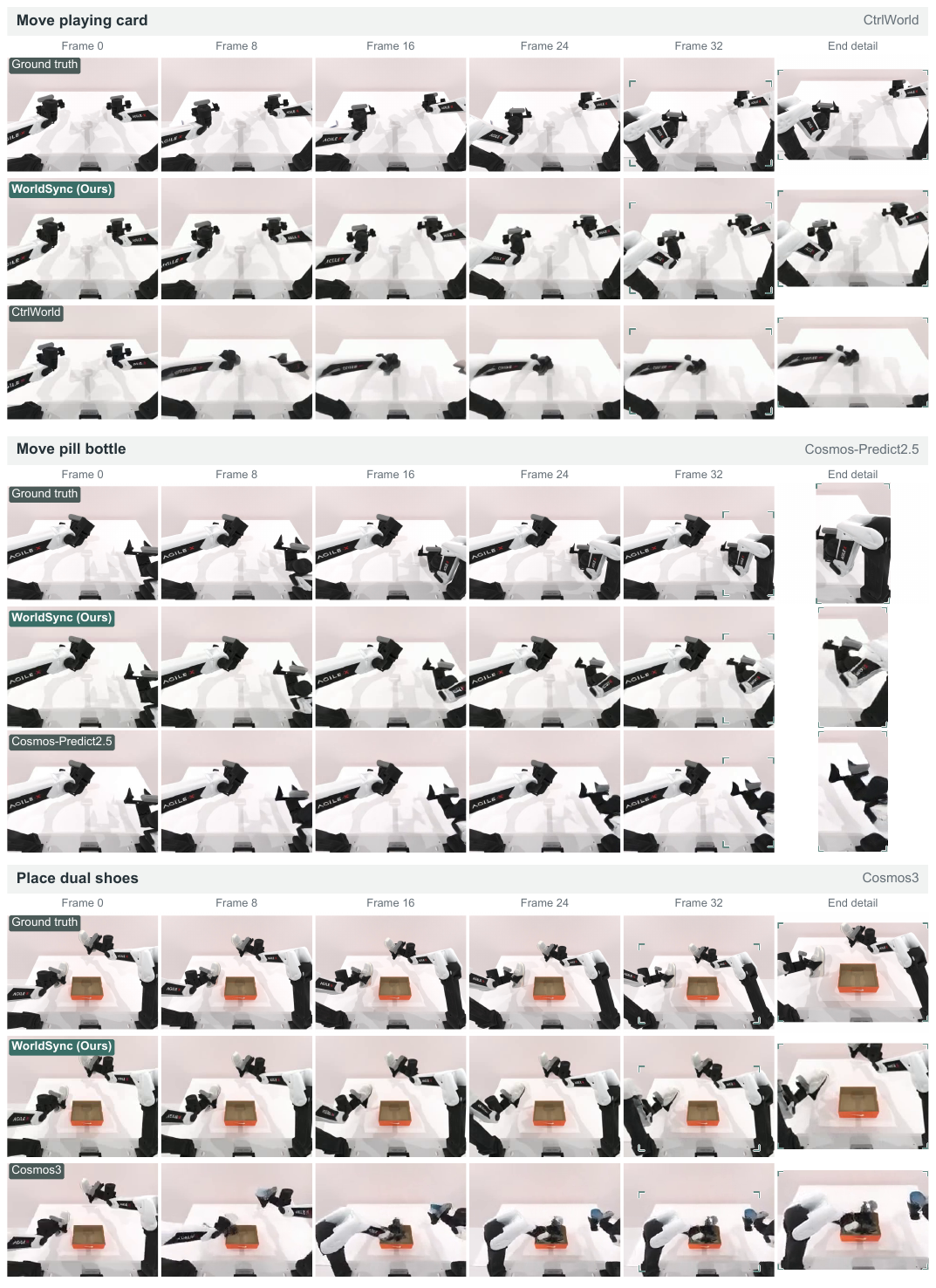}
    \caption{Qualitative comparisons of WorldSync with CtrlWorld,
    Cosmos-Predict2.5, and Cosmos3. Rows show ground truth, WorldSync,
    and the baseline for each task. The last column enlarges the marked final-frame region.}
    \label{fig:qual-comparison-a}
\end{figure*}

\begin{figure*}[p]
    \centering
    \includegraphics[width=\textwidth]{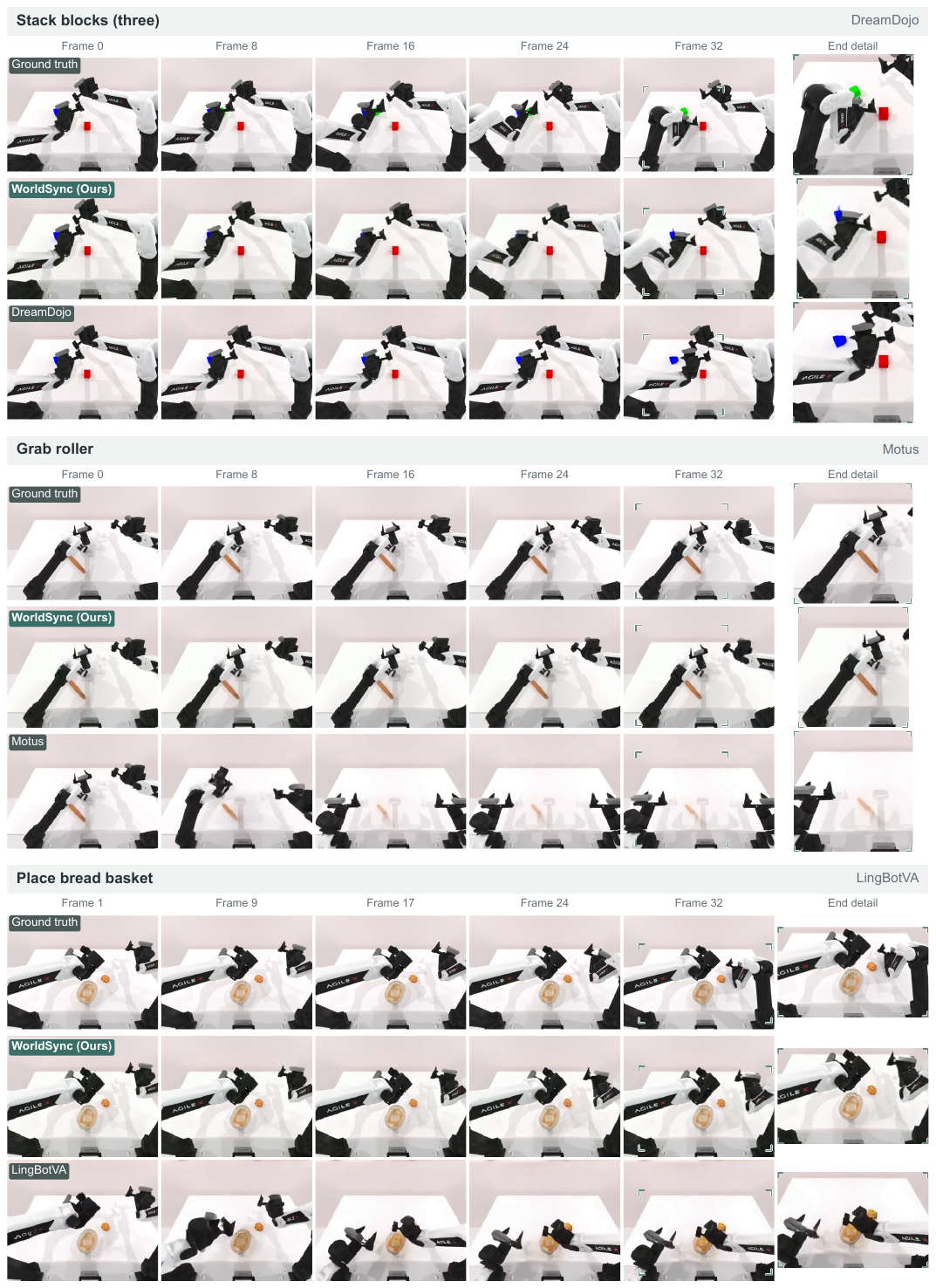}
    \caption{Qualitative comparisons of WorldSync with DreamDojo, Motus,
    and LingBotVA. Rows show ground truth, WorldSync, and the baseline for each task.
    The last column enlarges the marked final-frame region.}
    \label{fig:qual-comparison-b}
\end{figure*}

\begin{figure*}[p]
    \centering
    \includegraphics[width=\textwidth]{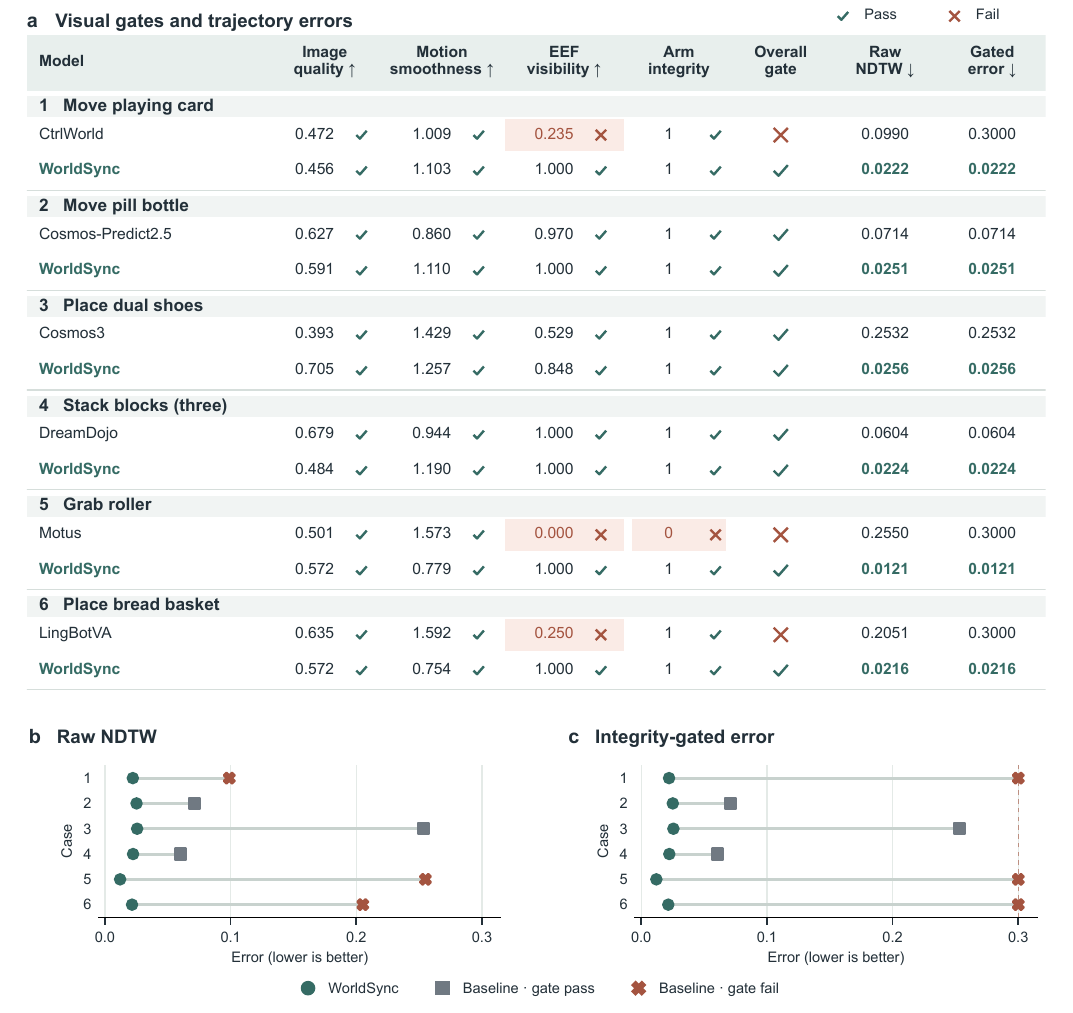}
    \caption{WorldEcho metrics for the six cases in
    Figures~\ref{fig:qual-comparison-a} and~\ref{fig:qual-comparison-b}.
    (a) Scores and gate decisions. (b,c) Raw NDTW and integrity-gated error.
    Image quality and motion smoothness pass at $\geq 0.33$ and $\geq 0.45$,
    respectively. EEF visibility passes when either tracking or text-prompted
    detection meets its requirement; arm integrity must equal 1. The displayed
    EEF fraction comes from tracking when that branch passes and from text
    detection otherwise; the fractions do not share a universal 0.5 cutoff.
    The overall gate requires all four checks; gated error equals raw NDTW
    when passed and $0.3$ otherwise.}
    \label{fig:qual-case-metrics}
\end{figure*}

Figures~\ref{fig:qual-comparison-a} and~\ref{fig:qual-comparison-b}
compare WorldSync with six baseline models on simulation tasks from WorldEcho.
For each comparison, both models receive the same initial observation and
action sequence, with the simulator replay providing ground truth.
Rows show ground truth, WorldSync, and the corresponding baseline at matched
frame indices. The final column enlarges the marked region using the same
relative image coordinates across rows, highlighting local pose and shape
differences.
Figure~\ref{fig:qual-case-metrics} reports the corresponding WorldEcho
metrics, computed on the complete generated videos rather than the displayed
frames or enlarged regions.

Figure~\ref{fig:qual-comparison-a} shows differences in arm structure and
gripper motion. In \textit{Move playing card}, CtrlWorld produces distorted
arm and gripper shapes. WorldSync better preserves their structure and
follows the downward movement in the ground-truth replay. In
\textit{Move pill bottle}, Cosmos-Predict2.5 largely retains the initial
right-gripper rotation, whereas WorldSync captures part of the wrist
rotation. In \textit{Place dual shoes}, Cosmos3 moves the grippers toward
the shoe box. WorldSync keeps them to its sides, more closely matching
the ground-truth replay.

Figure~\ref{fig:qual-comparison-b} shows differences in wrist motion and
arm placement. In \textit{Stack blocks (three)}, DreamDojo largely preserves
the initial left-wrist pose. WorldSync follows the direction of the
wrist motion, although its amplitude remains smaller than in the
ground-truth replay. In \textit{Grab roller}, Motus retracts the extended
arm, while WorldSync more closely maintains its ground-truth placement.
In \textit{Place bread basket}, LingBotVA moves the grippers toward the
basket, whereas WorldSync better preserves their placement relative to
ground truth.

The per-case metrics in Figure~\ref{fig:qual-case-metrics} distinguish
visual validity from motion agreement. WorldSync passes all four visual
checks in all six cases and has lower raw NDTW in each comparison.
CtrlWorld, Motus, and LingBotVA fail the end-effector visibility
check; Motus also fails arm integrity.
Their integrity-gated errors therefore take the fixed value of 0.3.
Cosmos-Predict2.5, Cosmos3, and DreamDojo pass all visual checks, but their raw NDTW
errors remain higher than WorldSync's (0.0714 versus 0.0251, 0.2532 versus
0.0256, and 0.0604 versus 0.0224, respectively). These cases show that passing the visual
gate does not ensure agreement with the conditioned motion.

The individual scores also expose differences that the overall gate does
not rank. For example, DreamDojo has higher image quality than WorldSync
(0.679 versus 0.484), despite its larger trajectory error.
Likewise, a passing arm-integrity decision does not rule out local shape
distortions, as the CtrlWorld example illustrates. The visualizations and
per-case metrics therefore provide complementary evidence.

These selected examples illustrate closer agreement with ground-truth
motion, alongside remaining differences in wrist rotation and object
appearance. They were chosen after assessing trajectory error, visual
integrity, and the complete generated sequences. The comparisons illustrate
specific visual discrepancies; aggregate performance over the full
evaluation set is reported in Table~\ref{tab:main}.

\clearpage

\end{document}